\documentclass[dvipsnames]{article}
\usepackage{PRIMEarxiv}

\usepackage[utf8]{inputenc} % allow utf-8 input
\usepackage{hyperref}       % hyperlinks
\usepackage{url}            % simple URL typesetting
\usepackage{booktabs}       % professional-quality tables
\usepackage{amsfonts}       % blackboard math symbols
\usepackage{nicefrac}       % compact symbols for 1/2, etc.
\usepackage{microtype}      % microtypography
\usepackage{lipsum}
\usepackage{fancyhdr}       % header
\usepackage{graphicx}       % graphics
\graphicspath{{media/}}     % organize your images and other figures under media/ folder

\usepackage{graphicx}
\usepackage[most]{tcolorbox}
\usepackage{caption}
\usepackage{subcaption}
\usepackage{longtable}

\usepackage{amsmath,amssymb,amsfonts}
\usepackage{textcomp}

\usepackage{verbatim}

\usepackage{xcolor}
\usepackage{url}
\usepackage{xr-hyper}
\usepackage{hyperref}
\usepackage{pdfpages} 
\usepackage{bm}
\usepackage{multirow}
\usepackage{multicol}
\usepackage{array}
\newcolumntype{L}{>{\arraybackslash}m{3cm}}

\usepackage{fullpage}
\usepackage{rotating}
\usepackage{stmaryrd}
\usepackage{proof}

\usepackage{tikz}
\usetikzlibrary{bayesnet}
\usetikzlibrary{decorations.pathmorphing} % noisy shapes
\usetikzlibrary{fit}% fitting shapes to coordinates
\usetikzlibrary{backgrounds}	

\usepackage{amsthm}
\usepackage{amsmath}
\theoremstyle{definition}

\newtheorem{definition}{Definition}
\newtheorem*{definition*}{Definition}
\newtheorem*{hypothesis*}{Hypothesis}

\newtheorem*{remark*}{Remark}

\usepackage[ruled]{algorithm2e}
\usepackage{algorithmic} %% Used in Chapter 1 

\usepackage[style=numeric,sorting=none,firstinits=true]{biblatex}
\title{Deep incremental learning for financial temporal tabular datasets with distribution shifts
}

\author{
  Thomas Wong \\
  Imperial College London \\
  London\\
  \texttt{mw4315@ic.ac.uk} \\
   \And
  Mauricio Barahona \\
  Imperial College London \\
  London\\
  \texttt{m.barahona@imperial.ac.uk} \\
}

\begin{document}
\maketitle

\begin{abstract}
We present a robust deep incremental learning framework for regression-based ranking tasks on financial temporal tabular datasets which is built upon the incremental use of commonly available tabular and time series prediction models to adapt to distributional shifts typical of financial datasets.
The framework uses a simple basic building block (decision trees) to build hierarchical models of any required complexity to deliver robust performance under adverse situations such as regime changes, fat-tailed distributions, and low signal-to-noise ratios. 
As a detailed study, we demonstrate our scheme using XGBoost models trained on the Numerai dataset and show that a two layer deep ensemble of XGBoost models over different model snapshots delivers high quality predictions under different market regimes. We also show that the performance  of XGBoost models with different number of boosting rounds in three scenarios (small, standard and large) is monotonically increasing with respect to model size and converges towards the generalisation upper bound. We also evaluate the robustness of the model under variability of different hyperparameters, such as model complexity and data sampling settings. Our model has low hardware requirements as no specialised neural architectures are used and each base model can be independently trained in parallel.
\end{abstract}

\keywords{Machine Learning, Time Series Prediction, Deep Learning, }

\section{Introduction} 
\label{section:NumeraiSunshine-overview}

%% At the request of reviwers, need to add more references for background
%\subsection{Background on IL}

Many important applications of machine learning (ML), such as the Internet of Things (IoT) \cite{song2018situ} and cyber-security \cite{buczak2015survey}, involve data streams, where data is regularly updated and predictions are made \textit{point-in-time}. 
Such applications pose challenges to standard ML approaches, specifically with regard to the balance between model learning and their update in response to new data arrivals \cite{Gama14,}.

Incremental learning (IL) techniques~\cite{belouadah2021comprehensive,wu2019large,van2022three} are used to adapt deployed machine learning systems to changes in data streams. For example, in image classification systems, class incremental learning \cite{zhu2021class,NEURIPS2022_c8ac22c0,NEURIPS2022_ae817e85} is used where the categories of images cannot be known in advance. A key challenge in IL is the presence of distributional shifts in data (or concept drifts) \cite{Jie19,} which results in model degradation \cite{bayram2022concept,} during inference, i.e., deterioration of out-of-sample performance when the model learns relationships from the training set that significantly differ from those in the test set. 

Reinforcement Learning (RL) \cite{arulkumaran2017deep,NEURIPS2022_d112fdd3,NEURIPS2022_eb4898d6} provides an alternative approach to prediction tasks in systems under data innovation. In RL, a model (agent) learns a policy to optimise its reward by interacting and eliciting a response from the environment. 
RL is therefore useful when the actions of the model influence the environment, and when multiple agents interact with each other \cite{NEURIPS2020_77441296,}. However, if the actions of models have no influence on the data stream, 
%when the reward of agents is not used as input features. 
%In other words, the input features (environment) are \textit{independent} to the actions of the agent (predictions of the model) 
(i.e., there is no feedback between agent and environment), then RL reduces to incremental learning. 
Furthermore, applying trained RL agents to unknown situations (e.g., trading \cite{deng2016deep}, self-driving cars \cite{NEURIPS2021_0d5bd023}, or robotics \cite{kober2013reinforcement}) remains a challenge, and complex algorithms have been introduced to bridge the gap between controlled environments and real-life situations 
\cite{arndt2020meta,higgins2017darla}. Hence the applicability of RL models can suffer from lack of robustness\cite{ma2018improved} and interpretability \cite{mott2019towards,} of agent behaviour, and from the large amount of computational resources required. 

The deep incremental learning (DIL) framework introduced here is a hybrid approach which allows predictions from base learner models to be reused in future predictions for tasks on data streams. 
Unlike RL,  
%which allows flow of information between agent and environment in both directions, 
deep incremental learning only allows a single direction of information flow, from one layer to the next. 
%like a waterfall. 
Importantly, the \textit{point-in-time} nature of predictions is preserved so that no look-ahead bias is introduced. DIL can be thought of as an extension of model stacking \cite{naimi2018stacked,} but taking into account the stream nature of the data. 
Here, we consider an incremental problem in finance, which consists of ranking stocks for neutral portfolio optimisation applied to obfuscated data streams of tabular features and targets, corresponding to stocks and computed features. Such data sets are affected by strong non-stationarity and distribution shifts caused by regime changes in the market.  Here, we expand on our previous work \cite{wong2023online} and develop an IL framework that uses  different data and feature sampling schemes and deep incremental model ensemble techniques appropriate for data streams with a high level of concept drift and non-stationarity.

\begin{figure}[htb!]
    \centering
    \includegraphics[width=\textwidth]{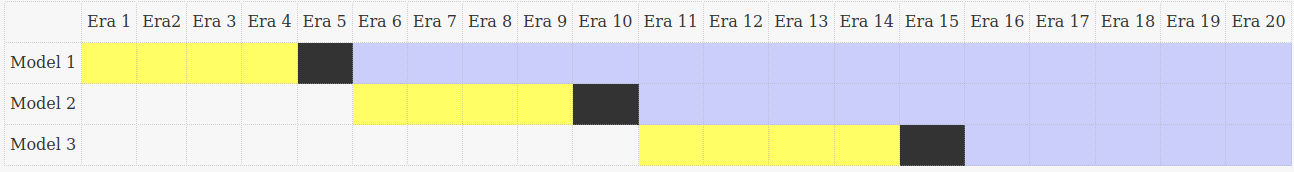}
    \caption{Schematic of how model predictions are reused in an incremental learning model. Consider three models (Models 1-3) each trained over a training period of 4 eras (weeks) and with a lag of 1 era (week). Model 1 is trained using information (both features and targets)  up to Week 4 and after the 1-week lag, predictions are obtained for era 6 onwards. The features from Weeks 6-9 are combined with predictions from Model 1 to train Model 2. Similarly, Model 3 is trained using data from eras 11 to 14, plus predictions from Models 1 and 2. }
    \label{fig:gantt}
\end{figure}

%\paragraph{How IL is different from traditional machine learning} 
Adopting an IL approach is crucial for data streams, as traditional assumptions of machine learning algorithms are not applicable. For instance, 
single-pass cross-validation that splits the data into \textit{fixed} training, validation and test periods is not suitable. 
%Instead, an IL framework should be used to retrain and/or update model parameters. 
Under an IL framework, a model is represented by a continuous stream of parameters. Further, the procedure adds new hyperparameters to the model, such as training size and retrain period, which
%In practice, these hyperparameters could be selected based on computational resources available rather than optimised. The size of memory will limit the maximum training size and the amount of GPU or other processing units available will limit how often the models are retrained/updated. 
%When retrain period is greater than 1, there is an extra degree of freedom in when to start the IL procedure. For example, if models are retrained every 10th era, then starting the IL procedure at Era 1, Era 2, ... to Era 10 would give 10 different training procedures using different training sets. This choice could 
 have a non-negligible impact on prediction performances for non-stationary datasets \cite{hoffstein2020rebalance}.
The distinction between features, targets and predictions is also blurred in the IL setting, as predictions from models learnt from different spans of data  can be used as additional features when building other models, and targets can be created by subtracting against the predictions made. Therefore, model training is a \textit{multi-step} problem, rather than a \text{single-step} problem. For an illustration of these issues, see Fig.~\ref{fig:gantt}.

Reusing model predictions within an IL framework provides a natural hierarchical structure, in which successive models can be interpreted as an improvement of previous ones in response to distributional shifts in data---this is akin to a feedback learning loop where model predictions correct themselves incrementally. Importantly, the prediction quality of each of the models can be inspected independently. Further, the IL setting allows models to process data streams with a finite memory usage by fixing the number of previous models that can be used by a model, so that the size of the training set (consisting of new features and predictions of previous models) remains bounded. 
%Information from previous eras can be passed indirectly to the latest machine learning models. 
There are many other possibilities for the design of IL models to deal with concept drifts in data. See \cite{Gama14,Jie19,} for a survey on recent methods in modelling data streams with different change detection and adaptive learning techniques.
 
Our framework applies this hierarchical IL setting to a \textit{collection} of machine learning models in parallel, which can be thought of as layers of models. 
 %The notion is borrowed from MLP, where each node within a layer is now an independently trained model. 
 %
However, in contrast to standard neural network architectures, such as the multilayer perceptron (MLP), the training is done in a single forward pass without back-propagation. This approach allows us to train complex model with reduced computational resources, as there is no need to put the whole model in distributed memory to pass gradients between layers. In this way, each model within a layer can be trained independently, and training becomes parallelised across GPUs without the need for specialised software packages to distribute data between GPUs.
Recent work in deep learning suggests that backpropagation is not strictly necessary for model training \cite{Hinton22}. For instance, Deep Regression Ensemble (DRE) \cite{Kelly22deep} is built by training layers of ridge regression models with random feature projections.
%similar to an MLP where some layers have frozen weights. 
The deep incremental model presented here, on the other hand, focuses on data streams and temporal data and imposes no restrictions on the ML models used as building blocks forming the layers.

\section{Temporal data formulations}

Our work deals with prediction tasks motivated by financial temporal data streams, whereby the ranking of a group of stocks needs to be predicted based on the information available at era $i$. Such temporal data streams are treated under different formulations. 

\subsection{Temporal Tabular Datasets}
Our temporal data is compiled into temporal tabular datasets, whereby the data at each time point is represented by features that have been computed from the time series up to that time. 

\begin{definition*}[Temporal Tabular Dataset]
%\paragraph{Temporal Tabular Datasets:}
A temporal tabular dataset is a set of matrices $\{ X_i, y_i \}_{1 \leq i \leq T}$ collected over time eras 1 to $T$. 
Each matrix $X_i$ represents data available at era $i$ with dimension $N_i \times M$, where $N_i$ is the number of samples in era $i$ and $M$ is the number of features describing the samples. 
The $y_i$ are the targets to be predicted from the features $X_i$, and can be single-dimensional or multi-dimensional.
The definition of the features is fixed throughout the eras, in the sense that the same computation is used to obtain the same number of features $M$ at each era. 
%In practical applications, the number of data samples in each era is assumed to be bounded. 
%
Although the features can be in different formats (i.e., numerical, ordinal or categorical), they are usually transformed into equal-sized or Gaussian-binned numerical (ordinal) values. %The datasets in this study are all standardised into discrete bins.
Note that the number of data samples $N_i$ does not have to be constant across time. 
%% Data Lag
%\paragraph{Data lag:} Unlike standard online learning problems, where newly arrived data are used immediately to generate predictions and to update the models, 
%
\end{definition*}

\begin{remark*}[Data Lag]
Unlike standard online learning problems, where newly arrived data are used immediately to generate predictions and to update the models, in financial applications there is usually a fixed time lag for the targets from an era to become known (also known as \textit{data embargo}). If the data embargo is, e.g., equal to $5$ eras, the targets of era $t$ become known at era $t+5$, and only then can they be used to calculate the quality of predictions according to a suitably chosen metric. 
%For many applications, the temporal tabular dataset can grow to \textit{infinite} size. 
\end{remark*}

\subsection{Time Series Data} 
In contrast, many traditional methods use time series directly to infer models for prediction. 

\begin{definition*}[Multivariate time series]
%\paragraph{Multivariate time series}
A multivariate time series of $T$ steps and $N$ channels can be represented as a matrix $\mathcal{X}_T = (\bm{x}_1, \bm{x}_2, \dots, \bm{x}_i, \dots, \bm{x}_T)  \in \mathbb{R}^{N \times T} $, where $1 \leq i \leq T$ and each (column) vector $\bm{x}_i \in \mathbb{R}^N$ contains the values of the $N$ channels at time $i$. In many applications, the number of channels $N$ is assumed to be fixed throughout time, with regular and synchronous sampling, i.e. the values in each vector from the $N$ channels arrive simultaneously at a fixed frequency. 
\end{definition*}

Although here we will concentrate on methods to predict temporal tabular datasets, there is a large variety of time series models that predict the time series directly.

\begin{definition*}[Time Series Model] 
%\paragraph{Time Series Model:}
Given a time series $ \mathcal{X}_T \in \mathbb{R^{N\times T}}$,  
%where $T$ is the size of the time dimension and $M$ is the number of features. 
a (one-step ahead) time series model is a function $f: \mathbb{R^{N\times T}} \mapsto \mathbb{R}^N$ that predicts the vector $\bm{x}_{T+1}$ from $X_T$. 
In practice, the function $f$ is often learned by training statistical or ML models using different instances of $\mathcal{X}_T$ obtained by shifting $T$ across the time dimension. 
\end{definition*}

A simple example of such a model, which will be used below, is the Exponential Moving Average (EMA).  Moving averages are commonly used to capture trends in time series as follows.

%\subsubsection{Exponential Moving Average (EMA)}
%\label{section:stats}

%Simple statistical rules can be applied on each feature time series \textit{independently} to summarise the history of the time series. 
 
%

\begin{definition*}[Exponential Moving Average]
Given a univariate time series $x_1,x_2,\dots,x_t,\dots$, the exponential moving average of the time series at time $t$ with decay $\alpha$ is defined as 
\begin{align}
\label{equation:EMA}
    y_t &= (1-\alpha) y_{t-1} + \alpha x_t
\end{align}
with initialisation $y_1 = x_1$.
\end{definition*}
%\end{color}

\begin{remark*}
    More complex time series models have been developed, including sequence models in deep learning, such as LSTM \cite{HochSchm97} and Transformers \cite{Bryan19}.  However, these models tend to be overparameterised and lack robustness to regime changes \cite{Shereen21}. They also involve heavy computational costs associated with the training and updating of models.
\end{remark*}

\subsection{Transforming time series into temporal tabular datasets: feature extraction} 
\label{section:NumeraiSunshine-feature-eng-multi}

There are a myriad of methods commonly used to transform multivariate time series into temporal tabular datasets. These feature engineering (FE) methods consist of feature extraction applied over a look-back window:

\begin{itemize}
\item Feature extraction: 
a function $f$ that maps the time series $\mathcal{X}_T \in \mathbb{R}^{N \times T}$ to a feature space $f(\mathcal{X}_T) \in \mathbb{R}^M$ where $M$ is the number of features. Feature extraction methods can help reduce the dimension and noise in time series data.

\item Look-back window:
%As time series can have unbounded length, 
Feature extraction is applied to data within a look-back window (memory) of fixed length $k$. 
%is used to restrict the data size when calculating features. %To avoid look-ahead bias, features are computed using only preceding values. % that represent the state of time series at time $i$ can only be calculated using values obtained up to time $i$, which is $(\bm{x}_1, \bm{x}_2, \dots, \bm{x}_i)$. For most IL tasks, 
%A usual assumption is  data collected more recently are more important than data collected from the past. Therefore, analysis is often restricted to use the most recent $k$ data points only, which are $(\bm{x}_{i-k}, \bm{x}_{i+1-k}, \dots, \bm{x}_i)$. This represents the state of the time series at time $i$ with a look-back window of size $k$. Feature extraction methods are applied to data within the look-back window only. 
Multiple look-back windows can also be used to extract features that capture short-term and long-term trends, and concatenated to represent the state of the time series.  
\end{itemize}

%% Look-back windows

%To model multivariate time series effectively, better methods which can be applied on multiple time series in parallel are needed. In this study, both deterministic and random transformations are considered to create tabular features for the prediction task. 

In this paper, we will employ two feature engineering methods that have been proposed for financial time series:
\begin{itemize}
    \item \textit{Signature Transform (ST)}:
% Signatures
%\subsubsection{Signature Transforms} 
STs \cite{Lyons07, Chevyrev16, Terry22} are deterministic transformations, recently proposed by Lyons, which can extract features at increasing orders of complexity from multivariate data, including time series.  See~\cite{Terry22} for a review of different applications of signature transforms in machine learning. For details on how STs are applied to the Numerai dataset, see Section \ref{section:NumeraiSunshine-tsalgos} in the Supplementary Information.

\item \textit{Random Fourier Transform (RFT)}:
%\subsubsection{Random Fourier Transforms} 
RFTs have been used in \cite{kelly2022virtue} to model the return of financial price time series but can also be applied to extract features from time series at each time step. The key idea is to approximate a mixture model of Gaussian kernels with trigonometric functions \cite{Sutherland15}.  Details on how RFTs are applied on the Numerai dataset are given in the Supplementary Information, see Algorithm~\ref{alg:rft} in Section \ref{section:NumeraiSunshine-tsalgos}.
\end{itemize}

\begin{remark*}
As discussed in Section~\ref{section:NumeraiSunshine-tabular-inc} in more detail, once feature extraction methods have been applied and temporal tabular datasets generated, traditional ML models such as ridge regression, gradient-boosting decision trees (GBDTs), and multi-layer perceptron (MLP) networks can be used to carry out predictions point-wise in time~\cite{wong2023online}, without relying on complex and expensive advanced neural network architectures such as Recurrent Neural Networks (RNN), Long-Short-Term-Memory (LSTM) Networks or Transformers \cite{Bryan19}.     
\end{remark*}

\section{Machine learning for temporal data}
\label{section:NumeraiSunshine-ML-TS}

Before describing our deep incremental learning approach, we give some relevant background and brief links to standard methods used for prediction of temporal tabular data. These methods will be used as the building blocks of our incremental learning approach. 

\subsection{Prediction of Temporal Tabular Datasets from time series data:  Factor-timing models }

Factor-timing models~\cite{haddad2020factor,} are a well-used approach to produce predictions for a temporal tabular dataset from time series, whereby
the raw predicted values from a time series model (e.g., the EMA~\eqref{equation:EMA}) are converted into normalised rankings, which are then used as weights for the linear factor-timing model (see Algorithm \ref{alg:factor-timing}).  
As baseline for comparison, we apply below factor-timing models to time series that are derived from temporal tabular datasets through a transformation, as follows.
\begin{definition*}[Derived Time Series]
 \label{def:derivedts}
A transformation $f$ is applied to the tabular features $X_t$  and targets $y_t$ at era $t$  to generate a multivariate time series: 
$\bm{\chi}_t = f(X_t,y_t)$, 
where $f:(\mathbb{R}^{N_i \times M}, \mathbb{R}^{N_i \times 1}) \mapsto \mathbb{R}^{M}$. For example, $f$ can be the Pearson correlation between feature and targets.
%where $X$ and $y$ are the features and targets, $N_i$ is the number of observations at era $i$, $M$ is the number of features. 
\end{definition*}  
This procedure generates a time series of  \textit{feature performances} from the temporal tabular dataset, which can be used within a factor-timing model, as in Algorithm~\ref{alg:factor-timing} avoiding look-ahead bias.   
%To avoid look-ahead bias, we restrict to transformations that can be applied to each era independently. 

%A major limitation of this assumption is that it precludes the use of feature engineering methods such as Auto-Encoders which are applied across eras. However this assumption can make sure there is no look-ahead bias in the model as these derived time series would be used to build time series models. 

%There are two approaches to model temporal tabular datasets. The first is to apply standard machine learning algorithms for tabular datasets as usual, taking into account the temporal order of data during data sampling. 
%The second is to convert the temporal tabular datasets into the derived time-series, defined as \ref{def:derivedts} and then apply different time-series prediction algorithms followed by factor-timing models. 
%In particular, we will apply different feature engineering methods, such as Signature transforms \cite{Terry22} and Random Fourier transforms \cite{Sutherland15} on the derived time-series to create tabular features for performing time-series forecast (which are different from the original temporal tabular dataset given). 

\begin{algorithm}[hbt!]
\caption{Factor Timing Model}
\label{alg:factor-timing}
\KwIn{At era $t$: predicted values $\hat{y}_t \in \mathbb{R}^M$ from a time series model, and temporal tabular dataset $X_t \in \mathbb{R}^{N_t \times M}$ where $M$ is the number of features}
\KwOut{Factor-timing model predictions $\hat{z}_t \in \mathbb{R}^{N_t}$ }
Calculate normalised ranking of features $\hat{r}_t$ from predictions of time series model 
\[
    \hat{r}_t = \text{rank}(\hat{y}_t) - 0.5 ,
\]
where the rank function calculates the percentile rank of a value within a vector, so that $-0.5 \leq \hat{r}_t \leq 0.5$. \\
If needed, given upper bound $u$ and lower bound $l$ $-0.5 < l < u <0.5$, apply truncation to $\hat{r}_t$:
\begin{equation*}
    r_t = \max(\min(\hat{r}_t,u),l).
\end{equation*} 
Calculate linear factor-timing predictions $\hat{z}_t = X_t r_t $ 
\end{algorithm}

\subsection{Machine Learning Models for Temporal Tabular Dataset prediction} 
\label{section:NumeraiSunshine-tabular-inc}

In contrast to factor-timing models, ML methods can be applied directly to temporal tabular datasets for prediction tasks.
There is a rich literature comparing different machine learning approaches on tabular datasets \cite{mcelfresh2023neural,Shwartz21,Leo22,Arlind21,}. 
Several benchmarking studies \cite{mcelfresh2023neural,Shwartz21,Leo22,} have demonstrated that advanced deep learning methods, such as transformers \cite{Arik_Pfister_2021,} and other neural network (NN) models , underperform for  regression/classification tasks on tabular datasets relative to traditional approaches, such as GBDT or MLP models. In particular, recent research \cite{mcelfresh2023neural,} has shown that GBDTs with moderate hyperparameter tuning perform closely to much more complex NN models. 

Further, previous studies had focused on datasets with relatively small numbers of features and samples ($M < 200$ features,  $N_i < 10,000$ data rows or samples), whereas we are interested in large datasets with more than 1000 features and more than 200,000 data rows. For larger datasets, it has been shown~\cite{mcelfresh2023neural,} that GBDTs performed better than 11 neural-network-based approaches and 5 other baseline approaches, such as Support Vector Machines. GBDT models also display higher performance when feature distributions are skewed or heavy-tailed. 

Finally, our objective is the prediction of data streams that are not static or stationary, but rather dynamic and subject to distribution shifts.  Previous work has shown that GBDTs and MLPs outperform other deep learning approaches for temporal tabular datasets, with higher robustness and lower computational requirements for training (and retraining) of models~\cite{wong2023online,Shwartz21,Leo22}. 

In this paper, GBDT models are studied in detail for tabular prediction, as it has demonstrated strong performances in benchmarking studies \cite{mcelfresh2023neural,Shwartz21,Leo22, wong2023online} and there exist efficient implementations that allow scalable model training and inference. 

Details of the GBDT and MLP models can be found in section \label{section:NumeraiSunshine-XGBoost} and \label{section:NumeraiSunshine-DL} in the Supplementary Information.

\section{Deep (hierarchical) Incremental Learning algorithm for temporal data}

Our deep incremental learning model is built layer by layer, using component models of a given type (e.g., factor-timing models or GBDTs) composed hierarchically across layers, as follows.
At any time, we split our temporal dataset into segments of temporal history. Each segment is assigned to a layer, and for each layer we train an ensemble of models computed with different random seeds.  We thus define the number of layers $L$,  the sizes of the training data (`lookback window') for each layer $(a_1, \dots, a_L)$, and the number of models in the ensemble within each layer $(K_1, \dots, K_L)$. 

The models are learnt using information from different temporal segments sequentially and hierarchically, layer by layer, so that past predictions can be used to refine future predictions.
Operationally, at the start of the training in a layer $l$, we prepare the features and targets $\{ X_j^l, y_j^l \}$ that are shared by all $K_l$ component models within the layer using the most recent data from the specified lookback window. Importantly, the features used as inputs to a layer consist of both original features from the temporal tabular dataset plus predictions obtained from models trained in previous layers $i=1, \ldots, l-1$ (see Fig.~\ref{fig:gantt}).

Regarding the type of component models that form the ensemble in each layer, any model that uses tabular features as input and predicts tabular targets can be used. This expands the class of models from standard tabular models, such as GBDT and MLP, to other multi-step models, such as factor-timing models. 
The overall model is therefore a composition of such component models.
%each of which starts with tabular features and ends with tabular targets.

Each component model within a layer is trained in an incremental manner. This means that the model parameters are updated at regular intervals as new data arrives only using the data from the given lookback window. Other hyperparameters of the model (e.g., boosting rounds for GBDTs) remain unchanged. 
For example, if the dataset in total has 1000 eras and we update the models every 50 eras with lookback window equal to 600 eras we would obtain 9 models, with model training at Eras 600,650,700, \dots, 1000.

The component models within each layer can be trained in parallel, which allows the incremental learning model to be efficient and scalable.

The pseudocode in Algorithm \ref{alg:incremental-stack} outlines the overall structure of the computational framework.

\begin{algorithm}[H]
\caption{Deep IL model with model stacking}
\label{alg:incremental-stack}

\KwIn{Temporal Tabular Dataset $\{ X_i, y_i \}_{1 \leq i \leq T}$, number of layers $L$, the number of models within each layer $(K_1, \dots, K_L)$, sizes of training data in each layer $(a_1, \dots, a_L)$, data embargo $b$, }

\For{$1 \leq l \leq L$}{
    Assign the temporal window $w_l$ to layer $l$: $w_l=\{\sum_{w=1}^l (a_w+b) < j \leq \sum_{W=1}^{l-1} (a_w+b) + a_l \} $ \\ 
    Prepare training data for layer $l$: $\{ X_j^l, y_j^l \}_{j \in w_l}$, where the features $X_j^l$ can be any combinations of predictions from previous layers and the original features in the temporal tabular dataset. \\
    \For{$1 \leq k \leq K_L$}{
    Perform data and feature sub-sampling for each component model $\mathcal{M}_k^l$ \\
    Train component model $\mathcal{M}_k^l$ with regular updates\\
    Obtain predicted ranking of stocks using $\mathcal{M}_k^l$ from era $1+\sum_{w=1}^l (a_w+b)$ onward to be used in model training in subsequent layers \\
    }
}
\end{algorithm}

This framework leads to a deep hierarchical ensemble of models, where each layer takes advantage of model ensembling, and the integration of information across layers through functional composition enables the incremental learning necessary to adapt to non-stationarity and regime changes.
We now discuss briefly some characteristics of the model:

\paragraph{Hierarchical nature of the model and self-similarity: }

The proposed framework is hierarchical: the ensemble of models in any given layer, which is used to generate predictions in time beyond the latest data arrival, integrates hierarchically both data and predictions obtained from the models in the preceding layers, themselves fitted to previous time periods. Indeed, the model has characteristics of self-similarity, since the layered structure can be seen as performing a functional composition of learning models of the same type, e.g., the component models within each layer can be chosen to be GBDTs (or MLPs) so that the learning mechanism of each individual component is similar to the overall model, and the structure is extended repeatedly in a self-similar manner by interpreting a component model as a base learner for another component model in a higher layer. 
%
%The IL model demonstrates a hierarchical nature in modelling. In some applications, the model also shows self-similarity with a fractal-like structure. In other words, the overall model shared a similar structure with each of its components. 

\paragraph{Universal Approximation Property:} 

It is well known that MLP and GBDT models have the universal function approximation property \cite{cybenko1989approximation}, and Deep Learning models for sequences, such as LSTM \cite{schafer2006recurrent}, also have the universal function approximation property for any dynamical system. Since the DIL model is a composition of models each of which has the universal function approximation property, it also has the universal approximation property for the underlying stochastic process that drives the data generation of the temporal tabular dataset.

\paragraph{Model stacking: bagging and boosting across time} 
Our model can also be interpreted as a \textit{stacked model} with a total of $\sum_{i=1}^L K_i$ base learners,  such that $K_i$ base learners are trained in the $i$-th iteration, corresponding to each of the $L$ layers. Ideas from bagging and boosting are integrated within the model. Each layer consists of multiple models trained in parallel, as in bagging, so that variance is reduced by combining predictions from different models within a layer. Further, our model can be considered as a degenerate case of boosting, where the learning rate of the target is set to zero inter-layers, such that the target is not adjusted based on predictions from previous layers. However, the  architecture can be modified to allow for target adjustment (boosting) between layers if needed.

\paragraph{Adaptive nature of the model} 
A key characteristic of the DIL model is that it is designed to support \textit{dynamic} model training, with parameters of each component model updated regularly to adapt to distributional shifts in data.
Under the traditional machine learning framework, hyperparameters are selected by cross-validation on splits of the training data. Yet optimal hyperparameters based on a single test period might not work in future. In the DIL model,  
predictions from previous layers based on different model hyperparameters are combined in the successive layer, corresponding to a later span of time, acting as a dynamic soft selection of hyperparameters. 
%Soft hyperparameter selection is also related to Bayesian methods of learning the regularisation hyperparameter of regression models. Instead of attempting to derive the posterior distributions of the model hyperparameters, which is difficult when there are no closed-form solutions, the weight parameters can be considered as an approximation to the posterior distribution. 
It has been shown that stacking of models with different random seeds \cite{lakshminarayanan2017simple}, hyperparameters \cite{Florian20} and architectures \cite{zaidi2021neural} leads to robust performance for \textit{static} datasets.
The DIL model can thus be seen as an extension of stacking techniques to \textit{stream} datasets, so that models incrementally trained with incoming data streams are stacked to obtain more robust predictions.

\section{Prediction tasks for neutral portfolio optimisation using financial data from the Numerai competition}
\label{section:NumeraiRain-predtask}

\paragraph{Numerai dataset and prediction task} 
\label{section:numerai-sunshine-dataset}

%%%%%% Things added here to this intro

As discussed above, financial time series data can be used directly for prediction~\cite{percival_walden_2020, Bryan19}, yet such methods tend to be overfitted, making them less robust to regime changes and to the high stochasticity inherent to financial data. Alternatively, feature engineering is applied at each era to compute features that capture different aspects of the time history over look-back periods. This approach leads to a temporal tabular dataset, which can be used for prediction without considering time explicitly. The Numerai competition is based on one such professionally curated temporal tabular dataset, formed by matrices $X_i$ that contain $M$ stock market features (computed by Numerai) for $N_i$ stocks updated weekly (i.e., eras are weeks). The definition and computation of the features is fixed throughout the eras. Importantly, the dataset is \textit{obfuscated}, 
%so that the proprietary data generation process from financial datasets is not known to the participants. It is 
i.e., the identity of the stocks present each week is unknown.  The task is then to predict the stock rankings each week, from lowest to highest expected return. This ranking is used to construct a market-neutral portfolio.

\paragraph{Features and Targets}

Two versions of the Numerai dataset, V4.1 (Sunshine) and V4.2 (Rain) \cite{numerai-datav4.1,numerai-datav4.2} are used in this study, starting on 2003-01-03 (Era 1) and extending up to 2023-06-30 (Era 1070) \footnote{The data keeps updating every week}. The dataset is weekly, i.e., eras correspond to weeks.

\begin{figure}
    \centering
    \includegraphics[width=7.5cm]{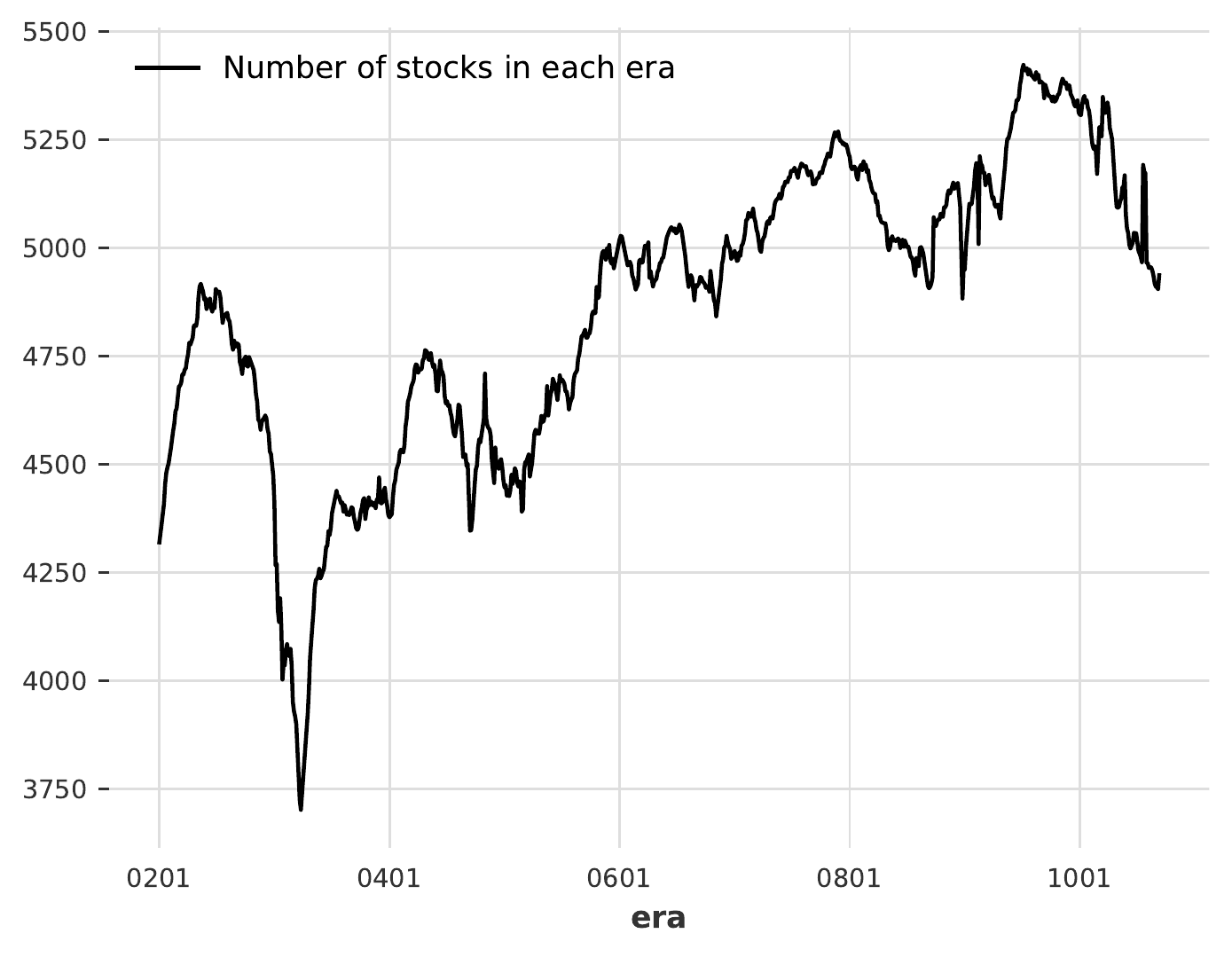}
    \caption{Number of stocks in each era from Era 201 to Era 1070 for v4.2 Numerai dataset.}
    \label{fig:Rain-EraSize}
\end{figure}

Each week, Numerai makes public a feature matrix of 1586 (V4.1)/ 2131 (V4.2) features for a changing selection of (unidentified) stocks, selected according to risk management rules by the Numerai hedge fund, plus several targets corresponding to stock returns normalised by different proprietary statistical methods. In Figure \ref{fig:Rain-EraSize}, the number of stocks in each week (era) from Era 201 to Era 1070 are shown, which demonstrates the number of stocks traded varied in each week. 

The features are normalised into 5 equal-sized integer bins, from -2 to 2, so that the bins have zero mean. The targets are scaled between 0 and 1, and grouped into 5 bins (0, 0.25, 0.5, 0.75, 1.0) following a Gaussian-like distribution, and then subtracting 0.5 to make the bins zero-mean. For a more extended discussion of the Numerai dataset, including features and targets, see Ref.~\cite{wong2023online}.

\begin{figure}
    \centering
    \includegraphics[width=7.5cm]{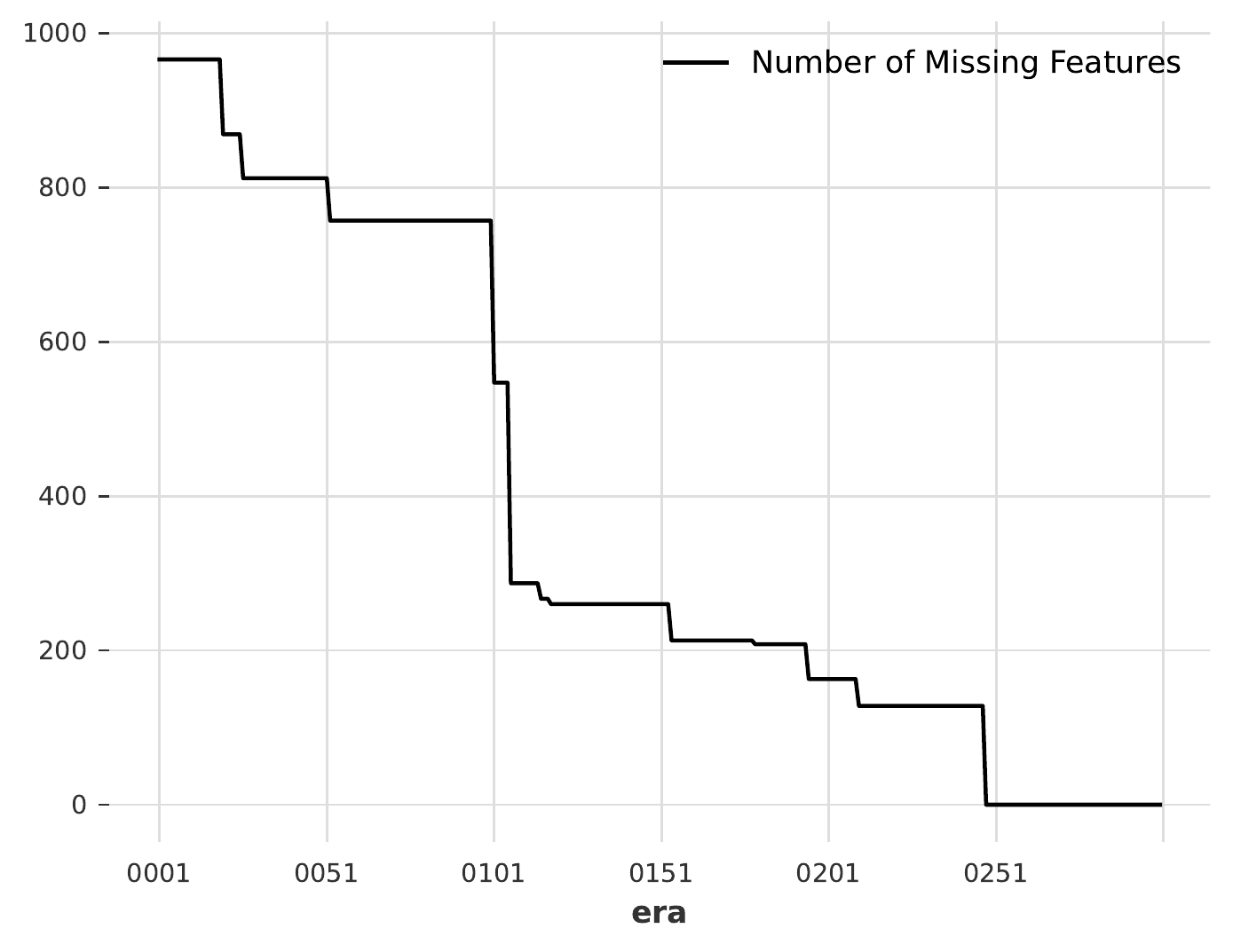}
    \caption{Number of features with missing values from Era 1 to Era 300 for v4.2 Numerai dataset.}
    \label{fig:Rain-MissingFeat}
\end{figure}

%% Changes to Dataset 
In V4.2 dataset, some features have completely missing values up to Era 251. In Figure \ref{fig:Rain-MissingFeat}, we show the number of features with completely missing values in each era between Era 1 and Era 300. In the first 100 eras, we have around $50\%$ of features with completely missing values. Therefore, we train XGBoost models using data from Era 201 onwards to ensure less than $10\%$ of features have completely missing values. 

%% Feature Categories
Each feature is now assigned to one or more groups. There are 10 feature groups in total, namely Intelligence, Charisma, Strength, Dexterity, Constitution, Wisdom, Agility, Serenity, Sunshine and Rain. For all feature groups except the last one (Rain), they represent features that behave similarly, as they are derived from similar data sources \cite{numerai-datav4.2,}. The Rain feature group consists of features that are created synthetically from features in other groups using information up to Era 585 \footnote{Numerai suggests most features are derived by fitting weights to the time-series of other features.}.

\paragraph{Data Lag}

The data lag for predictions depends on the practicalities of the data pipeline. For Numerai, a lower bound for the scoring target to be resolved is 5 weeks (4 weeks of market data and 1 week for data processing). To take account into both the data lag for the data generation process from Numerai, and the time needed to train models, a conservative data lag of 15 weeks is used here. 

\paragraph{Scoring Function} 
\label{formula:Numerai-Corr}

Numerai calculates a variant of Pearson correlation for all predictions in a single era $t$, as follows~\cite{numerai-corr}:
Let $y_p$ be the predictions ranked between 0 and 1, $y_t$ the targets centred between -0.5 and 0.5, $\Phi(\cdot)$ the (cumulative) distribution function of a standard Gaussian, $\textbf{sgn}(\cdot)$ and $\textbf{abs}(\cdot)$ the element-wise sign and absolute value function, respectively, then the Numerai correlation score for era $t$, $\rho_t$, is given by: 
\begin{align*}
    y_g &= \Phi^{-1}(y_p) \\
    y_{g15} &= \textbf{sgn}(y_g) \cdot \textbf{abs}(y_g)^{1.5} \\
    y_{t15} &= \textbf{sgn}(y_t) \cdot \textbf{abs}(y_t)^{1.5} \\
    \rho_t &= \textbf{Corr}(y_{g15},y_{t15}) \, 
\end{align*}
where $\textbf{Corr}(\cdot,\cdot)$ is the Pearson correlation function. Note that the 3/2 power is taken to emphasise the contribution from the highest and lowest predictions. 
The correlation score $\rho_t$ is collected for each era $t$ over the test period to calculate the following portfolio metrics: 
\begin{itemize}
    \item Mean Corr: average of $\rho_t$ over all eras in the test period
    \item Maximum Drawdown: maximum difference between the cumulative peak (high watermark) and the cumulative sum of correlation scores in the test period 
    \item Sharpe ratio: ratio of Mean corr and standard deviation of $\rho_t$ over all eras in the test period
    \item Calmar ratio: ratio of Mean Corr and Maximum Drawdown 
\end{itemize}
We will use these metrics to score our models throughout the paper. Specifically, high values of `Mean Corr', `Sharpe ratio' and `Calmar ratio' are all indicative of good model performance. We use the main target decided by Numerai, 'target-cyrus-v4-20' for scoring the trained models.

\paragraph{Example of concept drift}

The presence of regime changes is one of the reasons why machine learning trading strategies suffer from significant losses. Machine learning trading strategies learn historical patterns from a vast amount of financial data. When there are regime changes, these patterns become obsolete, or even incorrect, such that they are no longer able to predict the future return of financial assets. 

Regime changes are often unpredictable. For example, considering the return from Numerai hedge fund \cite{numerai-fund,}, the risk-adjusted return of hedge fund from September 2019 up to March 2023 is spectacular, where the maximum drawdown is less than $5\%$. However, from March 2023 there are 4 consecutive months of negative returns, giving a cumulative drawdown of more than $33\%$. Indeed, most risk-management metrics based on historical performances, such as Value-at-Risk (VaR)~\cite{Prado2015,} would not be able to foresee this downturn. 

\begin{figure}
    \centering
    \includegraphics[height=7.5cm]{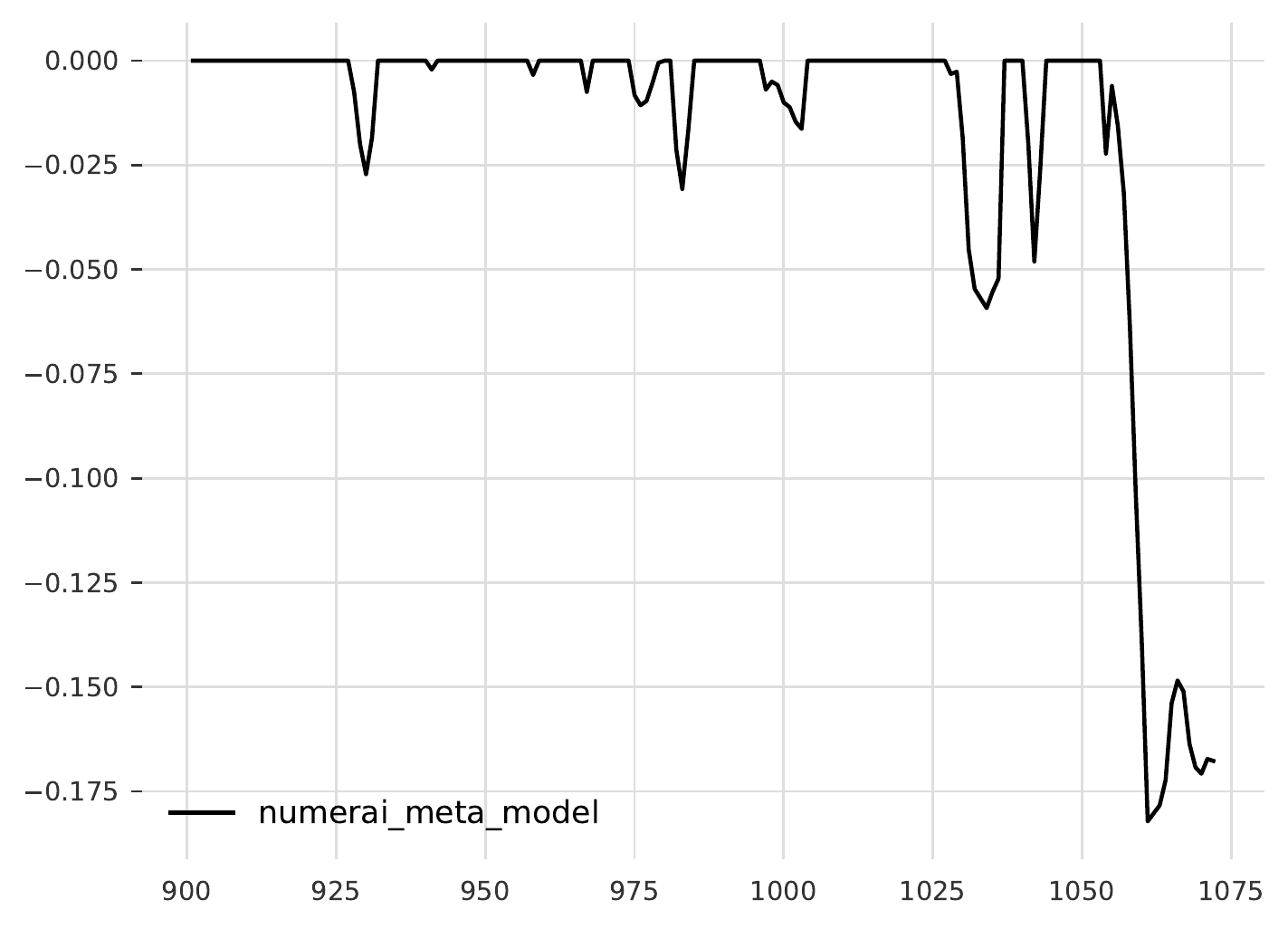}
    \caption{Underwater (Drawdown) plot of the Numerai Meta Model between Era 901 to Era 1070. A large drawdown is experienced by the model between Era 1055 to Era 1070. }
    \label{fig:Rain-Underwater}
\end{figure}

The challenging period for Numerai hedge fund corresponds to Era 1055 to Era 1070 in the dataset. Similar to the hedge fund, predictions from models submitted by participants in the competition also suffered from a large drawdown in the same period. In Figure \ref{fig:Rain-Underwater}, we show the Underwater (Drawdown) plot of the Numerai Meta Model. The drawdown between Era 1055--Era 1070 is around 4 times bigger than historical drawdown, suggesting there could be concept drift in the data. 

%% Performance of Models in Different Regimes 
Here, we define market regime \textbf{post hoc} based on performances. 2023-02-17 (Era 1051) to 2023-06-30 (Era 1070) is defined as the bear market. 2020-04-04 (Era 901) to 2023-02-10 (Era 1050) is defined as the bull market. In Table \ref{table:Rain-ExamplePerformances}, we report the performances of the Numerai Meta Model from Era 901 to Era 1070 for the whole period and under both market regimes. Under bull market, we have a better than average performance while under bear market we have a negative performance. Over a long enough period, model predictions have a positive return but models can experience large drawdown in bear market, causing a lot of volatility to the portfolio.

\begin{table}[hbt!]
\centering
\begin{tabular}{|l|l|l|l|}
\hline
Regime  &   Mean Corr  & Sharpe  & Calmar  \\ \hline
All     &   0.0175     & 0.7915  & 0.0962      \\ \hline
Bull (Eras 901-1050) &   0.0207     & 1.0085  & 0.3491      \\ \hline
Bear (Eras 1051-1070)   &  -0.0062     & -0.3220 & -0.0341     \\ \hline
\end{tabular}
\caption{Performances of Numerai Meta Model from Era 901 to Era 1070 under different market regimes. }
\label{table:Rain-ExamplePerformances}
\end{table}

\section{Incremental Learning for Numerai prediction: Non-hierarchical models}
\label{section:NumeraiRain-IL}

%Here we present the results of training IL models made up from component models of different type for prediction tasks in the Numerai dataset. 

Before presenting results from our hierarchical (deep) incremental learning model, we develop non-hierarchical incremental learning models for the Numerai dataset. These types of models have already been used in the literature \cite{kelly2022virtue,kelly_malamud_2021,zhao2023doubleadapt,} and will serve here both as a baseline comparison and to guide some our choices in model type, training methods and hyperparameter selection. We note that although these models are updated incrementally (i.e., they do incorporate information of new data arrivals) they do not incorporate information hierarchically across multiple layers, and hence fail to generalise well, due to severe distribution shifts in the data.   

To enhance the breadth of our comparison, we study here two types of IL models: (i) factor-timing models that use explicit time series derived from the Numerai dataset, and (ii) ML algorithms (GBDTs, MLP) for tabular datasets which are used directly on the Numerai temporal tabular dataset.

\subsection{Factor Timing Models}

We generate three factor-timing (FT) models (based on Exponential Moving Average, Signature Transform, and Random Fourier Transform), all of which follow the setup in Algorithm~\ref{alg:factor-timing} but are generated using specific transformations of the data, as follows.

We obtain a multivariate time series from the V4.2 dataset $\{\tilde{X}_t,y_t\}_{t=1}^{1070}$ as described in Section~\ref{def:derivedts}, i.e., we generate the time series $\{ \bm{\chi}_t\}_{t=1}^{1070}$, where each $\bm{\chi}_t \in \mathbb{R}^{2132}$ is derived by computing the correlation between each feature and the target $y_t$.
Once the time series is computed, we train factor timing models at each era using all the available data up to that point, bar the data embargo of 15 eras.
In particular, we train the following FT models: 
%one using an EMA model of the time series and two other using transforms that generate features from the time series:
\begin{itemize}
\item
Exponential Moving Average factor-timing model: An EMA model~\eqref{equation:EMA} is computed for each of the 2132 feature series independently.  This multivariate model is used to produced predictions $\hat{y}_t \in \mathbb{R}^{2132}$ for each era $t$, which are then used within the FT model to produce model predictions $\hat{z}_t \in \mathbb{R}^{N_t}$, as given by Algorithm~\ref{alg:factor-timing}.  These predictions are then scored using our portfolio metrics. 

\item
Feature Transform factor-timing models (ST and RFT): From a random subset of the 2132 variables of the time series $\bm{\chi}_t$  we generate transformed features (ST or RFT) with lookback period using all available data.
This process is repeated for a varying number of randomly drawn subsets of the variables in $\bm{\chi}_t$ to explore the importance of model complexity $c$,  defined as the ratio of number of features and length of the time series. For example, for a time series of length $T=600$, we may wish to generate models with complexity $c=2$.  Therefore, we obtain $c \cdot T= 1200$ ST features by taking 60 subsets randomly sampled from $\bm{\chi}_t$, where each random subset of 4 time series generates 20 ST features (taking signatures up to level 2). An analogous procedure is followed for RFT.   
The $c \cdot T$ transformed features (ST/RFT) from all subsets are then concatenated, and ridge regression with L2-regularisation is applied to generate the linear model for $\hat{z}_t \in \mathbb{R}^{N_t}$. 

We note that following recent research in high dimensional ridgeless regression \cite{Hastie19,kelly2022virtue,},
we average the results of ridge regression over a range of regularisation parameters (0.01,0.1,1.0,10.0,100.0) that cover a spectrum of models, from dense to sparse. 

\end{itemize}

%\paragraph{Training set and hyperparameter optimisation over evaluation set}
The FT models are retrained at every era using all data available up to that point; hence by construction these models are all \emph{incremental}. In Algorithm \ref{alg:Sunshine-FactorTiming}, we describe the incremental learning procedure to train FT models.

\begin{figure}[hbt!]
     \centering
        \subfloat[EMA models]{
          \includegraphics[width=15cm]{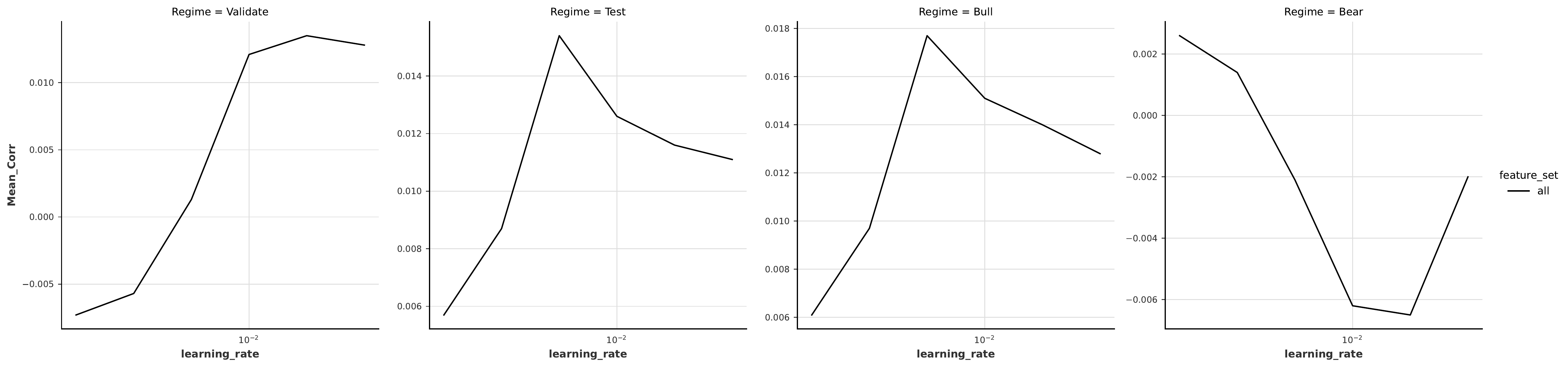}
        }
        \\
        \subfloat[Feature Transform factor-timing models]{
          \includegraphics[width=15cm]{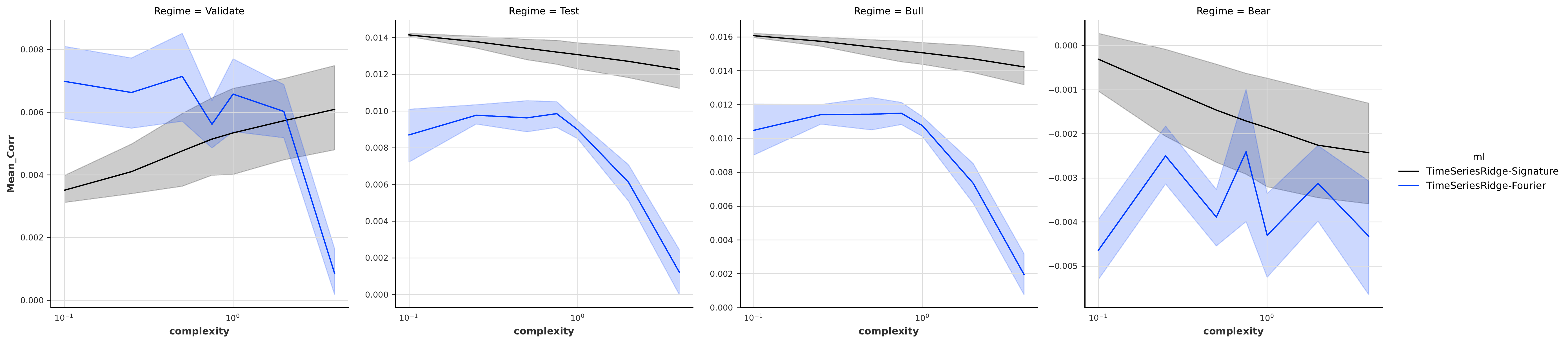}
        }
        \\
    \caption{Mean Corr of EMA and Feature Transform factor-timing models under different market regimes}
    \label{fig:Rain-Trend-Summary}
\end{figure}

\begin{algorithm}[H]
\caption{Factor Timing Models}
\label{alg:Sunshine-FactorTiming}
\KwIn{Data embargo $b=15$}
\For{$401 \leq i \leq 1070$}{
    Calculate feature performances time series $\{ \bm{\chi}_t\}_{t=1}^{1070}$, where each $\bm{\chi}_t \in \mathbb{R}^{2132}$ is obtained using the procedure described in Section~\ref{def:derivedts}
}
\For{$ 801 \leq i \leq 1070$}{
    Prepare training data by slicing the feature performances time series from Era $1$ to $D_i-b$ \\
    Train Factor Timing models (EMA, ST. RFT). \\
    Get one-step ahead prediction $y_{i+1}$ for era $D_{i+1}$ \\
    Create factor timing predictions $z_{i+1}$ using Algorithm \ref{alg:factor-timing} \\
}
\end{algorithm} 
To optimise the key hyperparameters of the FT models (decay $\alpha$ for EMA models, and complexity $c$ for ST/RFT models), we evaluate their one-step ahead performance over the \textit{validation} period from Era 801-Era 885.  
Figure \ref{fig:Rain-Trend-Summary}(a) shows the performances of EMA models with different weight decays under different market regimes for weight decays $\alpha = [ 0.00125,0.0025,0.005,0.01,0.02,0.04]$. Note that the best weight decay for Mean Corr in the validation period is 0.02, whereas it is 0.005 in the test period. This is another example of the effect of regime changes and why time-series cross-validation cannot always select the best weight decay for out-of-sample data. In particular, and as expected, models with a smaller decay constant perform better in Bear market but are the worst in Bull market. 
%suggesting model behaviours vary under different market regimes. 
%
The key hyperparameter for the signture FT models (complexity) is explored in 
Figure \ref{fig:Rain-Trend-Summary}(b) under different market regimes, averaged over 4 different random seeds. 
%Results for other risk metrics, Sharpe ratio and Calmar ratio are shown in Figure \ref{fig:Rain-Trend-Summary-Sharpe} and \ref{fig:Rain-Trend-Summary-Calmar} respectively. 
%
We train Feature Transform factor-timing models with different complexities $c=0.1,0.25,0.5,0.75,1,2,4$. RFT performs better than ST models in the validation period for small model complexity ($c<1$) but not in the test period. For RFT models, Mean Corr decreases as model complexities increases. For ST models, Mean Corr increases as model complexity increases in the validation period but not in the test period. This suggests using more complex factor-timing models do not always give better results for FT models and reinforces the suggestion that hyperparameters selected based on time-series cross-validation might not be robust in out-of-sample data.

\subsection{Benchmark IL model for XGBoost models} 
\label{section:NumeraiRain-benchmark}

Two key hyperparameters for IL models are (i) training size, which for a temporal tabular dataset corresponds to the number of eras of data to be used in training; and (ii) retraining period, which governs how often the model is retrained/updated using the latest data.

In this section, we train IL models with different training sizes and retrain periods, using XGBoost models with two different sets of hyperparameters:
\begin{itemize}
    \item a set found by grid search with fixed number of boosting rounds $B=5000$ and learning rate $L=0.01$, which we denote the Ansatz hyperparameter set
    \item  a set provided by Numerai in their example Python script, denoted here the Numerai hyperparameter set.
\end{itemize}
In each model retrain, we use \textbf{all} the available data from Era 201 to train the models. For example, at Era 1000 which is the $6th$ retrain of model, we use 800 eras of data from Era 201 to train the models. The first retrain at Era 801 uses 600 eras of data, which is roughly equal to the training period of the Numerai example models using the first 12 years of data ($\approx 574$ eras). 

For completeness, and as a reference comparison,  we also create benchmark models that are not regularly retrained, which we call \textbf{Ansatz-Fixed} and \textbf{Numerai-Fixed} respectively.

The details on the procedure to create hyperparameter sets and model training are described in Sections \ref{section:NumeraiRain-benchmark} and \ref{section:NumeraiSunshine-HyperOpt} in the SI. To speed up training, we train models using only around half of data in each era by removing observations with target equal to the Median value (0.5). We show in SI Section \ref{section:NumeraiRain-erasample} that this sampling method does not deteriorate model performance, while reducing computational costs by half compared to training using all data. 

Finally, in order to manage the computational constraints, we produce regular samples of the data eras in the training period such that only $25\%$ of the data eras is used in model training. We then train 4 models each using $25\%$ of data without overlap. For example, we use data from Era 1,5,9,$\dots$ to train the first model, and similarly for the other 3 models. We call this procedure \textbf{regular era sampling}.

We create benchmark models of size $B=1000,5000,10000,25000,50000$. The train size of models are fixed to 600 (with the last 15 eras of data for embargo) with the start of training data at Era 201. For models with $B \leq 5000$, we regularly retrain models every 50th era, which corresponds to updating the model once per year. We do not retrain models with $B \geq 10000$ due to computational limitations. The learning rates of the model are determined using the Ansatz formula $L=\frac{50}{B}$, which is explained in detail in Section \ref{section:NumeraiRain-LRopt} in SI. 

We report performances of the benchmark models from Era 801 to Era 1070 according to the following regimes: 
%The validation period is Era 801 to Era 885. The test period is between Era 901 and Era 1070, where 15 eras of data embargo are used. We also report performance based on market regimes within the test defined to understand if models behave differently under different regimes. 
\begin{itemize}
    \item Validation: Era 801 - Era 885 
    \item Test: Era 901 - Era 1070
    \begin{itemize}
    \item Bull: Era 901 - Era 1050
    \item Bear: Era 1051 - Era 1070
    \end{itemize}
\end{itemize}
%% Figure of Mean Corr 

\begin{figure}[hbt!]
     \centering
        \subfloat[Mean Corr]{
          \includegraphics[width=7.5cm]{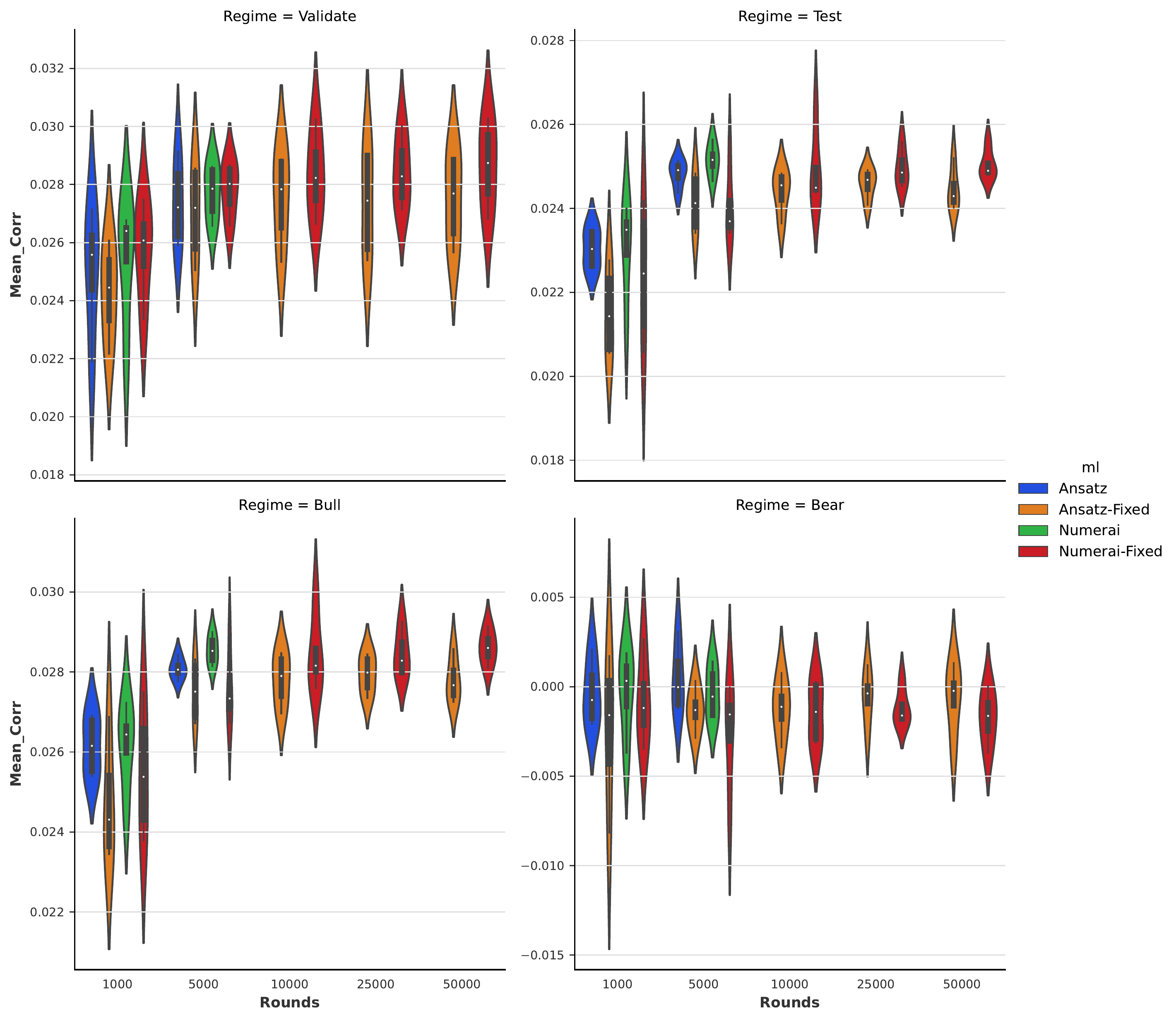}
        } 
        \subfloat[Sharpe]{
          \includegraphics[width=7.5cm]{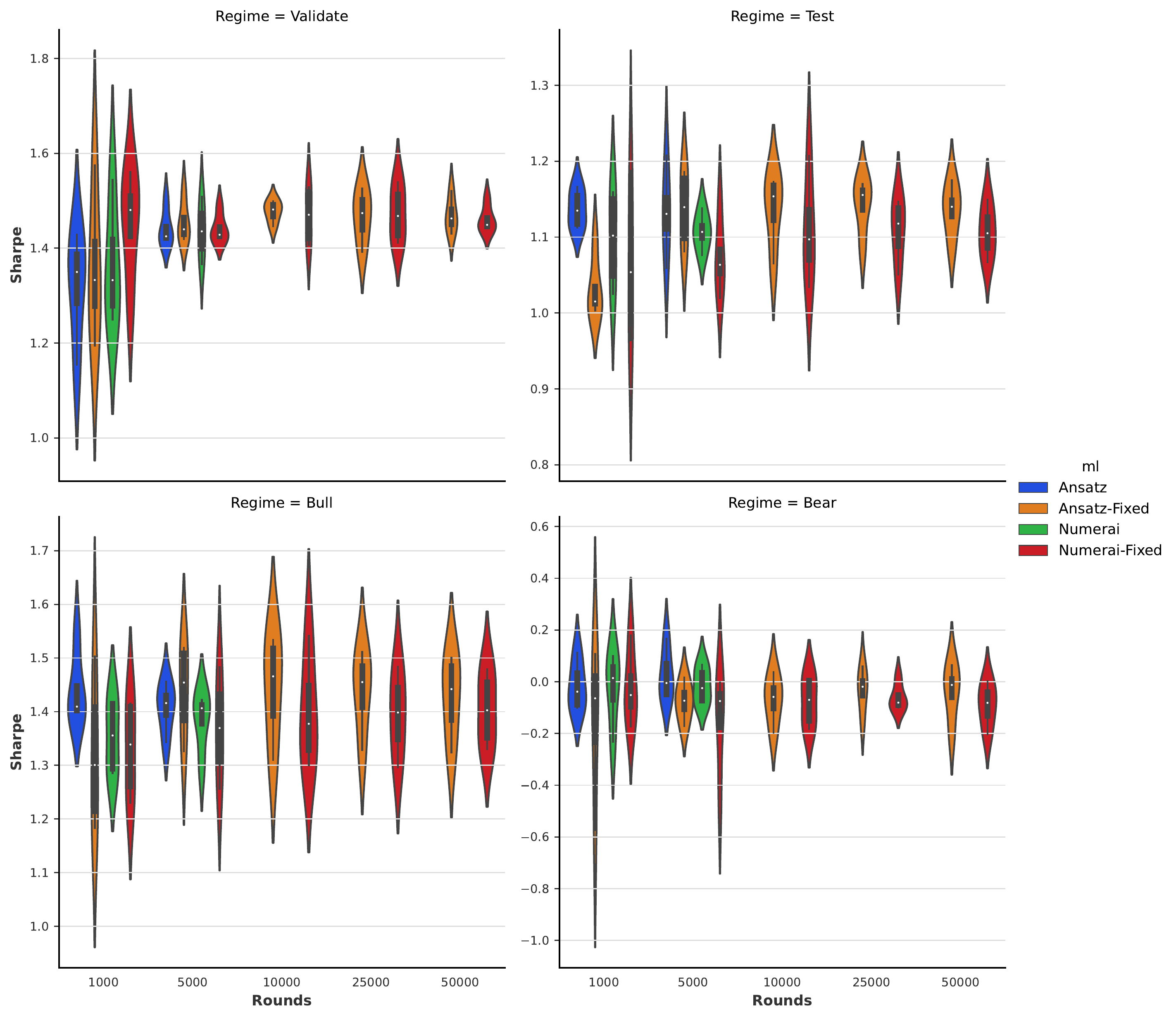}
        }
        \\
        \subfloat[Calmar]{
          \includegraphics[width=7.5cm]{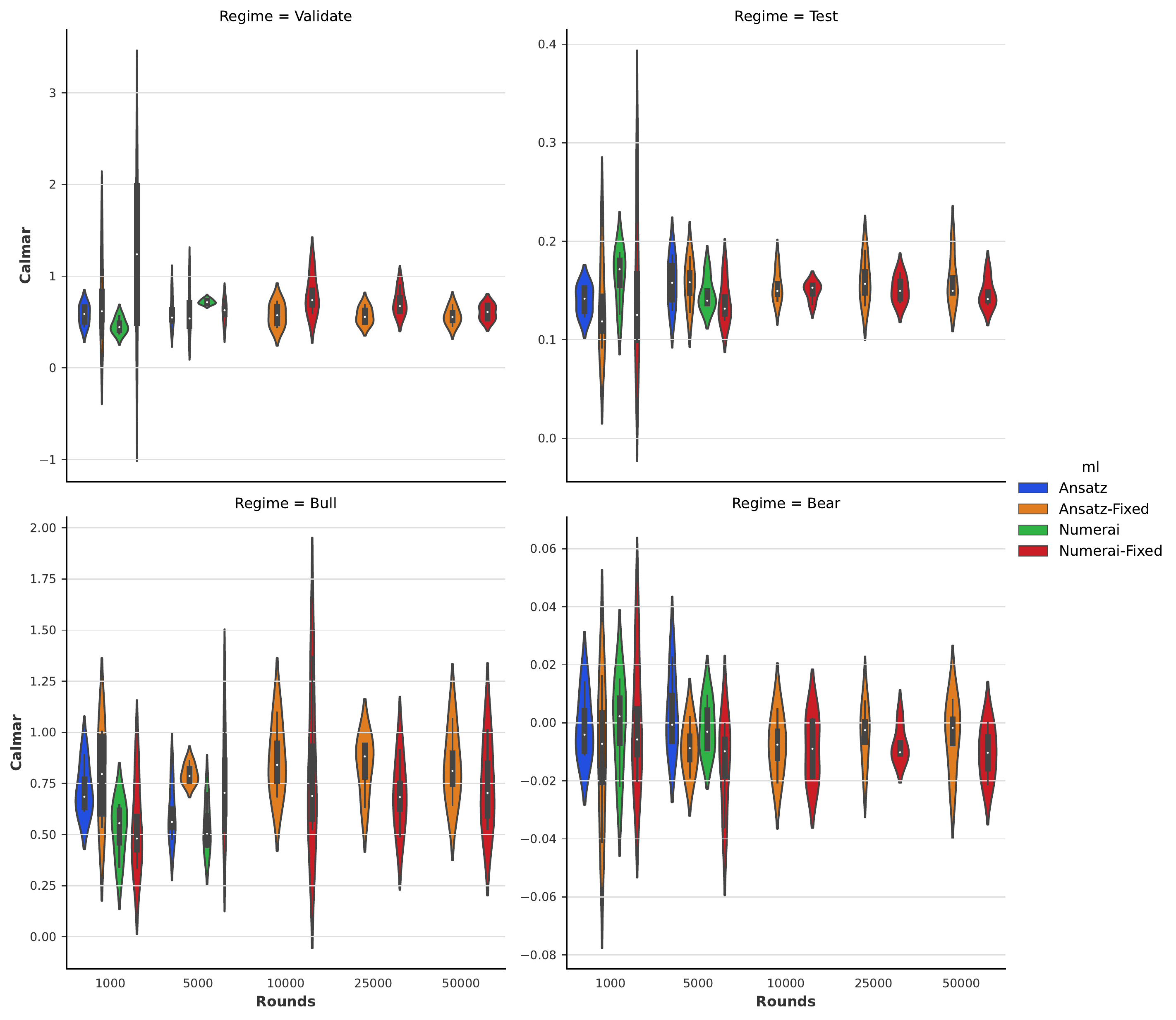}
        }
        \\
    \caption{Performances of benchmark XGBoost models with different number of boosting rounds $B=1000,5000,10000,25000,50000$ for risk metrics (a) Mean Corr, (b) Sharpe ratio and (c) Calmar ratio under different market regimes}
    \label{fig:Rain-BenchmarkSnapshot-Summary}
\end{figure}

Figure \ref{fig:Rain-BenchmarkSnapshot-Summary} shows the performance of the benchmark models with different number of boosting rounds $B=1000,5000,10000,25000,50000$. %A more detailed comparison for the benchmark models with $B=1000,5000$ is given in Figure \ref{fig:Rain-BenchmarkSnapshot-Summary2} in SI.
%% Size of models and performances 
Model performances for models with $B \geq 5000$ are not significantly different in both validation and test period. Within different market regimes in the test period, there are also no significant differences in model performance 
%There is little improvement of model performance 
for $B \geq 10000$. 

%% Retraining models help model performances  
Furthermore, there are no significant differences between each pair of models using the two different hyperparameter sets Ansatz and Numerai for a fixed data sampling method (regular retrain vs no retrain) in both the validation and test period. 
Yet as the number of boosting rounds increases, the performance differences between models with different hyperparameters narrows. This suggests that the differences between models performances are more due to differences between data sampling schemes used and to a lesser extent due to the hyperparameters. 

%% How we decide to use Ansatz not Numerai hyperparameters
To select our benchmark model hyperparameters, we consider both model performance and computational resources. Despite Numerai models having a slightly better Mean Corr than Ansatz models, they suffer from higher computational time and memory costs (explained in detail in SI Section \ref{section:NumeraiRain-benchmark}) and do not exhibit an improvement in Mean Corr in the validation period. Further, models with Ansatz hyperparameters also have a lower variance than those with Numerai hyperparameters across different risk metrics (Mean Corr, Sharpe ratio). Given their similar performance and characteristics, we select the Ansatz hyperparameter set to train different deep IL models in the next section.

\paragraph{How to measure the similarity between two GBDT models}

To measure the similarity between two GBDT models, we need to consider the overall structure similarity between two models also using feature importance, as this counts how many times a feature is used in a decision rule for a node within one of the trees in the GBDT model. It is not enough to consider only correlation between predictions because predictions that are similar but based on different decision rules can still offer diversification benefits to the ensemble by providing different learning pathways. If multiple independent learners arrive at similar predictions based on different information, the prediction becomes more robust to drifts in the data. 

\begin{definition*}[Structural Similarity of GBDT models]
For simplicity, a correlation based measure is used here to measure the overall similarity of two GBDT models, as follows
Given two GBDT models A and B with the same number of features $M$, let $R_A,R_B \in \mathbb{R^M_+}$ be the feature importance of the models, the structural similarity $\mathcal{S}(A,B) \in [-1,1]$ between two GBDT models is defined as the correlation between the normalised feature importance of the two models 
\begin{align}
    r_A &= \text{rank}(R_A) \in [0,1]^M \\
    r_B &= \text{rank}(R_B) \in [0,1]^M \\
    \mathcal{S}(A,B) &= \textbf{Corr}(r_A,r_B) 
    \label{eq:structural_similarity}
\end{align}
where $\textbf{Corr}(\cdot,\cdot)$ is the Pearson correlation function. 
\end{definition*}
Note that this measure considers the averaged contribution of each feature towards the model and ignores the interaction between features.

\subsubsection{Differences between benchmark models} 
\label{section:NumeraiRain-HyperDiff}

To understand the differences between models trained with different hyperparameters, we use the structural similarity measure~\eqref{eq:structural_similarity} 
%in Section \ref{section:NumeraiRain-deepIL} 
to understand the overall structural similarity between the Ansatz and Numerai benchmark models.

\begin{figure}
\begin{subfigure}{.48\textwidth}
\centering
\includegraphics[width=.8\linewidth]{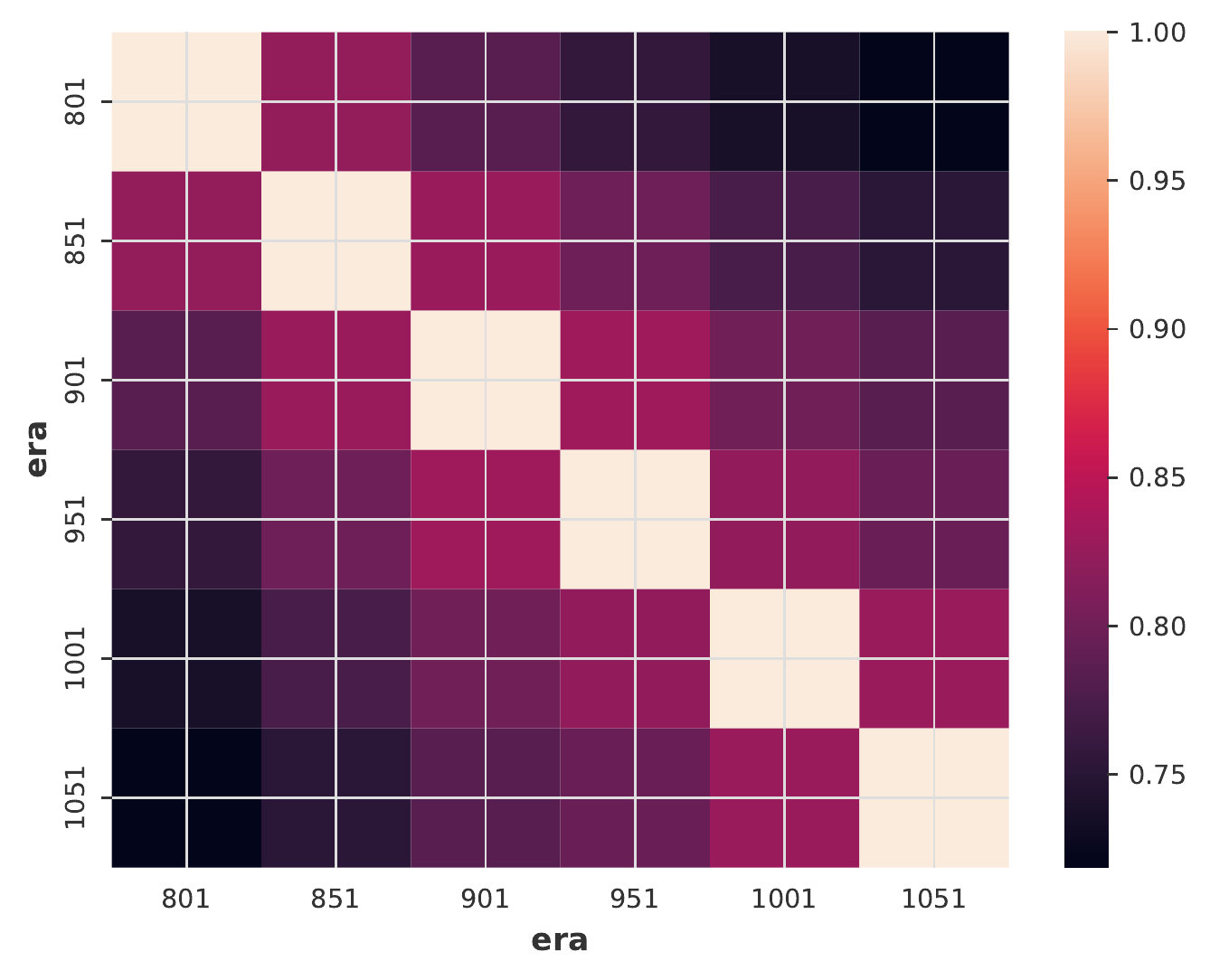}
\caption{Correlation between models with $B=1000$ and Ansatz hyperparameters}
\end{subfigure}
\quad
\begin{subfigure}{.48\textwidth}
\centering
\includegraphics[width=.8\linewidth]{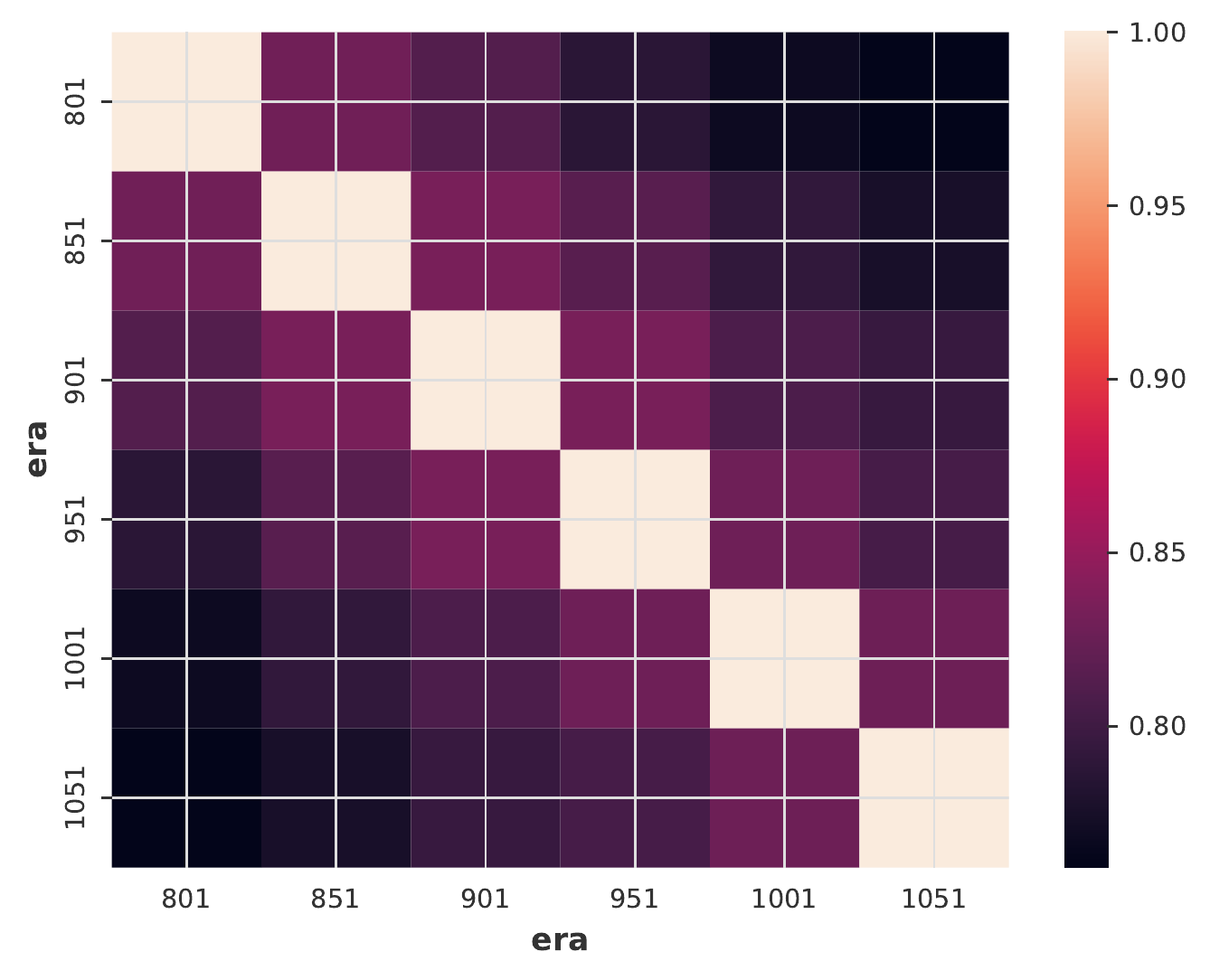}
\caption{Correlation between models with $B=1000$ and Numerai hyperparameters}
\end{subfigure}

\bigskip

\begin{subfigure}{.48\textwidth}
\centering
\includegraphics[width=.8\linewidth]{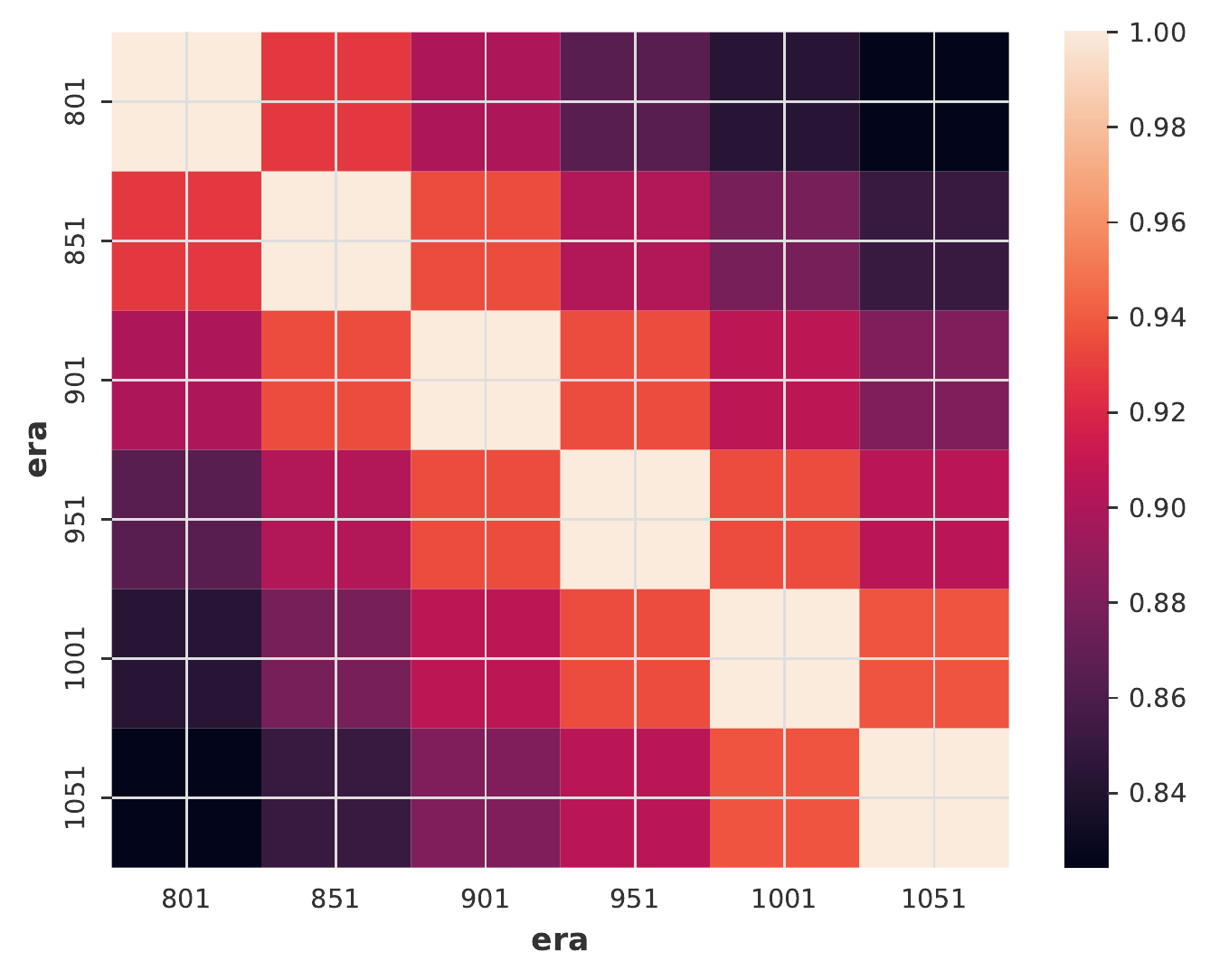}
\caption{Correlation between models with $B=5000$ and Ansatz hyperparameters}
\end{subfigure}
\quad
\begin{subfigure}{.48\textwidth}
\centering
\includegraphics[width=.8\linewidth]{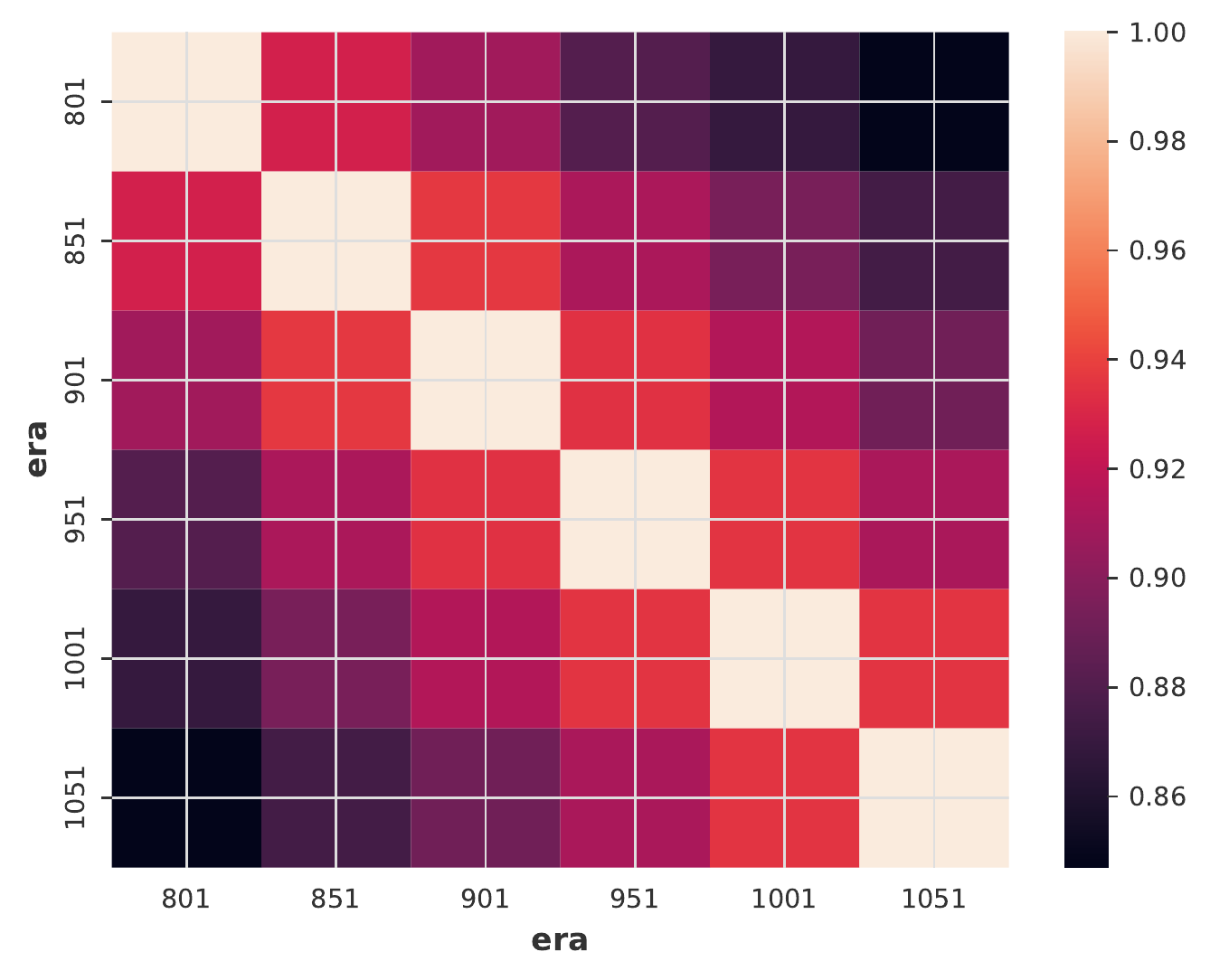}
\caption{Correlation between models with $B=5000$ and Numerai hyperparameters}
\end{subfigure}
\caption{Temporal correlation structure of benchmark models from Eras 801 to Era 1051.}
\label{fig:ModelCorr-Time}
\end{figure}

%% Temporal correlations 
In Figure \ref{fig:ModelCorr-Time}, we show the temporal correlation structure of the Ansatz and Numerai benchmark models with sizes $B=1000,5000$. As expected, the correlation between the retrained models decreases as the time gap between model retrains increases. Smaller models ($B=1000$) are less correlated with each other than larger models ($B=5000$), which can have $\mathcal{S}>0.9$ even after 150 eras. This observation supports our choice not to retrain models with sizes $B \geq 10000$, as the models are expected to be highly correlated with each other within the validation and test period.

\begin{figure}
\begin{subfigure}{.3\textwidth}
\centering
\includegraphics[width=.8\linewidth]{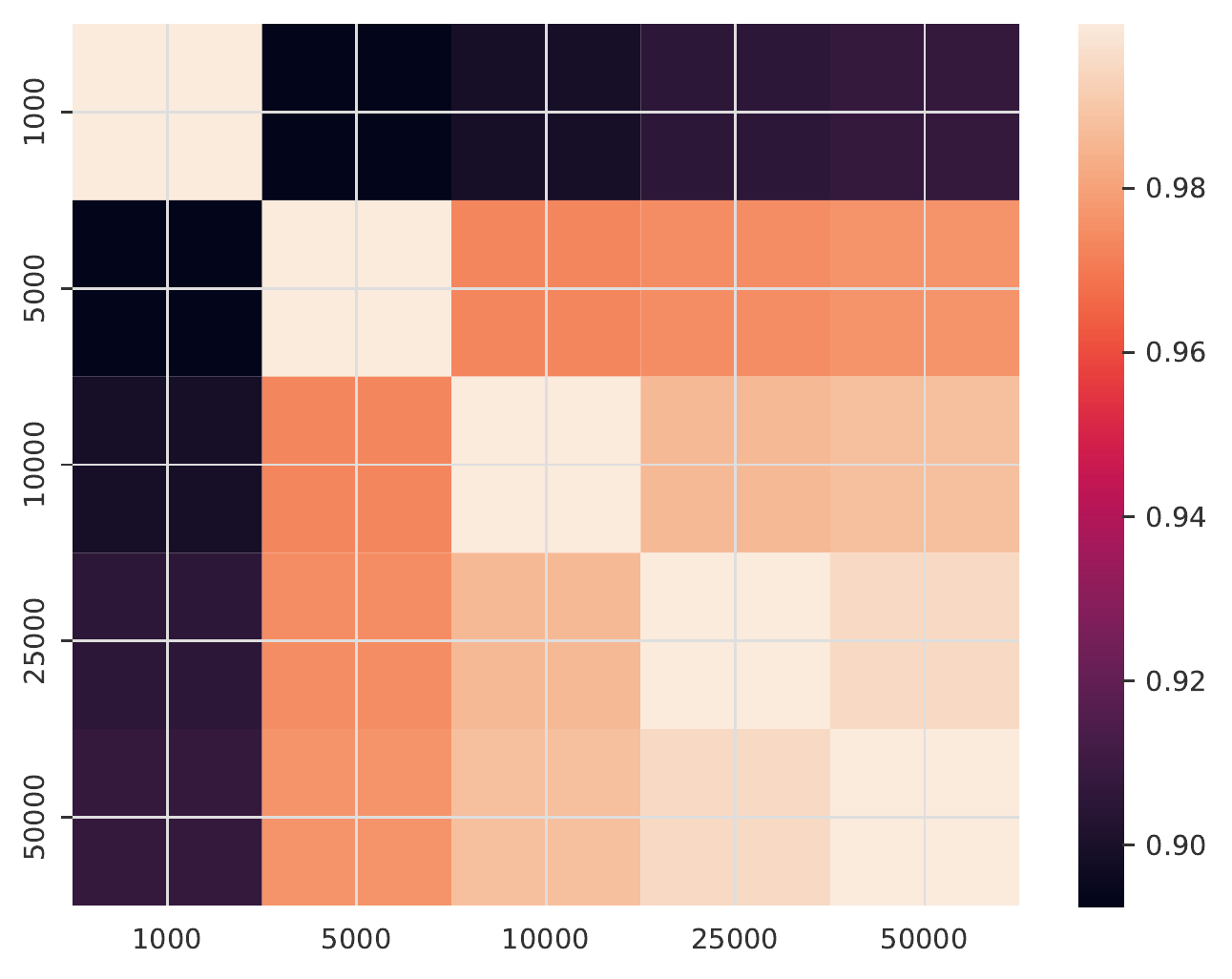}
\caption{Correlation between models with Ansatz hyperparameters}
\end{subfigure}
\hfill
\begin{subfigure}{.3\textwidth}
\centering
\includegraphics[width=.8\linewidth]{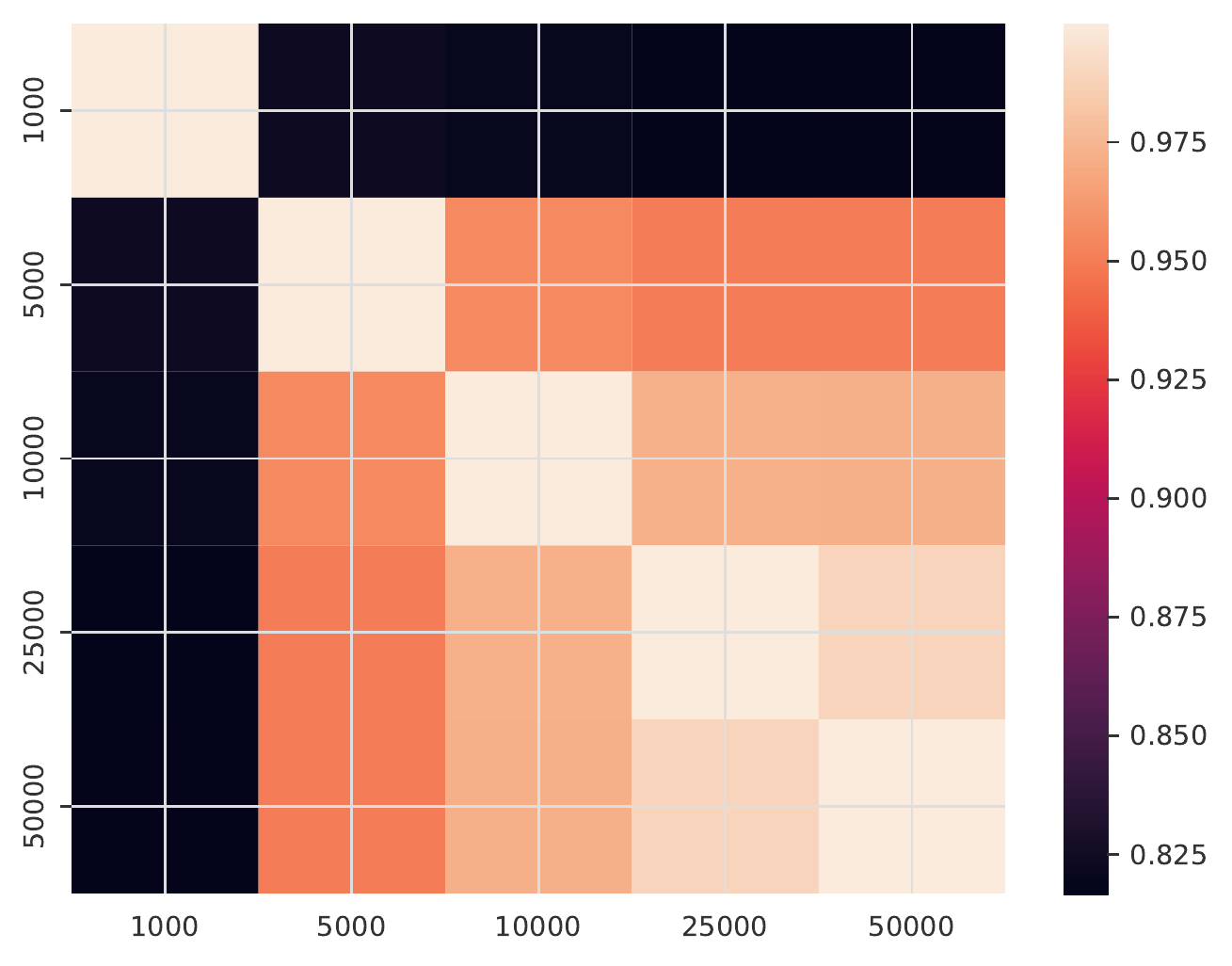}
\caption{Correlation between models with Numerai hyperparameters}
\end{subfigure}
\hfill
\begin{subfigure}{.3\textwidth}
\centering
\includegraphics[width=.8\linewidth]{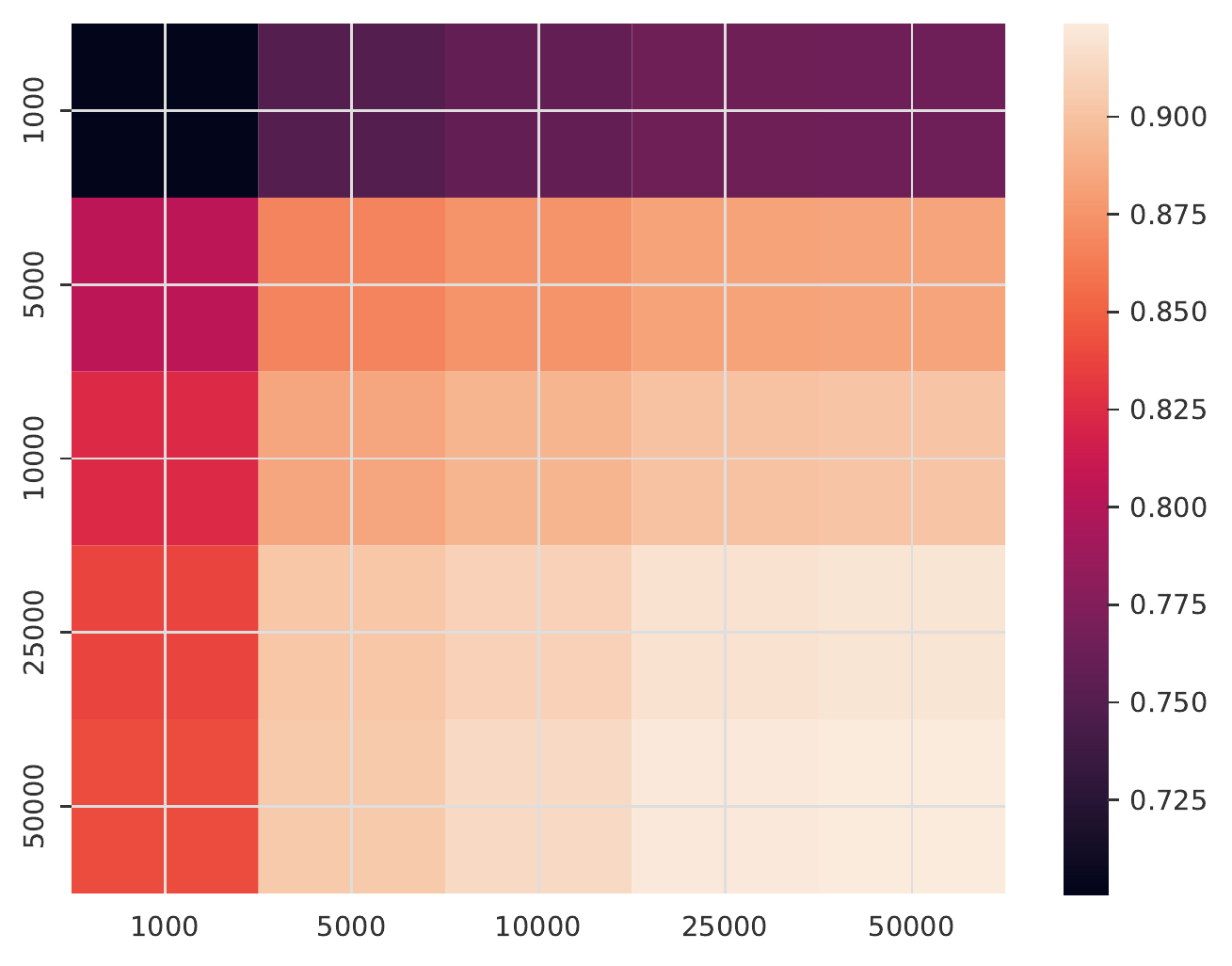}
\caption{Cross Correlation between Ansatz and Numerai models}
\end{subfigure}
\caption{Correlation structure of benchmark models of different sizes at Era 801}
\label{fig:ModelCorr-Size}
\end{figure}

%% Correlation of Sizes 
Figure \ref{fig:ModelCorr-Size} shows the correlation structure of benchmark models of different sizes at Era 801, the first model training time. For both Ansatz and Numerai hyperparameters, models with sizes $B \geq 5000$ are highly correlated. Cross-correlation between Ansatz and Numerai models of the same size is lower but the difference is negligible for models with sizes $B \geq 5000$. 

%% Convergences 
From the above observation, we hypothesise that models with different tree structure hyperparameters converge to the theoretical learning limit when the number of boosting rounds $B$ increases, on the condition that the learning rate $L$ of model is selected by the Ansatz formula. This means that hyperparameter optimisation is not necessary for large GBDT models so that we may pick any reasonable hyperparameter set (e.g., Ansatz) based on computational requirements.

Considering the fact model performances do not significantly improve beyond $B=5000$, we conclude it is not necessary to train \emph{single} models with size $B>5000$ as it consumes more computational resources while not providing meaningful gain in model performance. It is better instead to allocate the computational resources to train in parallel an ensemble of models with size $B \leq 5000$.

\section{Deep IL XGBoost Models}
\label{section:NumeraiRain-deepIL} 

We now deploy the full deep IL model with dynamic ensembling, in which models trained with different sampling schemes and hyperparameters are combined dynamically to create better models. 
This is inspired by our previous work on dynamic forecasting in financial data~\cite{wong2023online} and by models used in weather forecasting~\cite{Thoppil2021,}, where ensemble forecasting has been used to improve robustness of predictions.
%Both stock market prediction and weather forecasting share similar challenges of working with temporal data that is inherently unpredictable over long enough horizons. Ensemble forecasting \cite{Thoppil2021,} is widely used in weather forecast to improve robustness of predictions. 
Instead of creating predictions based on a single set of data/parameters, multiple sets of data/parameters are used to capture a range of scenarios, which represent possible trajectories for the evolution of weather or financial systems. 

% We report performances of the deep IL models from Era 801 to Era 1070 by the following regimes. The validation period is Era 801 to Era 885. The test period is between Era 901 and Era 1070, where 15 eras of data embargo are used. We also report performance based on market regimes within the test defined to understand if models behave differently under different regimes. 
%
% \begin{itemize}
%     \item Validation: Era 801 - Era 885 
%     \item Test: Era 901 - Era 1070
%     \item Bull: Era 901 - Era 1050
%     \item Bear: Era 1051 - Era 1070 
% \end{itemize}

A key assumption for model ensembling is to use a \textbf{diversified} set of base models that are not so correlated to achieve the variance reduction benefits during ensembling. As a result, we explore different sampling strategies to create diversified base models. In particular, we study model ensembles created with different (1) training sizes, (2) learning rates, (3) targets and (4) feature sets. 
Unless otherwise specified, we apply regular era sampling in training the XGBoost models in Layer 1, which in turn gives 4 different models for each deep IL ensemble strategy.

The deep IL models used a 2-layer model structure. The Layer 1 models are XGBoost models trained with different hyperparameters and settings described below. The Layer 2 models $N_k$, $1\leq k \leq 2$ are chosen as: 
\begin{itemize}
    \item $N_1$: Simple average over all predictions
    \item $N_2$: Ridge Regression with L2-regularisation $\alpha=1e-4$ and parameters are restricted to be non-negative.
\end{itemize}
The purpose of Layer 2 models is to refine predictions obtained in Layer 1. By combining predictions at individual observation (stock) level instead of model level, this approach is more flexible than the dynamic model selection. %used in Chapter \ref{chapter:NumeraiTitan}. 

\subsection{Ensemble strategies based on data sampling}

\subsubsection{Training Size Ensemble}

For incremental learning problems, it is not known in advance how much data is required for model learning. Trade-offs are made when deciding the training sizes. If more data is used, the training data can cover more historical regimes, but also have the risk of including data no longer relevant. If less data is used, the training data can adapt more quickly to concept drift, but can also increase the risk of overfitting the models towards the current data regime. Therefore, there is no universal rule to select the training set size. % Ad hoc decisions are often made based on prior knowledge on the dataset. 
%% Changing the standard train size 600 to 400-800 
The standard training set size recommended by Numerai is 600. Here, we explore if adjusting the training set sizes can improve model performances. 

In Algorithm \ref{alg:Rain-TrainingSizes}, the maximum training set size of Layer 1 models is increased to 800 eras and we train 5 models using the most recent $100\%,87.5\%,75\%,62.5\%,50\%$ of data. The number of boosting rounds is scaled with respect to training size. The learning rates of models is determined by the Ansatz formula $L=\frac{50}{B}$, using the scaled number of boosting rounds. 

%% In the above model, when the Layer 1 models are first trained at Era 801, the training set sizes of models would be 400,500,600,700,800. In the later model retrains, the gap between training set sizes of models would increase as more data are collected. This approach allows models to become more diverse as the overlap between training sets of the Layer 1 models decreases over time.  

%% Pseudo Code for Training Size Ensemble 
\begin{algorithm}[hbt!]
\caption{Deep IL XGBoost models over different training sizes}
\label{alg:Rain-TrainingSizes}
\KwIn{Number of boosting rounds $B=5000$, Max Training size of Layer 1 $X_1=800$, Data embargo $b_1=15$, $b_2=6$}
Set starting Era $D=801$ \\
Set Ansatz learning rate $L = \frac{50}{B}$ \\
Set Lookback ratios $r_1=1.0$, $r_2=0.875$, $r_3=0.75$, $r_4=0.625$, $r_5=0.5$ \\
\For{$1 \leq j \leq 5$}{
    Set Retrain period $T_j = \lfloor \frac{r_j X_1}{16} \rfloor $ \\
    \For{$1 \leq i \leq \lfloor \frac{270}{T_j} \rfloor $}{
        Set $D_1 = D + (i-1)T$ \\
        Set number of boosting rounds $B_j = r_j B$ \\
        Set learning rate $l_j = \frac{L}{r_j}$ \\ 
        Prepare training data $\mathcal{D}_j$ using $r_j$ proportion of data from Era 1 to $D_1-b_1 +(i-1)T$ \\
        Train Layer 1 XGBoost models $M_j^i$ with training data $\mathcal{D}_j$ using number of boosting rounds $B_j$ with learning rate $l_j$, other hyperparameters are unchanged. \\
        Obtain model predictions for $M_j^i$ from Era $D_1$ to Era $\min(D_1+T_j,1070)$ \\
    }
}
\For{$1 \leq j \leq 170$}{
    Set $D_2 = D + 99 + j$ \\
    \For{$1 \leq k \leq 2$}{
        Train Layer 2 models $N_k$ using the Layer 1 model predictions from Era $D_2-b_2-25$ to $D_2-b_2$ \\
        Obtain predictions from Layer 2 models $N_k$ for Era $D_2+1$ \\
    }
}
\end{algorithm}

\begin{figure}[htb!]
     \centering
        \subfloat[Mean Corr]{
          \includegraphics[width=15cm]{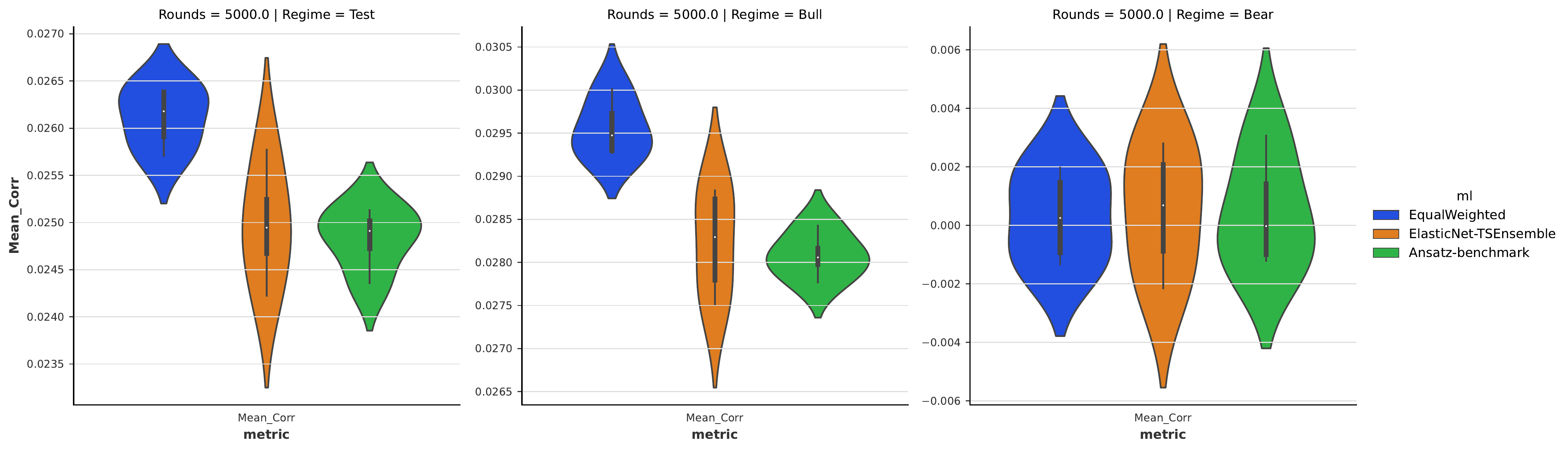}
        }
        \\
        \subfloat[Sharpe]{
          \includegraphics[width=15cm]{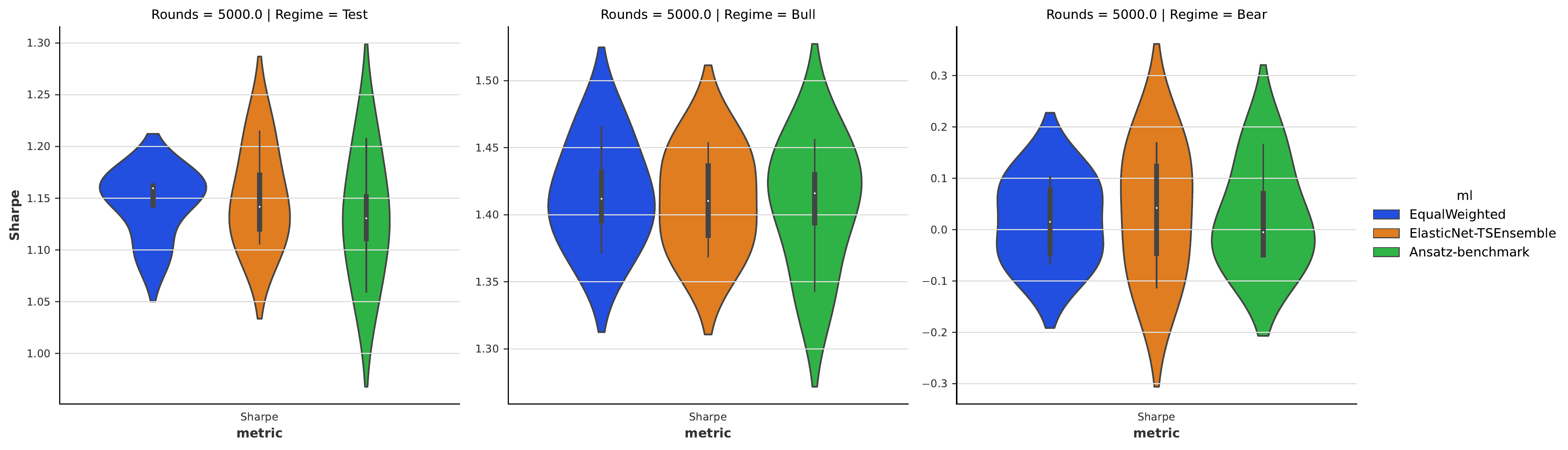}
        }
        \\
        \subfloat[Calmar]{
          \includegraphics[width=15cm]{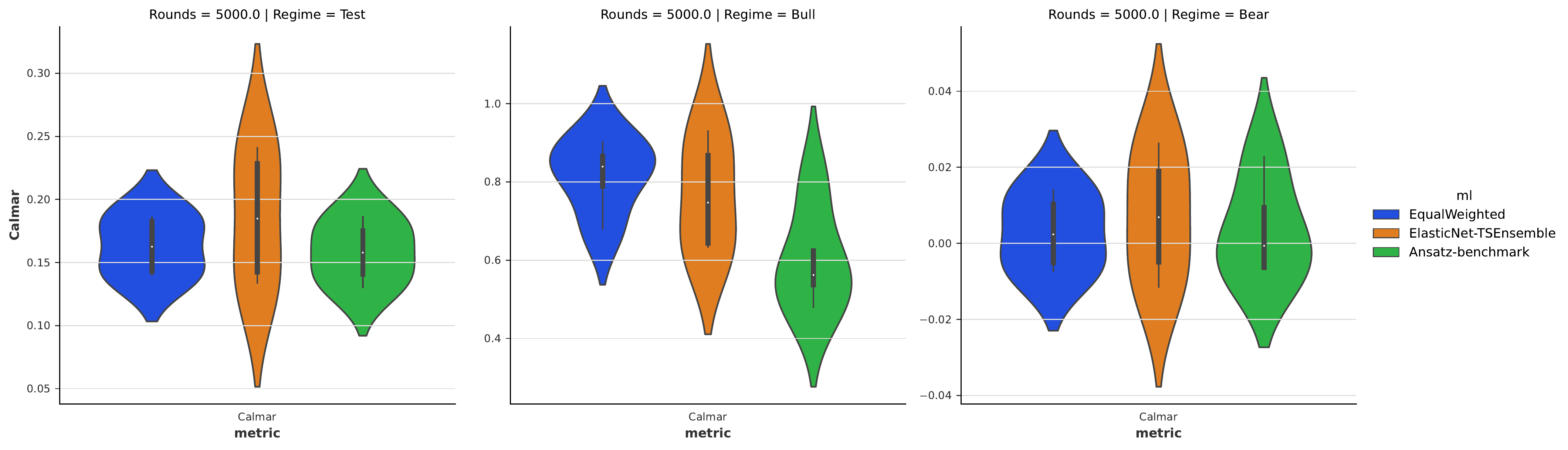}
        }
        \\
    \caption{Performances: (a) Mean Corr, (b) Sharpe ratio and (c) Calmar ratio of the deep IL XGBoost models with different training sizes under different market regimes with $B=5000$. }
    \label{fig:Rain-IL-TSEns}
\end{figure}

%% Results
In Figure \ref{fig:Rain-IL-TSEns}, we compare the performances of the two Layer 2 models (Elastic Net, Equal Weighted) with the benchmark Ansatz model of $B=5000$. The Equal Weighted model over all possible training set sizes achieves a higher Mean Corr than the benchmark model in the test period, yet  there is improvement in the Bull market but not in the Bear market. The Elastic Net model does not significantly improve the risk metrics compared to benchmark. Calmar ratio of Equal Weighted model is improved in the Bull market but not in the Bear market. %This suggests it is difficult to learn the optimal training set size for the Numerai dataset and using an equal weighted approach is more robust than models attempting to optimise that, such as Elastic Net. 

%% Layer 2 Models 

%% Train Sizes variation  

\begin{comment}

In Figure \ref{fig:ModelCorr-TrainSize}, XGBoost models with different training set sizes have structural similarity $0.7 <\mathcal{S} < 0.9$ at Era 801. While choosing a smaller training set size can decrease structural similarity, model performances are decreased at the same time, as shown in Figure \ref{fig:Rain-IL-TrainingSize-Layer1} in SI. Therefore, model ensembling based on training set size is not efficient to create model ensemble, compared to the methods suggested below. 

\begin{figure}[hbt!]
\includegraphics[width=.3\linewidth]{figure/chapter4/Rain_ModelCorr_TrainSize_Era801.pdf}
\caption{Structural similarity of models with different training set sizes at Era 801}
\label{fig:ModelCorr-TrainSize}
\end{figure}

\end{comment}

\newpage 
\subsection{Ensemble strategies based on different learning strategies}

\subsubsection{Learning Rate model ensemble (Complexity ensemble)}

Recent research suggests that features are learnt with different speeds within a neural network \cite{pezeshki2022multi,geiger2019jamming,}. Inspired by this idea, we combine GBDT models with different learning rates to learn models capturing both fast and slowing features. 

In Algorithm\ref{alg:Rain-LREns}, we combine XGBoost models with 5 different learning rates. For a given number of boosting rounds $B$, in addition to training the model of size $B$ as above, we train two larger models of size $2B$ and $4B$ and two smaller models of size $\frac{B}{2}$ and $\frac{B}{4}$ where the learning rate are adjusted by the Ansatz formula. To reduce computational costs, the two larger models are not regularly retrained. Only models with number of boosting rounds less than or equal to $B$ are regularly retrained. The Layer 2 models $N_k$ used in Algorithm\ref{alg:Rain-LREns} are the same as those used in Algorithm \ref{alg:Rain-TrainingSizes}.
%% LR Ens is equivalent to complexity ensemble using Ansatz formula 
Since the Ansatz formula is used to determine the learning rate $L$ and the number of boosting rounds $B$ pair for the Layer 1 models, the above procedure is equivalent to combining models with different complexities, where the number of boosting rounds $B$ is used to measure the complexity of GBDT models.

%% Pseudo Code for Learning rate ensemble 
\begin{algorithm}[hbt!]
\caption{Deep IL XGBoost models over different learning rates}
\label{alg:Rain-LREns}
\KwIn{Number of boosting rounds $B=5000$, Training size of Layer 1 $X_1=585$, Retrain Frequency $T=50$, Data embargo $b_1=15$, $b_2=6$}
Set starting Era $D=801$ \\
Set Ansatz learning rate $L = \frac{50}{B}$ \\
%% Models that regularly updated 
\For{$1 \leq i \leq 6$}{
    Set $D_1 = D +(i-1)T$ \\
     Prepare training data from Era 201 to $D_1-b_1 +(i-1)T$ \\
    \For{$1 \leq j \leq 3$}{
        Train Layer 1 XGBoost model $M_j^i$, with number of boosting rounds rounds $B_j = \frac{2B}{2^j}$ and learning rate $L_j = \frac{2^jL}{2}$ , other hyperparameters are unchanged. \\
        Obtain model predictions for $M_j^i$ from Era $D_1$ to Era $\min(D_1+50,1070)$ \\
    }
}
%% Models that are not regularly retraind
Prepare training data from Era 201 to 800 \\
\For{$4 \leq j \leq 5$}{
    Train Layer 1 XGBoost model $M_j$, with number of boosting rounds rounds $B_j = \frac{2^jB}{8}$ and learning rate $L_j = \frac{8L}{2^j}$ , other hyperparameters are unchanged. \\
    Obtain model predictions for $M_j$ from Era 801 to Era 1070 \\
}
\For{$1 \leq j \leq 170$}{
    Set $D_2 = D + 99 + j$ \\
    \For{$1 \leq k \leq 2$}{
        Train Layer 2 models $N_k$ using the Layer 1 model predictions from Era $D_2-b_2-25$ to $D_2-b_2$ \\
        Obtain predictions from Layer 2 models $N_k$ for Era $D_2+1$ \\
    }
}
\end{algorithm}

\begin{figure}[hbt!]
     \centering
        \subfloat[Mean Corr]{
          \includegraphics[width=15cm]{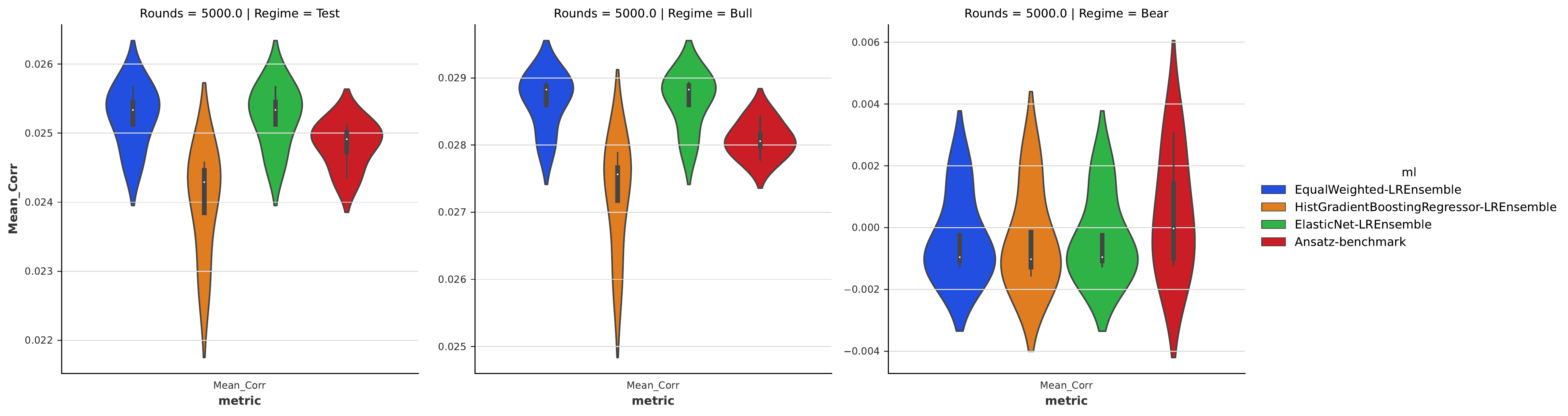}
        }
        \\
        \subfloat[Sharpe]{
          \includegraphics[width=15cm]{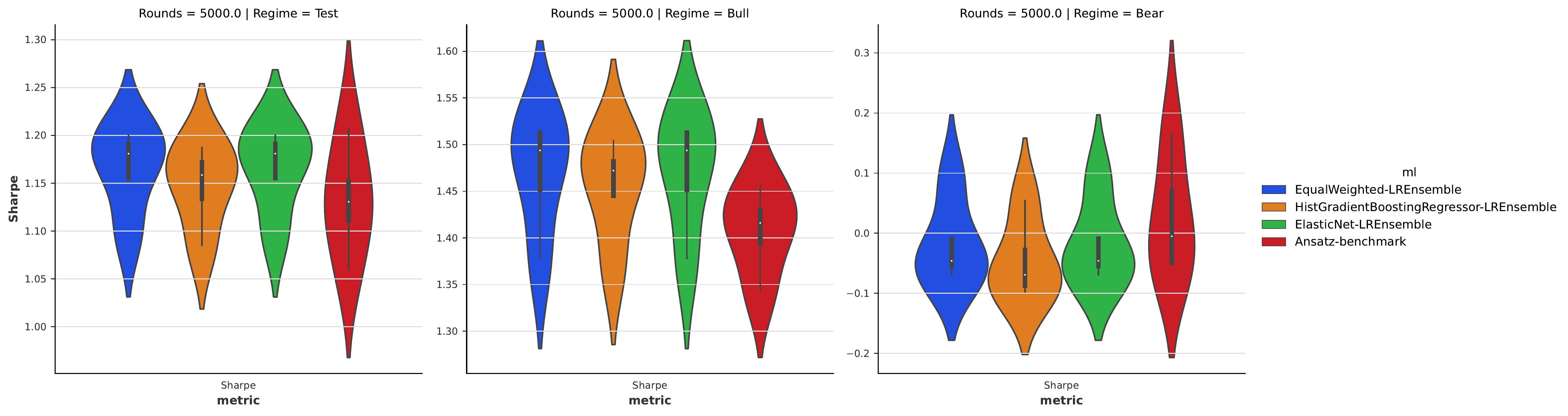}
        }
        \\
        \subfloat[Calmar]{
          \includegraphics[width=15cm]{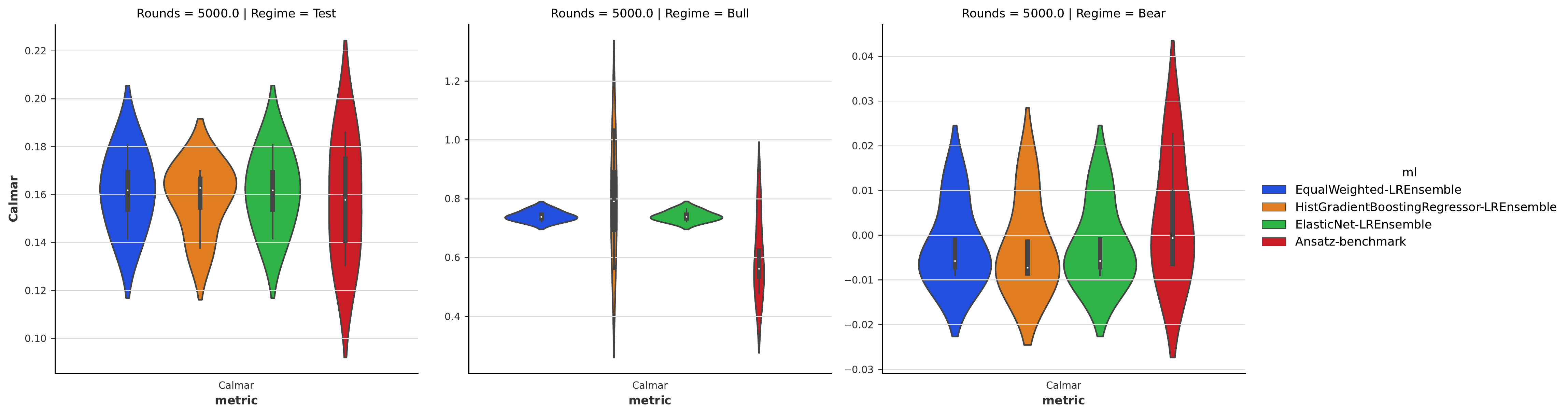}
        }
        \\
    \caption{Performances, (a) Mean Corr, (b) Sharpe ratio and (c) Calmar ratio of the deep IL XGBoost models with different learning rates under different market regimes. }
    \label{fig:Rain-IL-LREns}
\end{figure}

We run Algorithm \ref{alg:Rain-LREns} for $B=5000$ and performances for the two Layer 2 models are shown in Figure \ref{fig:Rain-IL-LREns}, compared with the Ansatz benchmark model with $B=5000$. Both the Equal Weighted and Elastic Net models improve Mean Corr and Sharpe ratio in the test period compared to the benchmark. Calmar ratio is also improved in the Bull market, but not in the Bear market. 

\begin{figure}[hbt!]
     \centering
        \subfloat[Mean Corr]{
          \includegraphics[width=12cm]{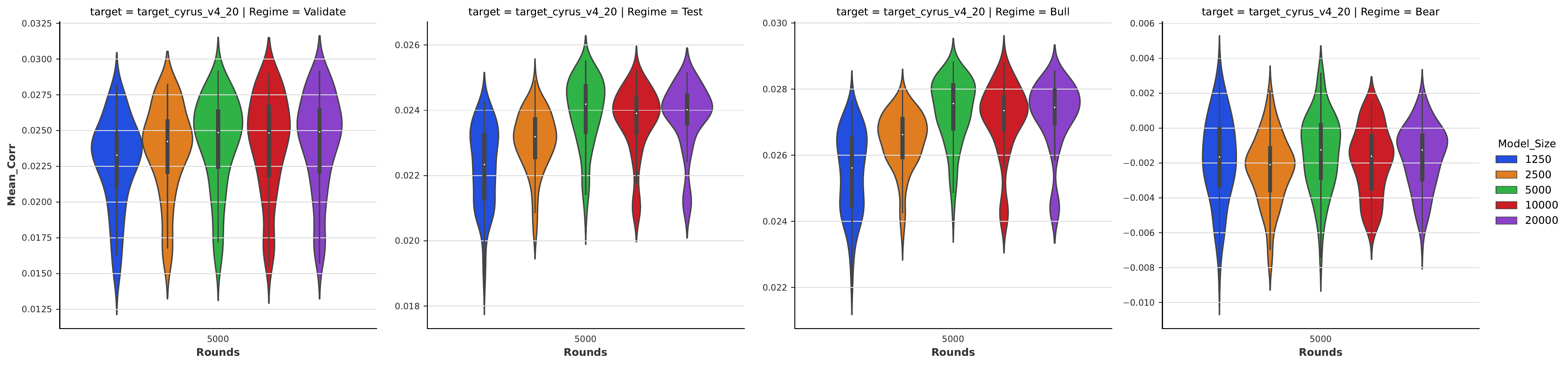}
        }
        \\
        \subfloat[Sharpe]{
          \includegraphics[width=12cm]{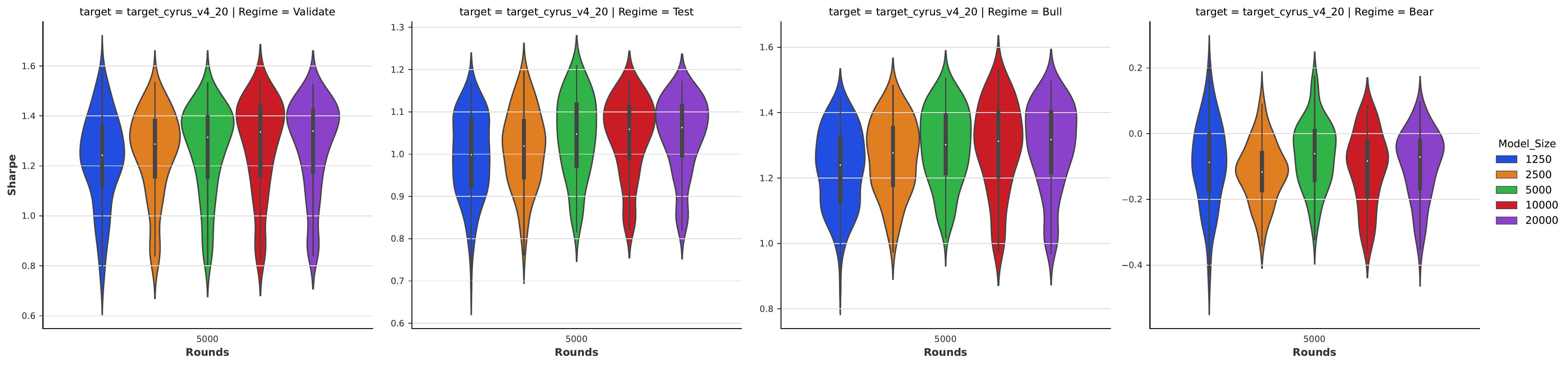}
        }
        \\
        \subfloat[Calmar]{
          \includegraphics[width=12cm]{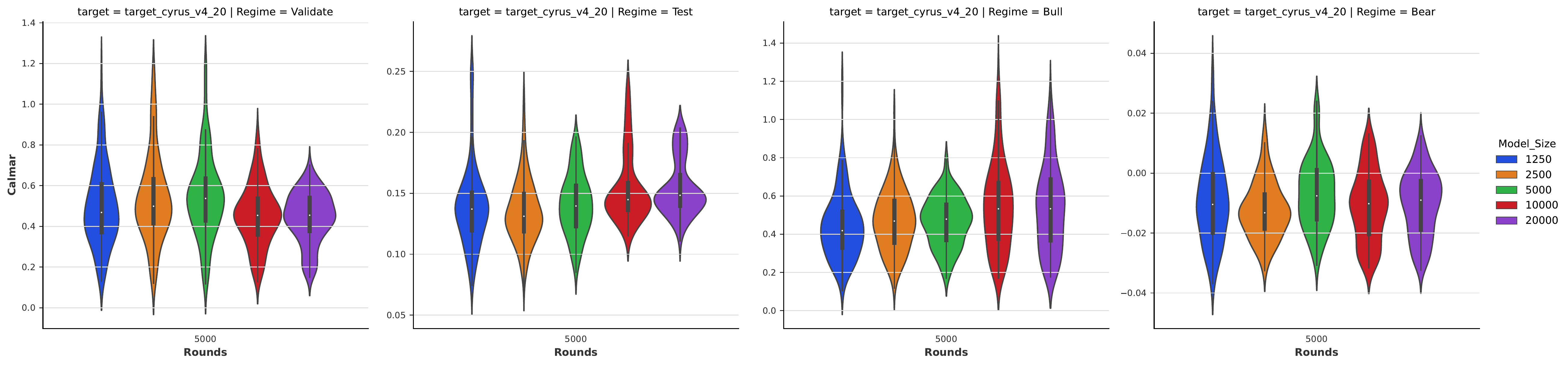}
        }
        \\
    \caption{Performances, (a) Mean Corr, (b) Sharpe ratio, and (c) Calmar ratio of the deep IL XGBoost models with different learning rates under different market regimes. }
    \label{fig:Rain-IL-LREnsemble-Layer1}
\end{figure}

%% LR Ens Results 
In Figure \ref{fig:Rain-IL-LREnsemble-Layer1} in SI, we show the learning curves of the 5 Layer 1 XGBoost models with different learning rates and the corresponding number of boosting rounds (1250,2500,5000,10000,20000). Although larger models perform slightly better than smaller models in the validation and test period, there are no significant differences in model performances in the Bear market. Therefore, there is no single optimal complexity across all regimes. This observation supports our proposed use of deep IL to combine the strength of models with different complexity (learning rates) so that the ensemble model is more robust.

%% Layer 2 Models 

\newpage 
\subsection{Ensemble strategies based on different targets}

%% Explain why use different targets to train models 

Feature projection was used to reduce drawdown of trading strategies in \cite{wong2023online}. In the V4.2 dataset \cite{numerai-datav4.2}, Numerai provides 5 different targets (Alpha-20D, Bravo-20D, Charlie-20D, Delta-20D, Echo-20D) in addition to the main scoring target (Cyrus-20D) which incorporates various risk management and hedging strategies. By design, these targets will offer a lower return (Mean Corr) but with lower risks (Max Drawdown and Volatility). The overall risk profile is improved even the portfolio return is reduced.

\subsubsection{Model ensemble with different targets and learning rates}

In Algorithm\ref{alg:Rain-LREns-Target}, we combine XGBoost models trained with the five different targets using different learning rates as in Algorithm \ref{alg:Rain-LREns}. In total, we train 25 Layer 1 XGBoost models to be combined in Layer 2. The Layer 2 models $N_k$ used in Algorithm\ref{alg:Rain-LREns-Target} are the same as those used in Algorithm \ref{alg:Rain-TrainingSizes}.

%% Pseudo Code for Learning rate ensemble 
\begin{algorithm}[hbt!]
\caption{Deep IL XGBoost models over different targets using different learning rates}
\label{alg:Rain-LREns-Target}
\KwIn{Number of boosting rounds $B=5000$, Training size of Layer 1 $X_1=585$, Retrain Frequency $T=50$, Data embargo $b_1=15$, $b_2=6$}
Set starting Era $D=801$ \\
Set Ansatz learning rate $L = \frac{50}{B}$ \\
Set Learning Targets $y_1,y_2,y_3,y_4,y_5$ to be Alpha-20D, Bravo-20D, Charlie-20D, Delta-20D, Echo-20D \\ 
%% Models that regularly updated 
\For{$1 \leq k \leq 5$}{
    \For{$1 \leq i \leq 6$}{
        Set $D_1 = D +(i-1)T$ \\
        Prepare training data from Era 201 to $D_1-b_1 +(i-1)T$ \\
        \For{$1 \leq j \leq 3$}{
            Train Layer 1 XGBoost model $M_{j,k}^i$, with number of boosting rounds rounds $B_j = \frac{2B}{2^j}$ and learning rate $L_j = \frac{2^jL}{2}$ using target $y_k$ , other hyperparameters are unchanged. \\
            Obtain model predictions for $M_j^i$ from Era $D_1$ to Era $\min(D_1+50,1070)$ \\
        }
    }
}
%% Models that are not regularly retraind
Prepare training data from Era 201 to 800 \\
\For{$1 \leq k \leq 5$}{
    \For{$4 \leq j \leq 5$}{
        Train Layer 1 XGBoost model $M_{j,k}$, with number of boosting rounds rounds $B_j = \frac{2^jB}{8}$ and learning rate $L_j = \frac{8L}{2^j}$  using target $y_k$, other hyperparameters are unchanged. \\
        Obtain model predictions for $M_j$ from Era 801 to Era 1070 \\
    }
}
\For{$1 \leq j \leq 170$}{
    Set $D_2 = D + 99 + j$ \\
    \For{$1 \leq k \leq 2$}{
        Train Layer 2 models $N_k$ using the Layer 1 model predictions from Era $D_2-b_2-25$ to $D_2-b_2$ \\
        Obtain predictions from Layer 2 models $N_k$ for Era $D_2+1$ \\
    }
}
\end{algorithm}

\begin{figure}[hbt!]
     \centering
        \subfloat[Mean Corr]{
          \includegraphics[width=15cm]{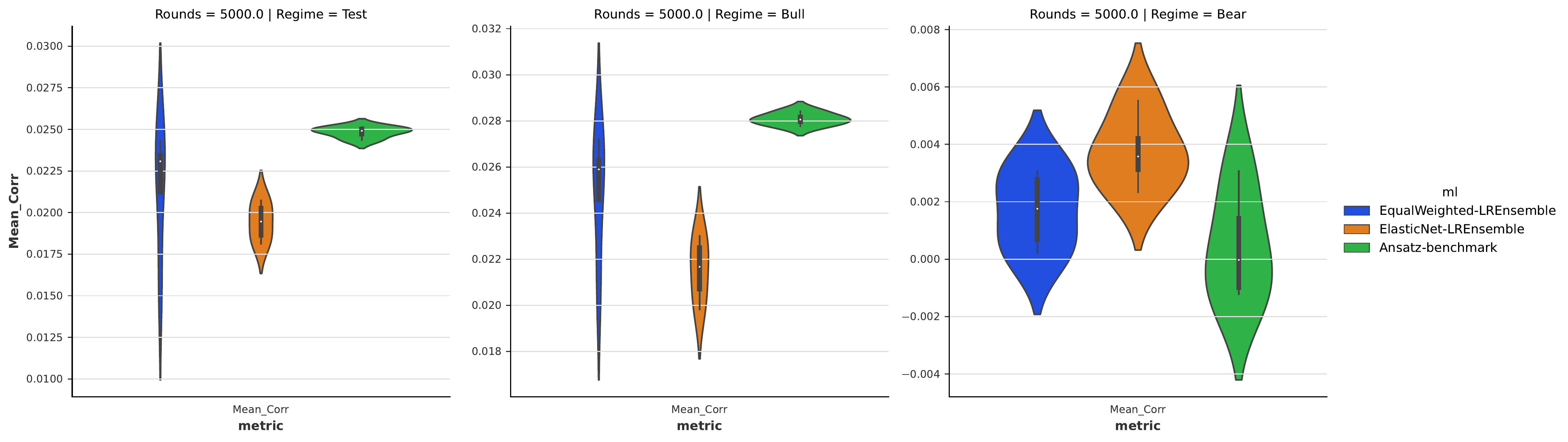}
        }
        \\
        \subfloat[Sharpe]{
          \includegraphics[width=15cm]{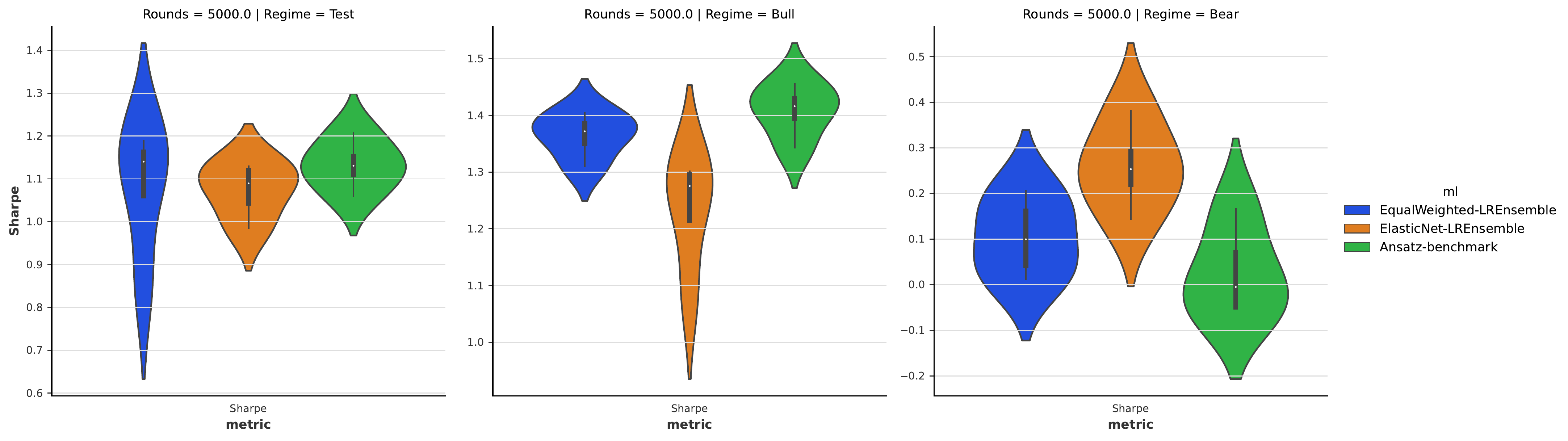}
        }
        \\
        \subfloat[Calmar]{
          \includegraphics[width=15cm]{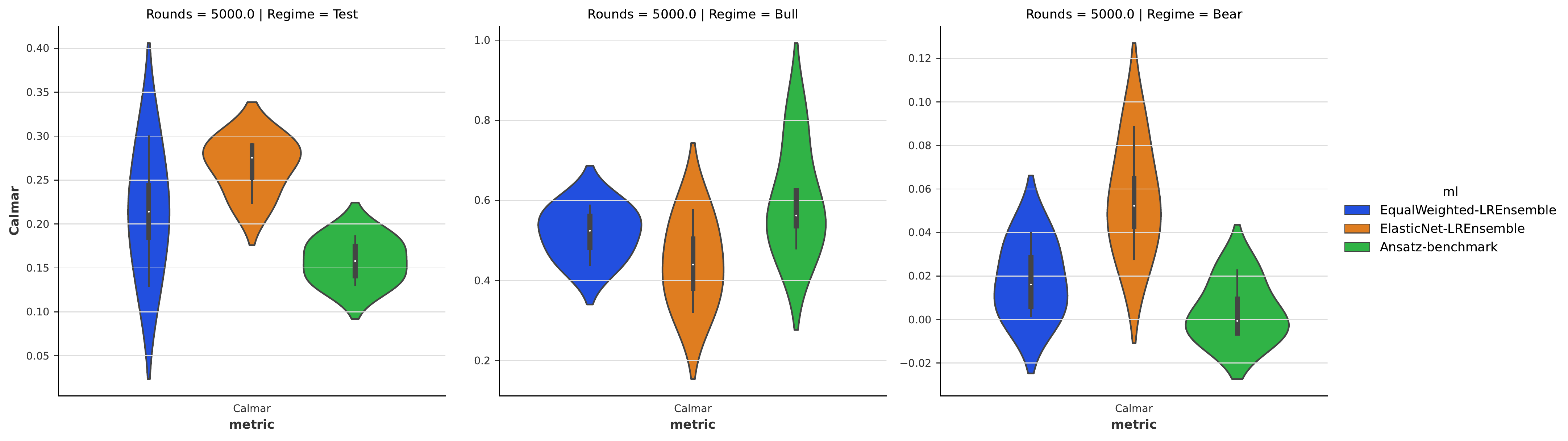}
        }
        \\
    \caption{Performances, (a) Mean Corr, (b) Sharpe ratio and (c) Calmar ratio of the deep IL XGBoost models with different targets and learning rates under different market regimes. }
    \label{fig:Rain-IL-LREns-Targets}
\end{figure}

In Figure \ref{fig:Rain-IL-LREns-Targets}, we compare the performances of the two Layer 2 models (Elastic Net, Equal Weighted) against the benchmark Ansatz model of $B=5000$.  The Equal Weighted model over all 25 Layer 1 XGBoost models with different targets over different learning rates achieves a higher Sharpe and Calmar ratio than the benchmark model in the test period at a lower Mean Corr ($\approx 90\%$ of the Benchmark model). Elastic Net can further improve the Calmar ratio but with a further lower Mean Corr ($\approx 75\%$ of the Benchmark model). 
The improvement of Sharpe and Calmar of models using different targets can be attributed to a lower downside in the Bear market. Employing various hedging strategies, such as using the risk-controlled targets in model training will result in a lower performance in Bull market. The diversification benefits can only been observed when there are regime changes in the data, such as during the Bear market where the benchmark unhedged strategy performs poorly. Therefore, to fairly access the merit of different hedging strategies, the test period needs to be long enough to cover different market regimes.

%% Layer 2 Models 

\newpage 
\subsection{Ensemble strategies based on feature sampling}

\subsubsection{Feature Sets model ensemble}
\label{section:NumeraiRain-FeatureJK}

Feature selection and sampling methods are useful in training models. Random feature sampling, a common procedure used at the \emph{local} level before the start of training each tree can be applied at the \emph{global} level before model training. By design, the learnt models are more diverse, as some features will never be used in the overall model rather than simply missing in some trees. It also lowers computational requirements of models as we do not have to fit all the data to the model. Here, we study Jackknife sampling \cite{Efron1981,} among other sampling techniques to build diversified models suitable for ensembling.

%% Why Feature Group Labels
The feature group labels in Numerai V4.2 dataset can be considered as a way of feature clustering using domain knowledge. Instead of analysing the high-dimensional temporal correlation structure of the features by clustering or dimensionality reduction methods, we use the labels provided by Numerai, which are created with the knowledge of data sources and feature generation process to correctly group features into different categories. This approach saves computational time and avoids identifying spurious relationships between features. 

%% What are the Jackknife sets 
Jackknife feature sets $\mathcal{F}_j$, for $1\leq j \leq 10$ are created as follows. For each for the 10 feature groups (Intelligence, Charisma, Strength, Dexterity, Constitution, Wisdom, Agility, Serenity, Sunshine, Rain), we remove that set from the 2132 features one at a time, and then use the remaining 9 groups to form the Jackknife feature sets $\mathcal{F}_{1} \dots \mathcal{F}_{10}$. The Jackknife feature sets are then used to train XGBoost models, using the procedure described in Algorithm\ref{alg:Rain-JK}. The Layer 2 models $N_k$ used in Algorithm\ref{alg:Rain-JK} are the same as those used in Algorithm\ref{alg:Rain-TrainingSizes}. 

%% How to create benchmark 
To evaluate the usefulness of feature group labels in model building, we compare our approach with two different baseline methods: (i) Deep IL XGBoost models over random feature sampling, as described in Algorithm\ref{alg:Rain-Random}; and (ii) benchmark XGBoost models trained with all the features using Ansatz hyperparameters. The Layer 2 models $N_k$ used in Algorithm\ref{alg:Rain-Random} are the same as those used in Algorithm\ref{alg:Rain-JK}.
The reason to use (i) as a benchmark is to calibrate if feature group labels offer information that is better than random in separating the features into groups representing different signal sources, thus creating information barriers between models so that they are forced to learn rules that are different from each other.  This would reduce correlation between predictions. The reason to use (ii) as a benchmark is to check if any form of feature selection is beneficial to model performance at all.

%% Pseudo Code for Jackknife and random feature sampling 
\begin{algorithm}[hbt!]
\caption{Deep IL XGBoost models over feature set Jackknife sampling}
\label{alg:Rain-JK}
\KwIn{Number of boosting rounds $B=5000$, Training size of Layer 1 $X_1=585$, Retrain Frequency $T=50$, Data embargo $b_1=15$, $b_2=6$}
Set starting Era $D=801$ \\
Set Ansatz learning rate $L = \frac{50}{B}$ \\
\For{$1 \leq i \leq 6$}{
     Set $D_1 = D  +(i-1)T$ \\
     Prepare training data from Era 201 to $D_1-b_1 +(i-1)T$ \\
    \For{$1 \leq j \leq 10$}{
        Train Layer 1 XGBoost models $M_j^i$, with feature set $\mathcal{F}_j$, other hyperparameters are unchanged. \\
        Obtain model predictions for $M_j^i$ from Era $D_1$ to Era $\min(D_1+50,1070)$ \\
    }
}
\For{$1 \leq j \leq 170$}{
    Set $D_2 = D + 99 + j$ \\
    \For{$1 \leq k \leq 2$}{
        Train Layer 2 models $N_k$ using the Layer 1 model predictions from Era $D_2-b_2-25$ to $D_2-b_2$ \\
        Obtain predictions from Layer 2 models $N_k$ for Era $D_2+1$ \\
    }
}
\end{algorithm}

\begin{algorithm}[hbt!]
\caption{Deep IL XGBoost models over random feature sampling}
\label{alg:Rain-Random}
\KwIn{Number of boosting rounds $B$, Training size of Layer 1 $X_1=585$, Retrain Frequency $T=50$, Data embargo $b_1=15$, $b_2=6$}
Set starting Era $D=801$ \\
Set Ansatz learning rate $L = \frac{50}{B}$ \\
\For{$1 \leq i \leq 6$}{
    Set $D_1 = D  +(i-1)T$ \\
    Prepare training data from Era 201 to $D_1-b_1 +(i-1)T$  \\
    \For{$1 \leq j \leq 10$}{
        Train Layer 1 XGBoost models $M_j^i$, with $50\%$ of the 2132 features selected by random without replacement, other hyperparameters are unchanged. \\
        Obtain predictions for $M_j^i$ from Era $D_1$ to Era $\min(D_1+50,1070)$ \\
    }
}
\For{$1 \leq j \leq 170$}{
    Set $D_2 = D + 99 + j$ \\
    \For{$1 \leq k \leq 2$}{
        Train Layer 2 models $N_k$ using the Layer 1 model predictions from Era $D_2-b_2-25$ to $D_2-b_2$ \\
        Obtain predictions from Layer 2 models $N_k$ for Era $D_2+1$ \\
    }
}
\end{algorithm}

\begin{figure}[hbt!]
     \centering
        \subfloat[Mean Corr]{
          \includegraphics[width=15cm]{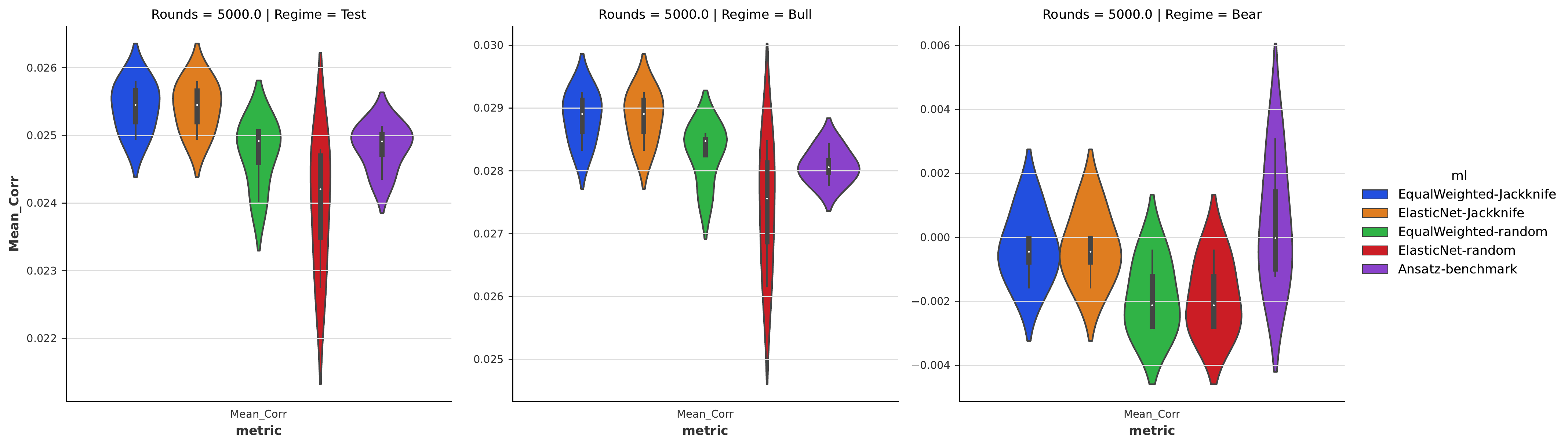}
        }
        \\
        \subfloat[Sharpe]{
          \includegraphics[width=15cm]{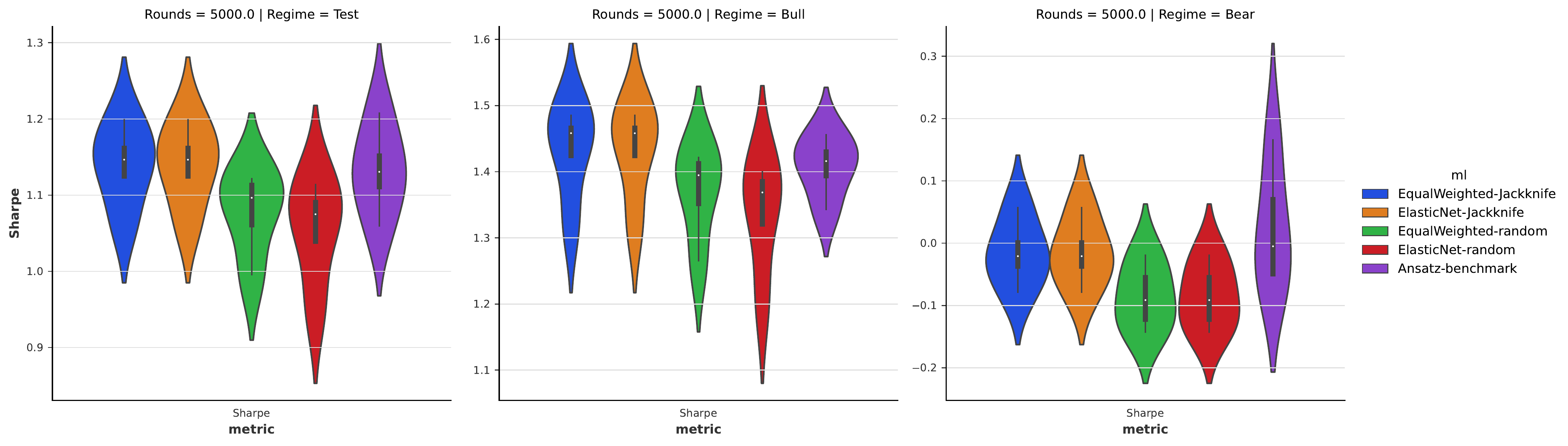}
        }
        \\
        \subfloat[Calmar]{
          \includegraphics[width=15cm]{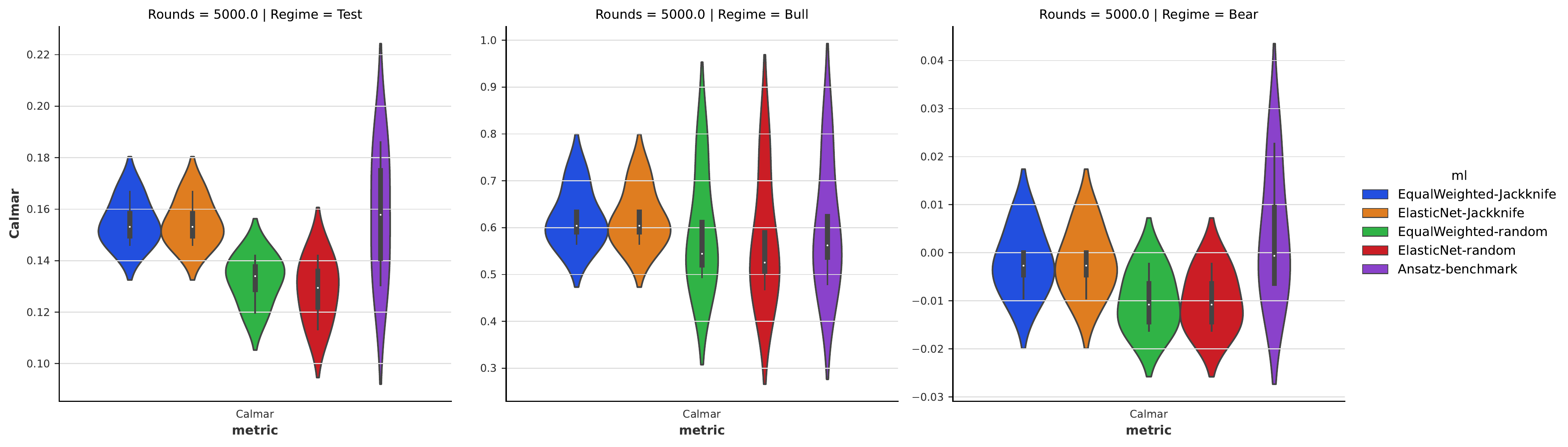}
        }
        \\
    \caption{Performances, (a) Mean Corr, (b) Sharpe ratio, and (c) Calmar ratio of the deep IL XGBoost models with Jackknife and random feature sampling under different market regimes. }
    \label{fig:Rain-IL-FeatureSampling}
\end{figure}

%% Results 
We run these two algorithms for $B=5000$. In Figure \ref{fig:Rain-IL-FeatureSampling} we show the performances of the four Layer 2 models from Jackknife and random feature sampling against the benchmark Ansatz model with $B=5000$, which is also regularly retrained. 
%% Feature JK 
The Layer 2 models from Jackknife sampling have a higher Mean Corr and Sharpe ratio than the models from random sampling and the benchmark model in both the validation and test period. The Layer 2 models using random sampling have comparable Mean Corr and Sharpe ratio with the benchmark model in both the validation and test period. Within models using Jackknife sampling, there are  no significant differences between the Equal Weighted and Elastic Net model. However, for models using random sampling, Elastic Net model underperformed relative to the Equal weighted model. The historical performances of models from random sampling are simply noise, and we are not supposed to be able to learn any useful patterns from them.

\begin{figure}[hbt!]
\begin{subfigure}{.3\textwidth}
\centering
\includegraphics[width=.8\linewidth]{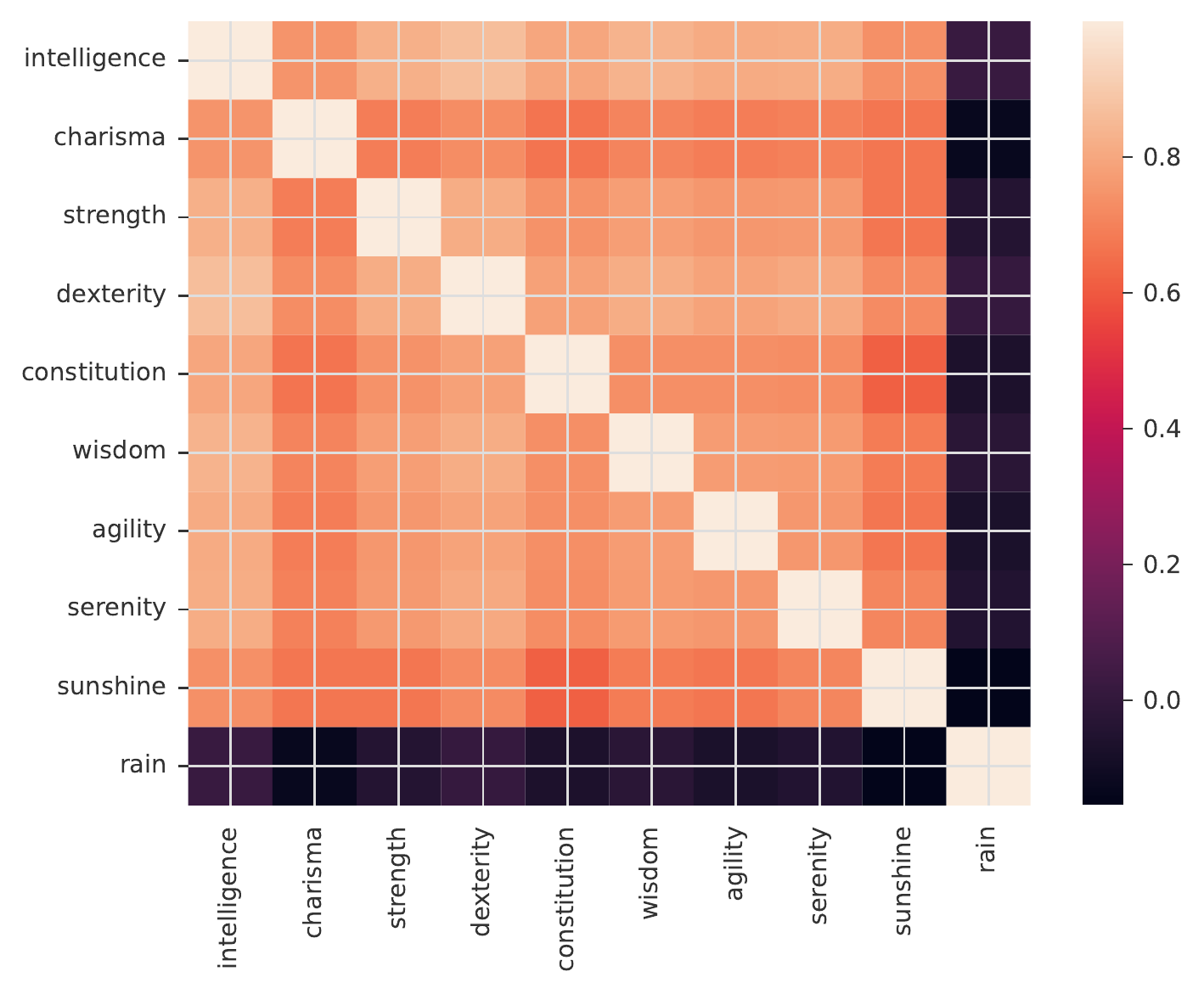}
\caption{Structural similarity of models with Jackknife feature sampling at Era 801}
\end{subfigure}
\quad
\begin{subfigure}{.3\textwidth}
\centering
\includegraphics[width=.8\linewidth]{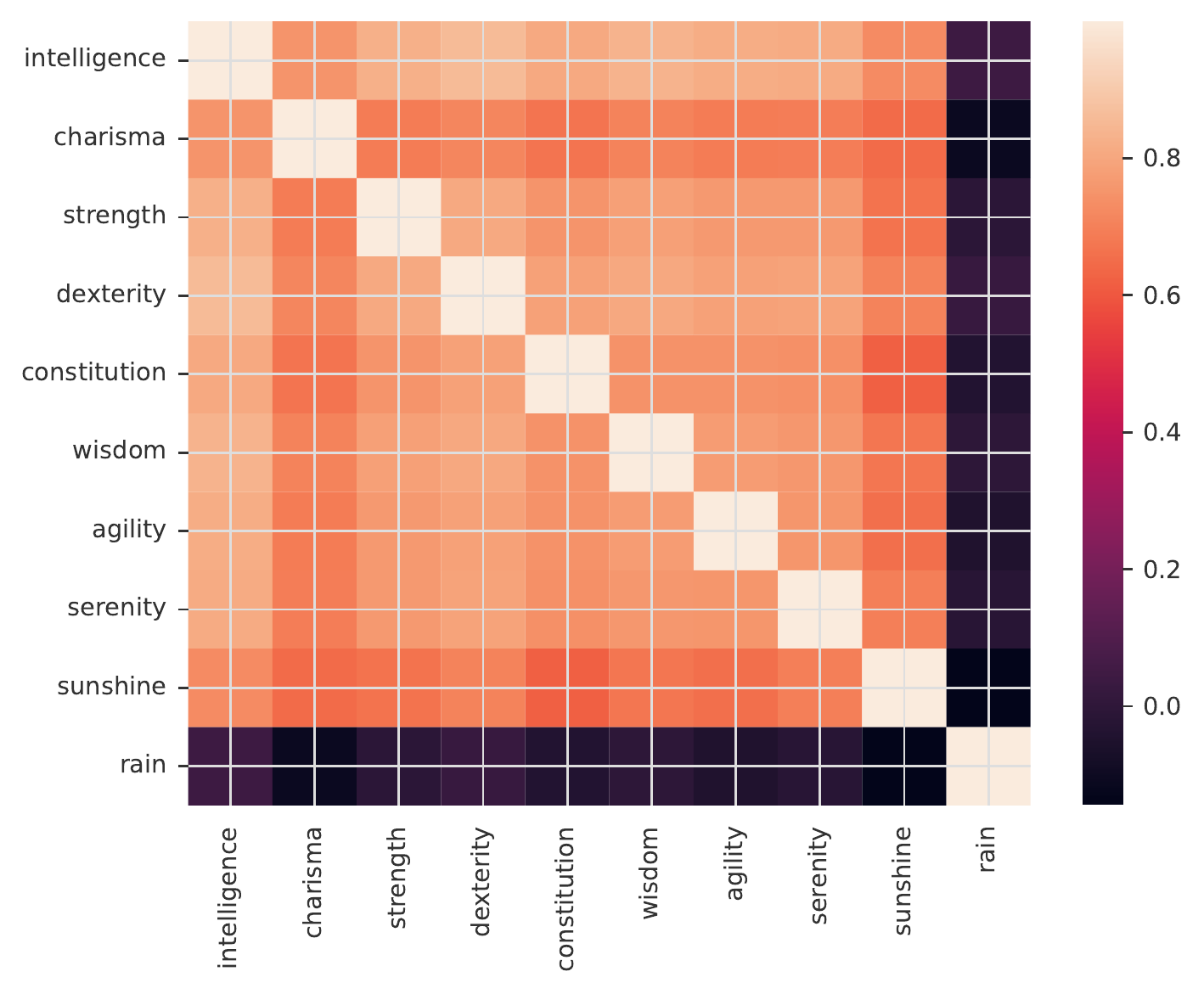}
\caption{Structural similarity of models with Jackknife feature sampling at Era 901}
\end{subfigure}
\quad
\begin{subfigure}{.3\textwidth}
\centering
\includegraphics[width=.8\linewidth]{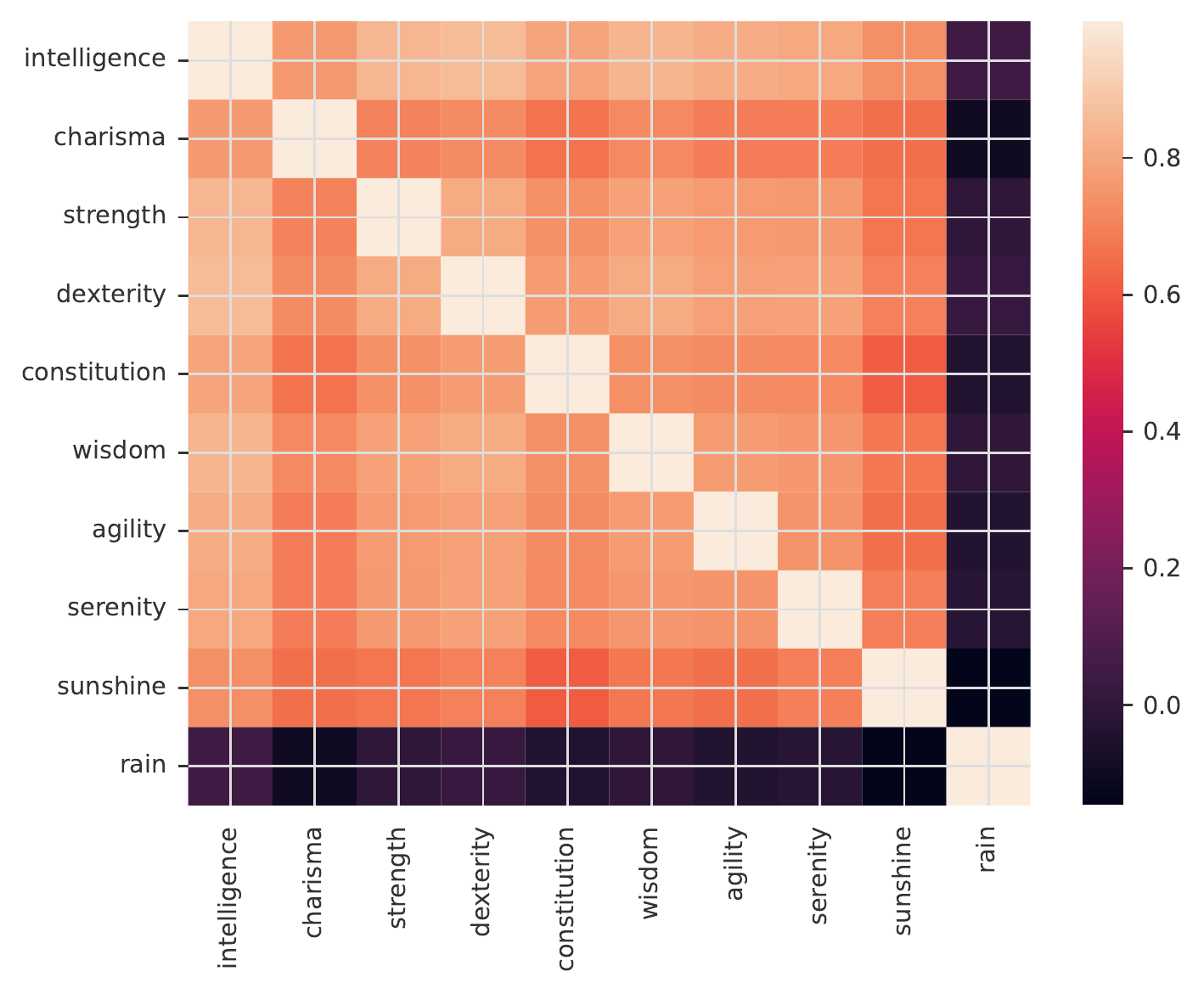}
\caption{Structural similarity of models with Jackknife feature sampling at Era 1001}
\end{subfigure}
\bigskip

\begin{subfigure}{.3\textwidth}
\centering
\includegraphics[width=.75\linewidth]{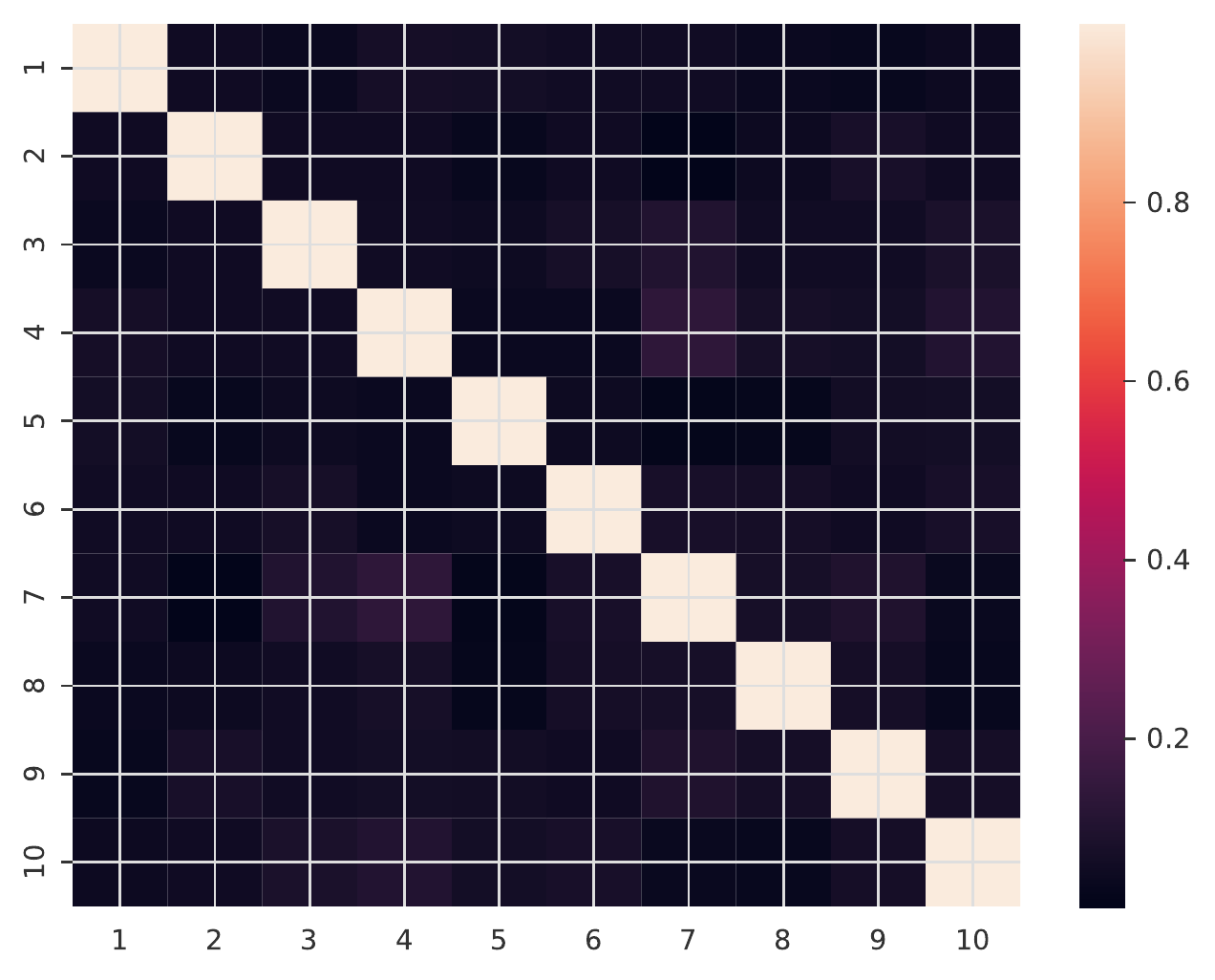}
\caption{Structural similarity of models with random feature sampling at Era 801}
\end{subfigure}
\quad
\begin{subfigure}{.3\textwidth}
\centering
\includegraphics[width=.75\linewidth]{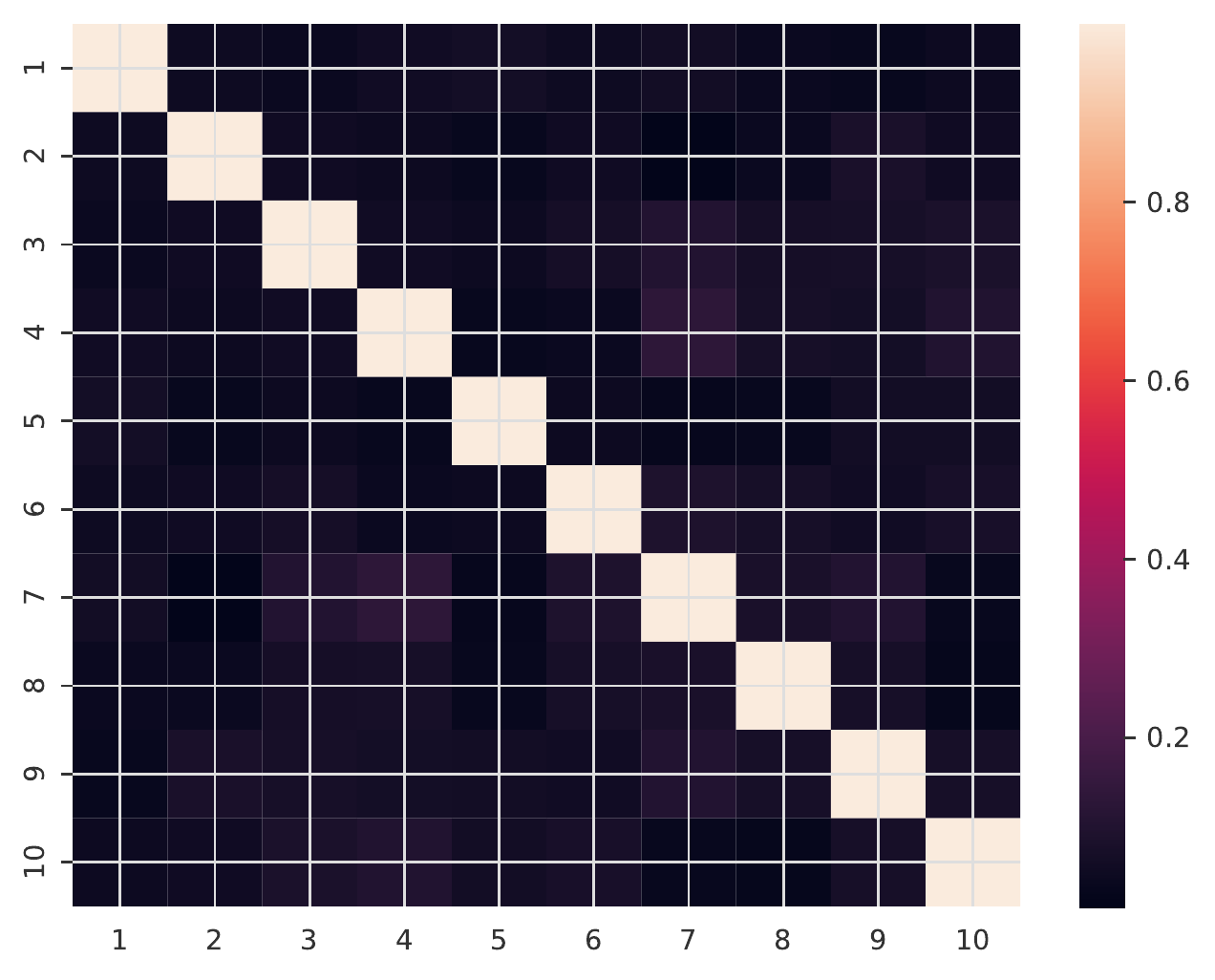}
\caption{Structural similarity of models with random feature sampling at Era 901}
\end{subfigure}
\quad
\begin{subfigure}{.3\textwidth}
\centering
\includegraphics[width=.75\linewidth]{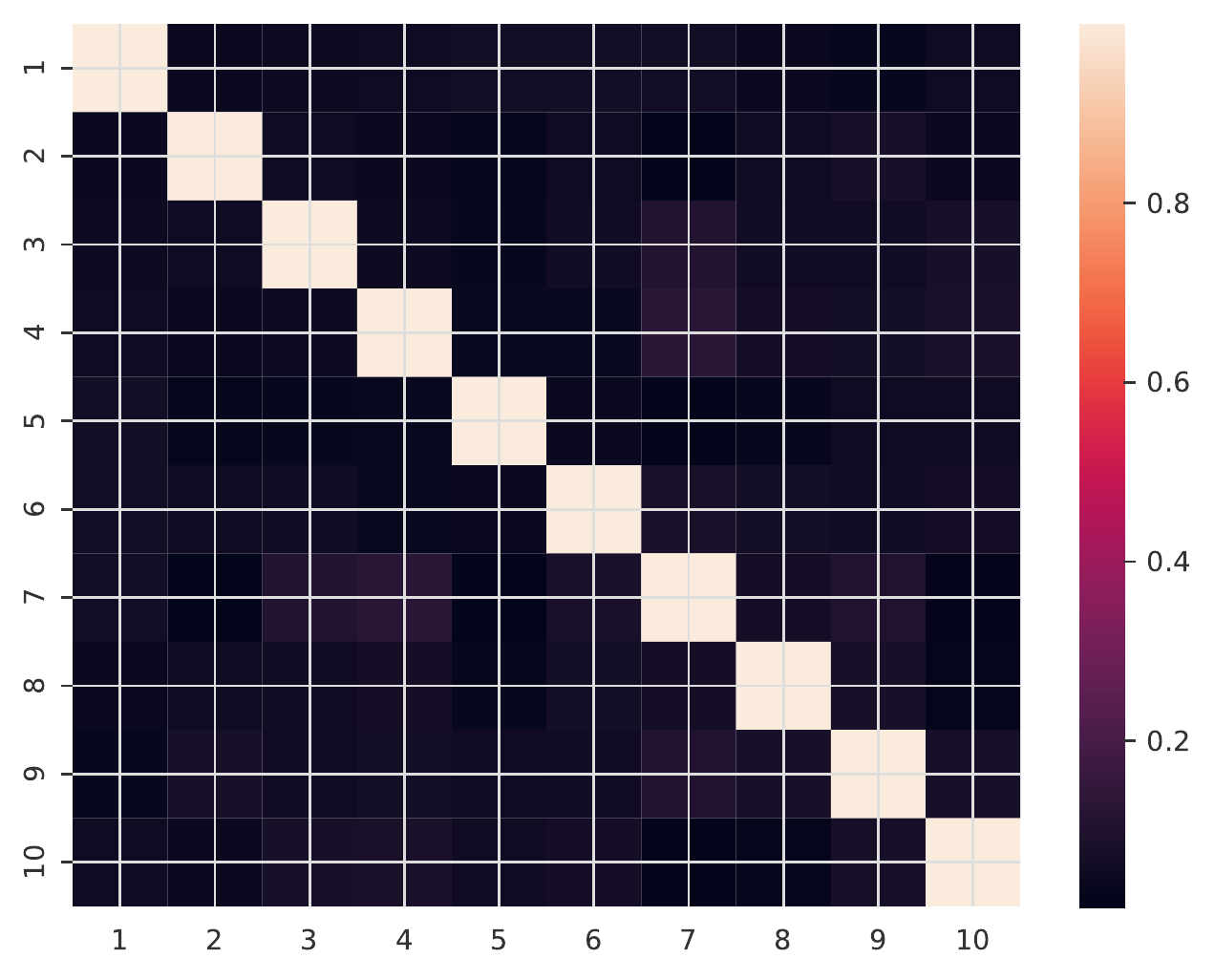}
\caption{Structural similarity of models with random feature sampling at Era 1001}
\end{subfigure}
\caption{Structural similarity of models with Jackknife and random feature sampling at Era 801, 901, 1001}
\label{fig:ModelCorr-FeatureSet}
\end{figure}

%% Structural Similarity of models 
In Figure \ref{fig:ModelCorr-FeatureSet}, we compare the correlation between the 10 Layer 1 XGBoost models obtained by Jackknife and random feature sampling. Models trained \textbf{without} rain feature set are uncorrelated to the rest of the models. Models trained by removing the other nine feature sets one at a time have correlation lower than 0.86 with average structural similarity of 0.66. Structural similarity of models obtained by Jackknife sampling is also stable across time, demonstrated by the similar heatmap representation of the models' structural similarity at Era 801,901,1001. Models obtained by random feature sampling do not have any stable structural similarity by design. 

%% Conclusion
While highly similar models offer limited diversification benefits in model ensembling, uncorrelated models created by random failed to generate better predictions. In conclusion, using domain knowledge about the data generation process, we can create models that are not over-similar to each other which can significantly improve model performances after ensembling.

%%% Structural Similarity between models 

%% Layer 2 Models 

\newpage 
\subsection{Dynamic Hedging based on model variances}
\label{section:NumeraiRain-DynamicHedging}

%% Disagreement between models 
Recent research suggests that disagreement between investors is an indicative signal of stock returns \cite{bali2023machine,}. In particular, under the Bull market, stocks that have the highest degree of disagreement between investors will under-perform relative to stocks that have the lowest degree of disagreement between investors. The opposite holds under the Bear market.

%% Dynamic Hedging 
Here, the variance between model predictions from different Layer 1 models are used as proxy of disagreement between investors. In Algorithm\ref{alg:model-disagrement}, two strategies are built based on predictions from the Layer 1 models, namely the Baseline model predictions based on the simple average and the Tail model predictions based on the standard deviation. The Tail risk model will buy stocks that the investors (Layer 1 models) disagree with each other the most and sell stocks that the investors agree with each other the most. Two approaches are then used to combine the Baseline and Tail Risk model predictions. With Static hedging, a linear combination of $60\%$ Baseline and $40\%$ Tail Risk is used for the whole test period. With Dynamic  hedging, the hedging ratio, which determines how much Tail Risk strategy is used, is adjusted according to the prevailing performances of the Tail Risk strategy. The hedging ratio is switched between two modes: (i) No hedging and (ii) $40\%$ Baseline and $60\%$ Tail Risk according to the most recent 50-week performance of the Tail Risk strategy.

\begin{figure}
    \centering
    \includegraphics[width=8cm]{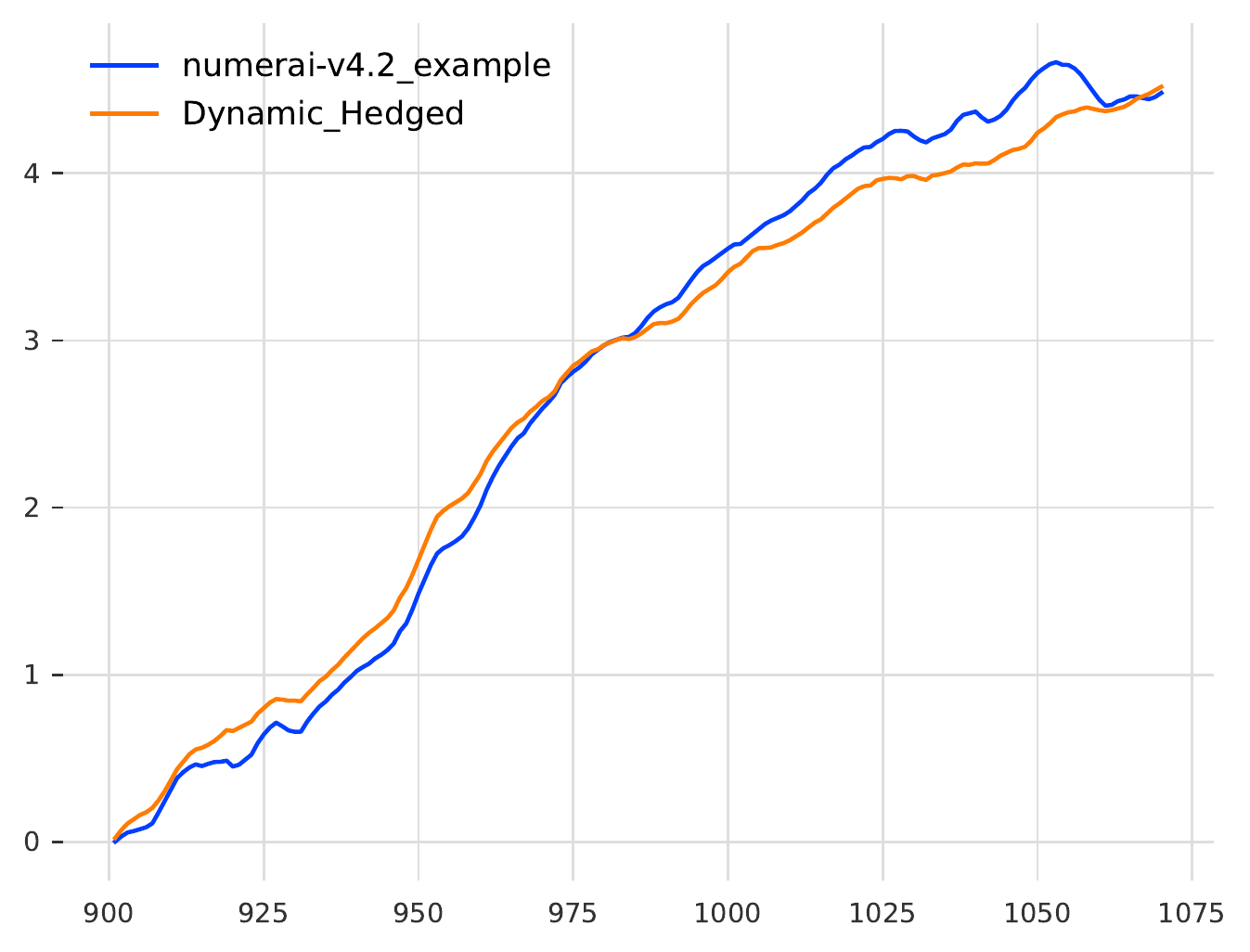}
    \caption{The portfolio return curve of the Dynamic Hedge model based on feature set Jackknife sampling and V4.2 Example Model from Era 901 to Era 1070 }
    \label{fig:Rain-DHedge-FeatureJK}
\end{figure}

%%% Dynamic Hedged Performances 
In Table \ref{table:Rain-DynamicHedge-FeatureJK}, we compare the dynamic hedged model from deep IL XGBoost ensemble using feature set Jackknife and the dynamic hedging strategy described above, with the example model provided by Numerai. To note, the example model is trained using $100 \%$ of data which is not replicated directly here due to memory limits. We used regular era sampling to train 4 models each with $25\%$ of data and then take the simple average over those. %As shown below, we have minimal loss in model performances compared to training a single model with all the data after averaging. 
The Tail Risk model works well when the Baseline model has poor performances, demonstrating its complementary nature. The Dynamic  hedged model achieves comparable Mean Corr with the example model provided by Numerai. The Sharpe ratio is improved from 0.9626 to 1.3169 while the Max Drawdown reduces from 0.2608 to 0.0237, a more than $90\%$ reduction. The portfolio return curve is much smoother for the Dynamic  hedged model, as shown in Figure \ref{fig:Rain-DHedge-FeatureJK}. 
%% Show in SI, demonstrate this does not work for random feature sampling or other ensemble ways. 
We have checked the dynamic hedging procedure on deep IL XGBoost models with random feature sampling, over different targets, learning rates and training sizes. Detailed results are shown in Tables \ref{table:Rain-DynamicHedge-FeatureRandom},\ref{table:Rain-DynamicHedge-Target},\ref{table:Rain-DynamicHedge-LearningRate},\ref{table:Rain-DynamicHedge-TrainingSize} in SI. The results show that dynamic hedged models from these ensembles are inferior to the above, as they have a lower Mean Corr and Sharpe ratio.

%% Model Disagreement 
\begin{algorithm}[hbt!]
\caption{Model Disagreement}
\label{alg:model-disagrement}
\KwIn{At era $t$: predicted values $\hat{y}_k^t \in \mathbb{R}^{N_t}$ from Layer 1 models $\mathcal{M}_k$, $1 \leq k \leq K$}
Calculate normalised predictions $\hat{r}_t$ from Layer 1 models 
\[
    \hat{r}_t = \text{rank}(\hat{y}_k^t) - 0.5 ,
\]
where the rank function calculates the percentile rank of a value within a vector, so that $-0.5 \leq \hat{r}_t \leq 0.5$. \\
Calculate Baseline model predictions (average) $\hat{y}_{mean}^t = \text{rank} \left( \frac{1}{K} \sum_{k=1}^K  \hat{y}_k^t \right) - 0.5 $ \\
Calculate Tail risk model predictions (standard deviation) $\hat{y}_{sd}^t = \text{rank} \left( \sqrt{\frac{1}{K} \sum_{k=1}^K  (\hat{y}_k^t - \hat{y}_{mean}^t)^2} \right) - 0.5$ \\
Calculate static hedged model predictions $\hat{h}_t = 0.6 \hat{y}_{mean}^t + 0.4 \hat{y}_{sd}^t$ \\
%% How to do dynamic hedging
Calculate the average recent performance of tail risk model $\bar{\rho_t}$, $ \bar{\rho_t} = \frac{1}{50} \sum_{i=t-56}^{t-7} \rho_i$, where $\rho_i$ is calculated by the scoring formula in Section \ref{formula:Numerai-Corr}.  \\
\If{$\bar{\rho_t} \geq 0$}{
    Set Hedge ratio $h=0.6$ \\
}
\Else{Set Hedge ratio $h=0$}
Calculate dynamic hedged model predictions $\hat{d}_t = (1-h) \hat{y}_{mean}^t + h \hat{y}_{sd}^t$ \\
\end{algorithm}

%% Layer 2 Models 
\begin{figure}[hbt!]
     \centering
        \subfloat[Mean Corr]{
          \includegraphics[width=15cm]{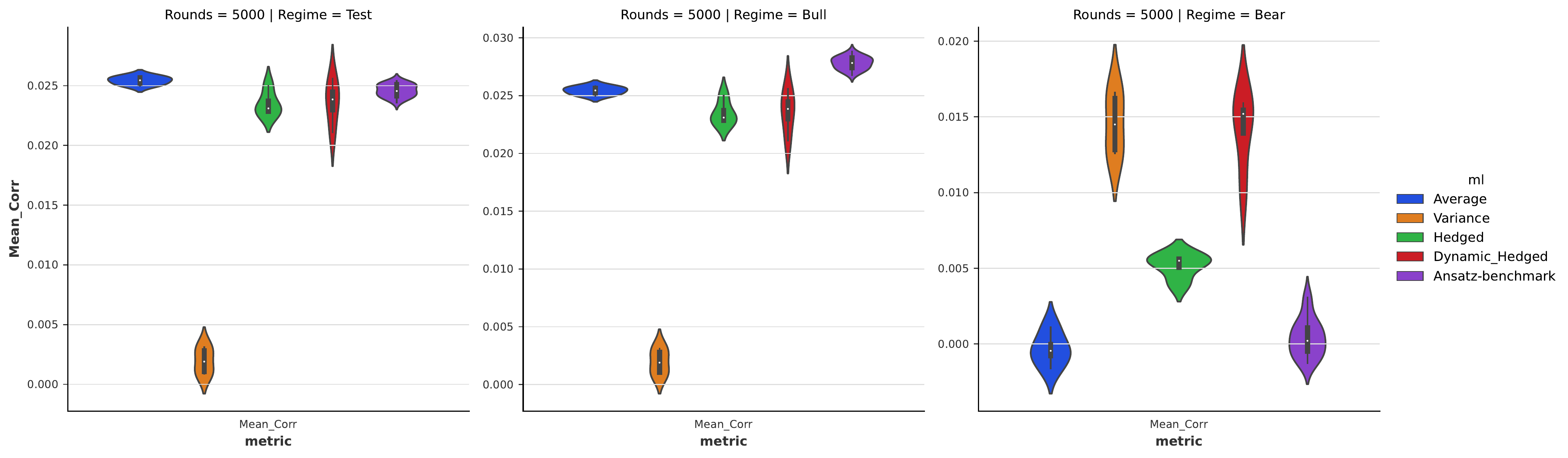}
        }
        \\
        \subfloat[Sharpe]{
          \includegraphics[width=15cm]{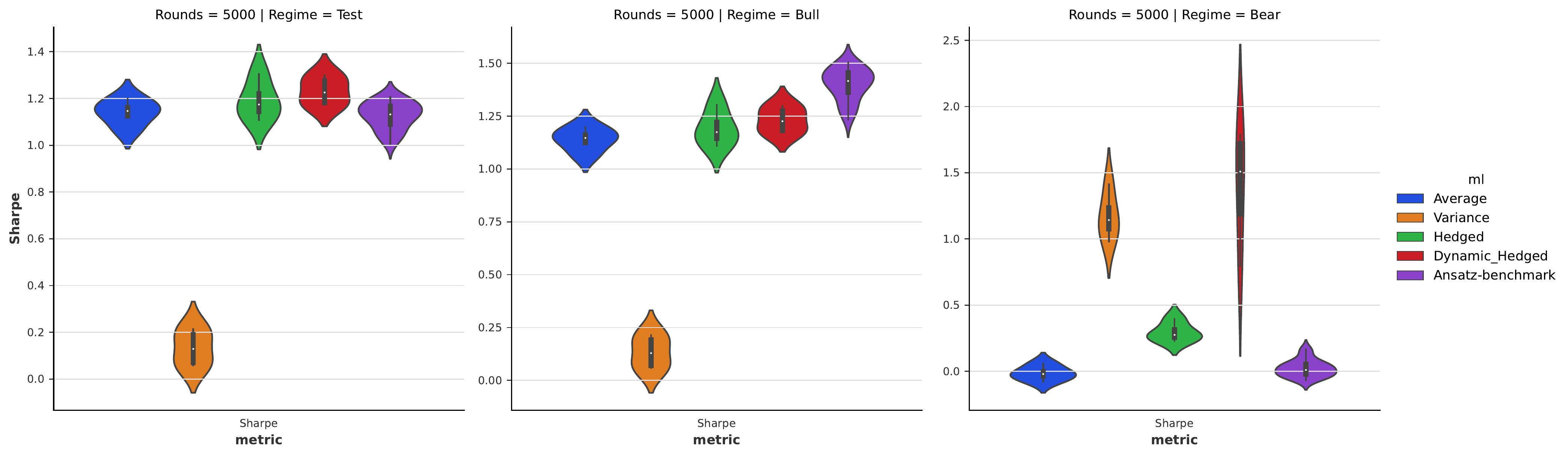}
        }
        \\
        \subfloat[Calmar]{
          \includegraphics[width=15cm]{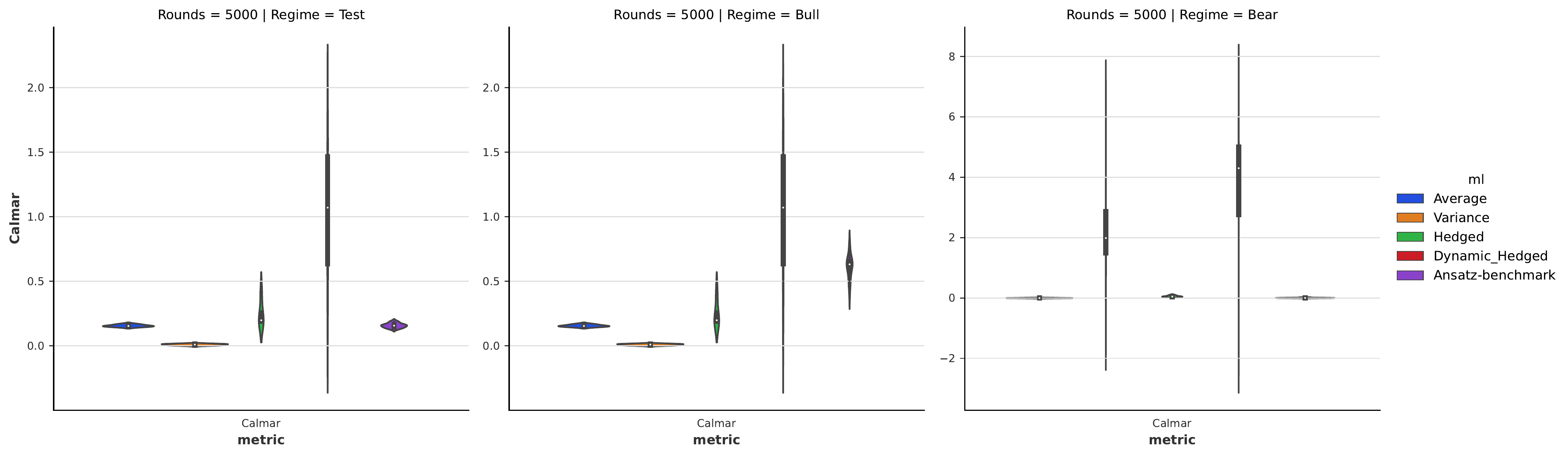}
        }
        \\
        \subfloat[Max Drawdown]{
          \includegraphics[width=15cm]{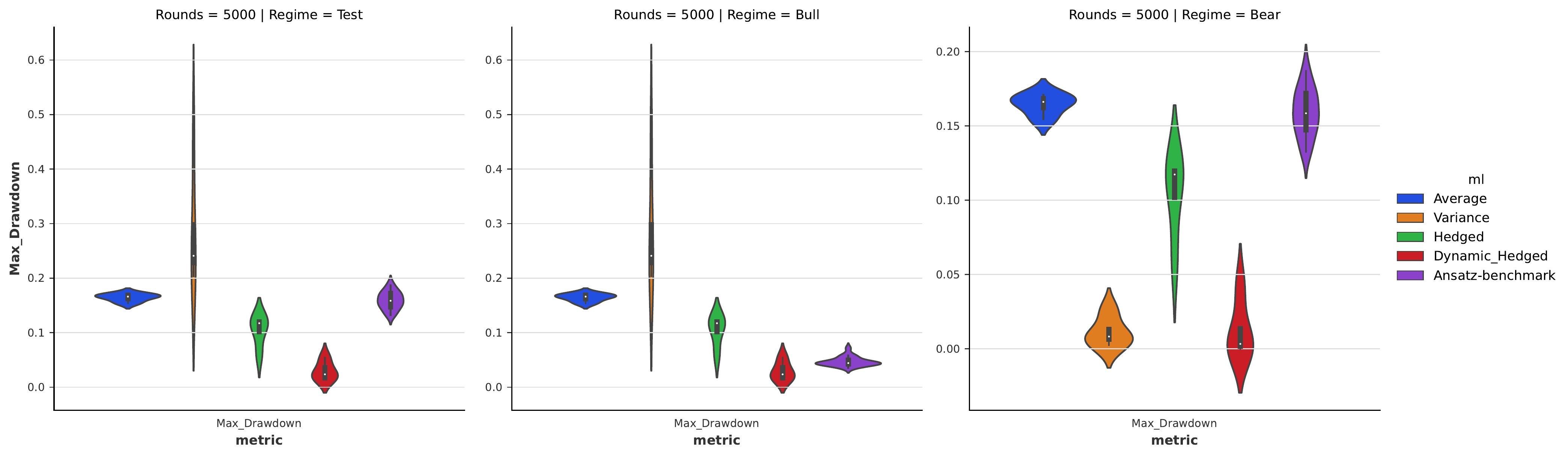}
        }
        \\
    \caption{Performances, (a) Mean Corr, (b) Sharpe ratio, (c) Calmar ratio and (d) Max Drawdown of the deep IL XGBoost models with Jackknife feature sampling and dynamic hedging under different market regimes. }
    \label{fig:Rain-IL-FeatureSampling-DynamicHedge}
\end{figure}

\begin{table}[hbt!]
\centering
\begin{tabular}{|l|c|l|l|l|}
\hline
Regime & Strategy & Mean Corr & Sharpe & Max Drawdown  \\ \hline
\multirow{5}{*}{Test}  & Example Model           & 0.0264  & 0.9626  & 0.2608  \\ \cline{2-5} 
                       & Baseline Model          & 0.0266  & 1.1725  & 0.1602  \\ \cline{2-5} 
                       & Tail Risk Model         & 0.0023  & 0.1601  & 0.2385  \\ \cline{2-5} 
                       & Static Hedged Model     & 0.0203  & 1.2159  & 0.0435  \\ \cline{2-5}
                       & Dynamic Hedged Model    & \textbf{0.0266}  & \textbf{1.3169}  & \textbf{0.0237}   \\ \hline
\multirow{5}{*}{Bull}  & Example Model           & \textbf{0.0307}  & 1.2512  & 0.0693 	 \\ \cline{2-5} 
                       & Baseline Model          & 0.0302  & \textbf{1.4780}  & 0.0396 \\ \cline{2-5} 
                       & Tail Risk Model         & 0.0006  & 0.0435  & 0.2385  \\ \cline{2-5} 
                       & Static Hedged Model     & 0.0215  & 1.3014  & 0.0309  \\ \cline{2-5}
                       & Dynamic Hedged Model    & 0.0283  & 1.3844  & \textbf{0.0237}  \\ \hline
\multirow{5}{*}{Bear}  & Example Model           & -0.0060 & -0.2306 & 0.2608  \\ \cline{2-5} 
                       & Baseline Model          & -0.0002 & -0.0080 & 0.1602  \\ \cline{2-5} 
                       & Tail Risk Model         & \textbf{0.0153}  & \textbf{1.2929}  & \textbf{0.0000}  \\ \cline{2-5} 
                       & Static Hedged Model     & 0.0109  & 0.7489  & 0.0435  \\ \cline{2-5}
                       & Dynamic Hedged Model    & 0.0137  & 1.1500  & 0.0220   \\ \hline
\end{tabular}
\caption{Performances of Dynamic Hedged deep IL XGBoost ensemble model based on feature set Jackknife sampling and V4.2 Example Model from Era 901 to Era 1070 under different market regimes. }
\label{table:Rain-DynamicHedge-FeatureJK}
\end{table}

Model ensembles created based on Jackknife feature set sampling use knowledge about the dataset and thus offer a better approximation of disagreement between investors. Therefore, the Tail Risk model created based on the variance between models trained with Jackknife feature set sampling is the most effective hedging strategy in the Bear market.

\newpage 
\section{Conclusion} 

%% Summary
In this study, both traditional tabular and factor-timing models have been studied for the IL problem on the temporal tabular dataset from Numerai. Traditional tabular models, if retrained regularly can adapt to distribution shifts in data. On the other hand, factor-timing models failed to adapt to distribution shifts. 

%% GBDT is a robust choice
\paragraph{GBDT are robust ML models}

We found that GBDT models performed best for the Numerai datasets, agreeing with the findings of \cite{mcelfresh2023neural,}, which demonstrate the robust and superior performances of GBDT models on large datasets. This is also partly due to the nature of features being binned values from continuous underlying measures, which favours models based on decision rules rather than regression. With suitable designs of the training process, such as a slow learning rate with a large number of boosting rounds, we can train XGBoost models with good performance, slowly converging to the theoretical optimal.

%% Robustness Hyperparameters 
GBDT models are also highly scalable and have robust performance over slightly perturbed hyperparameters. The larger the GBDT model, the smaller the effect of model hyperparameters on the learning process and model performances. GBDT models are numerically stable, without the convergence issues that commonly plague the training of neural networks. 
%For example, vanishing gradient is a common issue in training MLPs therefore different variations of activation functions are proposed. 

%GBDT also less suspect to variation in model training randomness due to ensemble nature of model. Any large enough GBDT models will have the random effects of feature and data selection averaged out. Unlike neural networks, where model performances highly depend on the initial random weights, as suggested in the Lottery Ticket Hypothesis \cite{frankle2018the,}. 

\paragraph{Feature and Data Sampling}

Data management and forgetting mechanism is an integrated part of an IL pipeline \cite{Gama14} to build robust prediction models on a data stream. The impact of data sampling methods is usually over-looked in most quantitative finance research and even in hedge funds \cite{hoffstein2020rebalance}. Data and feature sampling methods can have significant effects on model performances. 

%% Era Sampling 
Removing data with targets equal to the Median value can reduce computational time by half without significant loss to model performances. This demonstrates the point that well designed sampling procedures can be used to filter data for effective model training. 

%% Feature Sampling 
Feature sampling can also be used to increase diversity of models by enforcing constraints on features that are allowed to be used or interact in a model. Using feature set group labels can create feature sets that are more efficient in creating diverse model ensembles than random sampling. 

%% Training Size 
In all incremental problems we encounter the stability-plasticity dilemma \cite{carpenter1993normal,}, which is the trade-off between the ability of ML models to adapt to new patterns and preserve existing knowledge. It is not known in advance which data sampling method will have the optimal performance and therefore ensembling models with different training sizes with equal weights is often a robust strategy when there are no additional information to decide how much data to use for model training. 

%% Model retrain 
Retraining benchmark models regularly can improve performances significantly compared to using the same model without updating. In general, model performances improves with the frequency of retrain  but the requirements on computational resources also increase. Therefore, trade-offs between computational costs and the marginal gain in model performances are made for practical IL systems.

\paragraph{Learning rates and model complexity}

%% Ansatz Formula 
We derive an Ansatz formula to determine the learning rate $L$ for a GBDT model given a fixed number of boosting rounds $B$. We show the formula is optimal for our benchmark GBDT models over a wide range of sizes, from $B=1000$ to $B=50000$. 

%% Learning Rate Ensemble 
Combining models with different degree of complexity, created with different number of boosting rounds with learning rates derived using the Ansatz formula, can improve model performances compared to models using a fixed number of boosting rounds. The optimal model complexity is regime dependent and therefore using deep IL techniques to dynamically combine model predictions can reduce downside risks in models.

\paragraph{Connection with Model Stacking/Selection}

Stacking is a simple but highly effective technique to combine different ML model predictions. The concept of stacking is not limited to machine learning. In finance, portfolio optimisation is studied in detail to improve investment returns, where a convex optimisation is solved at each time step to find the linear combination of assets or strategies that maximise risk-adjusted return.
Under the IL framework, model stacking can be performed dynamically. Here, we combine the predicted rankings from different ML models at each era with different weights. Instead of considering model stacking as a \textit{separate} step to model training, model stacking can be incorporated as an integrated part in the IL framework, as an extra layer in the IL model.

\paragraph{Hedging against regime changes}

Using the variance between models within the ensemble as a signal, tail risk strategy can be created to hedge the baseline prediction strategy based on simple average of different component models. Regime changes in data can be captured by uncertainty of model predictions, as a higher variance between models within the ensemble suggests a lower confidence of the predictions. Therefore, prediction based on variance would perform well under regime changes, such as the Bear market period identified in this study. The best performance is achieved when the component models are trained using different combinations of feature subsets based on economic knowledge about the dataset. In this case, the variance between models are the most informative approximation of disagreement between investors in the stock market.

\paragraph{Further Work}

In most practical applications, \textit{multiple} machine learning methods are used together to create an ensemble prediction. The IL model presented in this paper provides a comprehensive way to integrate different ML models in a consistent and systematic way to create point-in-time predictions. With a multi-layer structure and modularised design within each layer, the deep IL model can flexibly model datasets with different complexities and structures. Further work can be done by integrating different deep tabular models into the model and bench-marking different machine learning methods under the IL framework.

Within our incremental learning framework, we retrain each XGBoost model from scratch without using any information from previous ones. Currently, new methods \cite{WANG2022288,liu2020diverse,} have been developed which adapt towards concept drift in data by adding a suitable amount of trees to existing GBDT models. Different approaches, such as reusing a certain amount of base learners (trees) from previous trained GBDT models or updating the weights of trees dynamically depending on the severity of concept drift can be explored in future work.

We only consider the simplest form of deep learning models, MLP in this paper. Recent research suggests regularisation techniques \cite{Arlind21,} can improve performances of neural networks models over a wide range of network architecture. Further work can be done to investigate if careful design of the model training process with suitable regularisation can improve the scalability and model performances.

\section{Acknowledgements}
This work was supported in part by the Wellcome Trust under Grant 108908/B/15/Z and by the EPSRC under grant EP/N014529/1 funding the EPSRC Centre for Mathematics of Precision Healthcare at Imperial. MB also acknowledges support by the Nuffield Foundation under the project ``The Future of Work and Well-being: The Pissarides Review". We thank Numerai GP, LLC for providing the datasets used in the study.

\section{Data and Code Availability} 
The data and code used in this paper are available at \url{https://github.com/barahona-research-group/THOR-2}.

%Bibliography
\newpage
\printbibliography

\newpage 
\section{Supplementary Information}

\subsection{Algorithms of different benchmark machine learning models studied} 
\label{section:NumeraiSunshine-tsalgos}

\subsubsection{Signature Transforms} 

Signature transforms are applied on continuous paths. A path $X$ is defined as a continuous function from a finite interval $[a,b]$ to $\mathbb{R}^d$ with $d$ the dimension of the path. $X$ can be parameterised in coordinate form as $X_t = (X_t^1,X_t^2,\dots,X_t^d)$ with each $X_t^i$ being a single dimensional path. 

% Iterated Integrals 
For each index $ 1 \leq i \leq d$, the increment of $i$-th coordinate of path at time $t \in [a,b]$, $S(X)_{a,t}^i$, is defined as 
\begin{equation*}
    S(X)_{a,t}^i = \int_{a<s<t} \mathrm{d}X_s^i = X_t^i - X_a^i
\end{equation*}
As $S(X)_{a,\cdot}^i$ is also a real-valued path, the integrals can be calculated iteratively. A $k$-fold iterated integral of $X$ along the indices $i_1,\dots,i_k$ is defined as 
\begin{equation*}
    S(X)_{a,t}^{i_1,\dots,i_k} = \int_{a<t_k<t} \dots \int_{a<t_1<t_2}   \mathrm{d}X_{t_1}^{i_1}  \dots \mathrm{d}X_{t_k}^{i_k} 
\end{equation*}

% Definition of Signature 
The Signature of a path $X: [a,b] \mapsto \mathbb{R}^d$, denoted by $S(X)_{a,b}$, is defined as the infinite series of all iterated integrals of $X$, which can be represented as follows 
\begin{align*}
    S(X)_{a,b} &= (1, S(X)_{a,b}^1, \dots, S(X)_{a,b}^d,  S(X)_{a,b}^{1,1}, \dots ) \\
               &= \bigoplus_{n=1}^{\infty} S(X)_{a,b}^n
\end{align*}

An alternative definition of signature as the response of an exponential nonlinear system is given in \cite{Terry22}. 

% Log Signature 
Log Signature can be computed by taking the logarithm on the formal power series of Signature. No information is lost as it is possible to recover the (original) Signature from Log Signature by taking the exponential \cite{Chevyrev16,Terry22}. Log Signature provides a more compact representation of the time series than Signature. 
\begin{equation*}
    log S(X)_{a,b} =  \bigoplus_{n=1}^{\infty}  \frac{(-1)^{(n-1)}}{n} S(X)_{a,b}^{\bigotimes n} 
\end{equation*}

%%% Theoretical properties of signatures 
%%% Multiplicative Functional 
%%% Universal Property of Signature in predictions (flexible) 
Signatures can be computed efficiently using the Python package signatory \cite{kidger2021signatory}. The signature is a multiplicative functional in which Chen's identity holds. This allows quick computation of signatures on overlapping slices in a path. Signatures provide a unique representation of a path which is invariant under reparameterisation \cite{Chevyrev16, Terry22}. Rough Path Theory suggests the signature of a path is a good candidate set of linear functionals which captures the aspects of the data necessary for forecasting. In particular, continuous functions of paths are approximately linear on signatures \cite{pmlr-v130-lemercier21a}. This can be considered as a version of the universal approximation theorem \cite{cybenko1989approximation} for signature transforms.

%% Interpretation of Signatures
% Level 1 Signature corresponds to the difference of two series (tail-head). When log price series are given as input, it corresponds to log return 
% Basic Statistical features can be recovered from signatures 

\paragraph{Limitations for signature transforms in high dimensional datasets}
The number of signatures and log-signatures increases exponentially with the number of channels. For time series with a large number of channels, random sampling can be applied to select a small number ($5 < N < 20$) of time series with replacement from the original time series on which signature transforms are applied. Random sampling can be repeated a given number of times to generate representative features of the whole multivariate time series. Similar ideas are considered in \cite{James20}, in which random projections on the high dimensional time series are used to reduce dimensionality before applying signature transforms.  

%% Lookback window
Let $\tilde{X}$ be a multivariate time series with $T$ time-steps and $d$ dimensional features, denote $\tilde{X}_s \in \mathbb{R}^d$ be the observation of the time series at timestep $s$. Procedure \ref{alg:lookback} can be used to obtain paths, which are slices of time series with different lookback windows. Random Signature transforms \ref{alg:randomsig} can then be used to compute the signature of the path, which summarises the information of the time series. 

\begin{algorithm}[hbt!]
\caption{Lookback Window Slicing}\label{alg:lookback}
\KwIn{time series $\tilde{X} \in \mathbb{R}^{T \times d}$, lookback $\delta$}
\KwOut{paths $X_t \in \mathbb{R}^{t \times d}$}
\For{$1 \leq t \leq T$}{
    Set start of slice $s_1 = \max(1, t - \delta)$ \;
    Set end of slice $s_2 = t$ \;
    $X_t = (\tilde{X}_{s_1},\tilde{X}_{s_1+1}, \dots, \tilde{X}_{s_2}) $ \;
}
\end{algorithm}

%% Random Signature Transform Algorithms 
\begin{algorithm}[hbt!]
\caption{Random Signature Transform}\label{alg:randomsig}
\KwIn{path $X_t \in \mathbb{R}^{t \times d}$, level of signature $L$, number of channels $C$, number of feature sets $p$,} 
where $d > C$ \;
\KwOut{log signatures $s_t \in \mathbb{R}^{pN}$ }
Define $N= \text{Number of Log Signatures of a path with } C \text{ channels up to level }  L $ \;
\For{$1 \leq i \leq p$}{
    Sample with replacement $C$ Columns from $X_t$, defined as $\tilde{X}^i_t$ \;
    Compute the Log Signatures $s_t^i \in \mathbb{R}^N$ of $\tilde{X}^i_t$ \;
}
Combine all log signatures $s_t = (s_t^1, \dots, s_t^p)$ 
\end{algorithm}

\subsubsection{Random Fourier Transforms} 
Random Fourier Transforms are used in \cite{kelly2022virtue} to model the return of financial price time series. They can be applied to the feature performance time series at each time step as in Algorithm \ref{alg:rft}. The key idea is to approximate a mixture model of Gaussian kernels with trigonometric functions \cite{Sutherland15}.  

\begin{comment}
A price time series $P_t \in \mathbb{R}^T$ is first transformed into a return series $X_t \in \mathbb{R}^{T \times d} $ by taking the percentage change of price at different lookback intervals. Let $\delta_1, \dots, \delta_d \in \mathbb{N}$ be a given a list of lookback intervals, 
\begin{equation*}
    X_{t,\delta_i} = \frac{P_t - P_{t-\delta_i}}{P_{t-\delta_i}}
\end{equation*}
where $1 \leq t \leq T$ and $1 \leq i \leq d$     
\end{comment}

%% Random Fourier Transform Algorithms 
\begin{algorithm}[hbt!]
\caption{Random Fourier Transform \cite{kelly2022virtue}}\label{alg:rft}
\KwIn{signal vector $x_t \in \mathbb{R}^d$, number of features sets $p$, }
\KwOut{transformed vector $s_t \in \mathbb{R}^{14p}$ }
\For{$1 \leq i \leq p$}{
    Sample $w_i \sim \mathcal{N}(0, I_{d\times d})$ \;
    Set grid $(\gamma_i)_{i=1}^14 = (0.1, 0.5, 1, 2, 4, 8, 16, 0.1, 0.5, 1, 2, 4, 8, 16)$ \;
    \For{$1 \leq j \leq 7$}{
        Set $ s_{t,14i+j} = \frac{1}{\sqrt{7p}} \sin(\gamma_j w_i^T x_t)$
    }
    \For{$8 \leq j \leq 14$}{
        Set $ s_{t,14i+j} = \frac{1}{\sqrt{7p}} \cos(\gamma_j w_i^T x_t)$
    }
}
\end{algorithm}

\subsubsection{Gradient Boosting Models} 
\label{section:NumeraiSunshine-XGBoost}

\paragraph{XGBoost Implementation}

XGBoost \cite{XGBoost} modifies the above "standard" gradient boosting algorithms with approximation algorithms in split finding. Instead of finding the best(exact) split by searching over all possible split points on all the features, a histogram is constructed where splitting is based on percentiles of features. XGBoost supports two different growth policies for the leaf nodes, where nodes closest to the root are split (depth-wise) or the nodes with the highest change of loss function are split (loss-guide). The default tree-growing policy is depth-wise and performs better in most benchmark studies. XGBoost also supports L1 and L2 regularisation of model weights. Other standard model regularisation techniques such as limiting the maximum depth of trees and the minimum number of data samples in a leaf node are also supported.

\paragraph{Model Snapshots}

For GBDT models, it is easy to extract model snapshots, defined as the model parameters captured at the different parts of the training process. This can be done without any additional memory costs at inference.

Model snapshots of a GBDT model can be obtained as follows. The snapshots start with the first tree and the number of trees to be used is set to be $10\%,20\%,\dots,100\%$ of the number of boosting rounds. This trivially gives 10 different GBDT models representing different model complexities from a \textit{single} model. 

%To avoid early convergence of the learning process, the learning rate can be set to a small value. 

\subsubsection{Deep Learning Models} 
\label{section:NumeraiSunshine-DL}

\paragraph{Training process}

PyTorch Lightning \cite{Falcon_PyTorch_Lightning_2019} is used to build neural network models as it supports modular design and allows rapid prototyping. %The learning rate of neural networks is found by the Learning Rate Finder over a parameter grid of $(1e-3,0.1)$. 
Early stopping is applied based on the validation set based on a given number of rounds (patience). The batch size of the neural network is set to be the size of each era. The Adam optimiser in PyTorch with the default settings for the learning rate schedule is used. L2-regularisation on the model weights is also applied. Gradient clipping is also be applied to prevent the gradient explosion problem for correlation-based loss functions. 

\paragraph{Architecture}

The network architecture is a sequential neural network with two parts, firstly a "Feature Engineering" part which consists of multiple feature engineering blocks and then the "funnel" part which is a standard MLP with decreasing layer sizes. 

Each feature engineering block has an Auto-Encoder-like structure, where the number of features is unchanged after passing each block. Setting a neuron scale ratio of less than 1 corresponds to the case of introducing a bottleneck to the network architecture so as to learn a latent representation of data in a lower dimensional space. %Setting a neuron scale ratio greater than 1 corresponds to the case of introducing random combinations of features which are then refined during the model training process.
Algorithm \ref{alg:encoding} shows how to create the feature engineering part of the network.

Funnel architecture, as used in \cite{Zimmer_Auto-PyTorch_Tabular_Multi-Fidelity_2021} is an effective way to define the neuron sizes in a network for different input feature sizes. Algorithm \ref{alg:funnel} shows how to create the funnel part of the network. 

Each Linear layer is followed by a ReLU activation layer and dropout layer where $10\%$ of weights are randomly zeroed. %Batch Normalisation is not used. 

\begin{definition}[Linear Layer] ~\\
    A Linear Layer $(M_1,M_2)$ within a sequential neural network is a transformation $X_2 = f(X_1)$ with input tensor $X_1 \in \mathbb{R}^{N \times M_1}$ and output tensor $X_2 \in \mathbb{R}^{N \times M_2}$ where $N$ is the batch size of data. For a given non-linear activation function $\sigma(\cdot)$ such as ReLU, let $W \in \mathbb{R}^{M_2 \times M_1}$ be the weight tensor and $b \in \mathbb{R}^{M_2}$ be the bias tensor to be learnt in the training process, the Linear layer is defined as 
    \begin{equation*}
        f(X_1) = \sigma(X_1 W^T + b)
    \end{equation*}
    % where the addition of bias tensor is performed by broadcast multiplication 
\end{definition}

\begin{algorithm}[hbt!]
\caption{Feature Engineering network architecture}
\label{alg:encoding}
\KwIn{Input feature size $M$, Number of encoding layers $L$, neuron scale ratio $r$}
\KwOut{Sequential Feature Engineering Network Architecture}
\For{$1 \leq l \leq L$}{
    Encoding Layer $l$: Linear layer $(M,M*r)$ \\
    Decoding Layer $l$: Linear layer $(M*r,M)$ 
}
\end{algorithm}

\begin{algorithm}[hbt!]
\caption{Funnel network architecture}
\label{alg:funnel}
\KwIn{Input feature size $M$, Output feature size $K$, Number of intermediate layers $L$, neuron scale ratio $r$}
\KwOut{Sequential Funnel Network Architecture}
Input Layer: Linear layer (M, $M*r$) \\
\For{$1 \leq l \leq L$}{
    Intermediate Layer $l$: Linear layer $(M*r^{l}, M*r^{l+1})$ 
}
Output Layer: Linear layer $(M*r^{L+1},K)$ \\
\end{algorithm}

\paragraph{Feature projection and Loss Function} 

Pearson correlation calculated on the whole \textit{era} of target and predictions is used as the loss function at each training epoch. Feature projection, if needed, can be applied from the outputs of network architecture. The neutralised predictions are further standardised to zero mean and unit norm. The negative Pearson correlation of the standardised predictions and targets is then used as the loss function to train the network parameters.

\newpage 

\subsection{Creating benchmark XGBoost models}
\label{section:NumeraiRain-benchmark}

Example models provided by Numerai are trained under different conditions with the models we presented here. In particular, random seeds and data sampling schemes are not reported from Numerai, such that we cannot replicate the results. 

Random seeds are unwanted sources of variability that needs to be controlled \cite{gundersen2023sources}. Models trained with all the data will have better performances by design and higher computational costs. Therefore we need to use the same data sampling schemes to fairly train models under the same conditions except that ones we want to change. For GBDT models, model performances also increases with the number of boosting rounds in general, therefore we also need to use an equal amount of rounds to train the models. 

Therefore, we create benchmark models using the Ansatz hyperparameters and hyperparameters from Numerai under the \textbf{same} random seeds and the \textbf{same} data sampling scheme. All the 2132 features are used in training. The target 'target-cyrus-v4-20' are used to train all the models. 

We use the data sampling scheme $S_2$ described in Section \ref{section:NumeraiRain-erasample} which keep observations with target not equal to the Median value (0.5) in training and trained each model using 25\% of data eras regularly sampled. We then obtain 4 benchmark models for each set of hyperparameters (Ansatz and Numerai). The training size of models is fixed to 600, with the training data starting at Era 201.

The Ansatz hyperparameters are found by grid search on different XGBoost models hyperparameters using a subset of features described in Section \ref{section:NumeraiSunshine-HyperOpt}. The key hyperparameters optimised are: Max Depth: 4, Data Sampling per tree: 0.75, Feature Sampling per tree: 0.75. The learning rate $L$ is given by formula $L= \frac{50}{B}$. 

Numerai provided the following hyperparameters \cite{numerai-hyperparameters,} for their example model based on LightGBM \cite{LightGBM}. The key hyperparameters are: Max Depth: 6, Data Sampling per tree: 1.0, Feature Sampling per tree: 0.1. The recommended the number of boosting rounds $B=30000$ with learning rates $L=0.001$, which is interpreted as using the formula $L=\frac{30}{B}$.

%% Computational costs 
As Ansatz hyperparameters uses more shallow trees to build trees than Numerai, it has a much lower memory consumption. On average, for a fixed number of boosting rounds $B$, the memory consumption of models with Ansatz hyperparameters are only around $30\%-35\%$ of that of models with Numerai hyperparameters. Computational time is a lower as fewer decision rules are learnt in each. On average, for a fixed number of boosting rounds $B$, the running time of models with Ansatz hyperparameters are only around $70\%-80\%$ of that of models with Numerai hyperparameters.

%% Memory for B=5000 (Numerai: 24885588, Ansatz:8566028)
%% Time for B=5000 (Numerai: 4 min 37s, Ansatz: 3 min 44s)
%% Time for B=25000 (Numerai: 22min 47s, Ansatz: 18 min -46s)

\begin{figure}[hbt!]
    \centering
     \includegraphics[width=12.5cm]{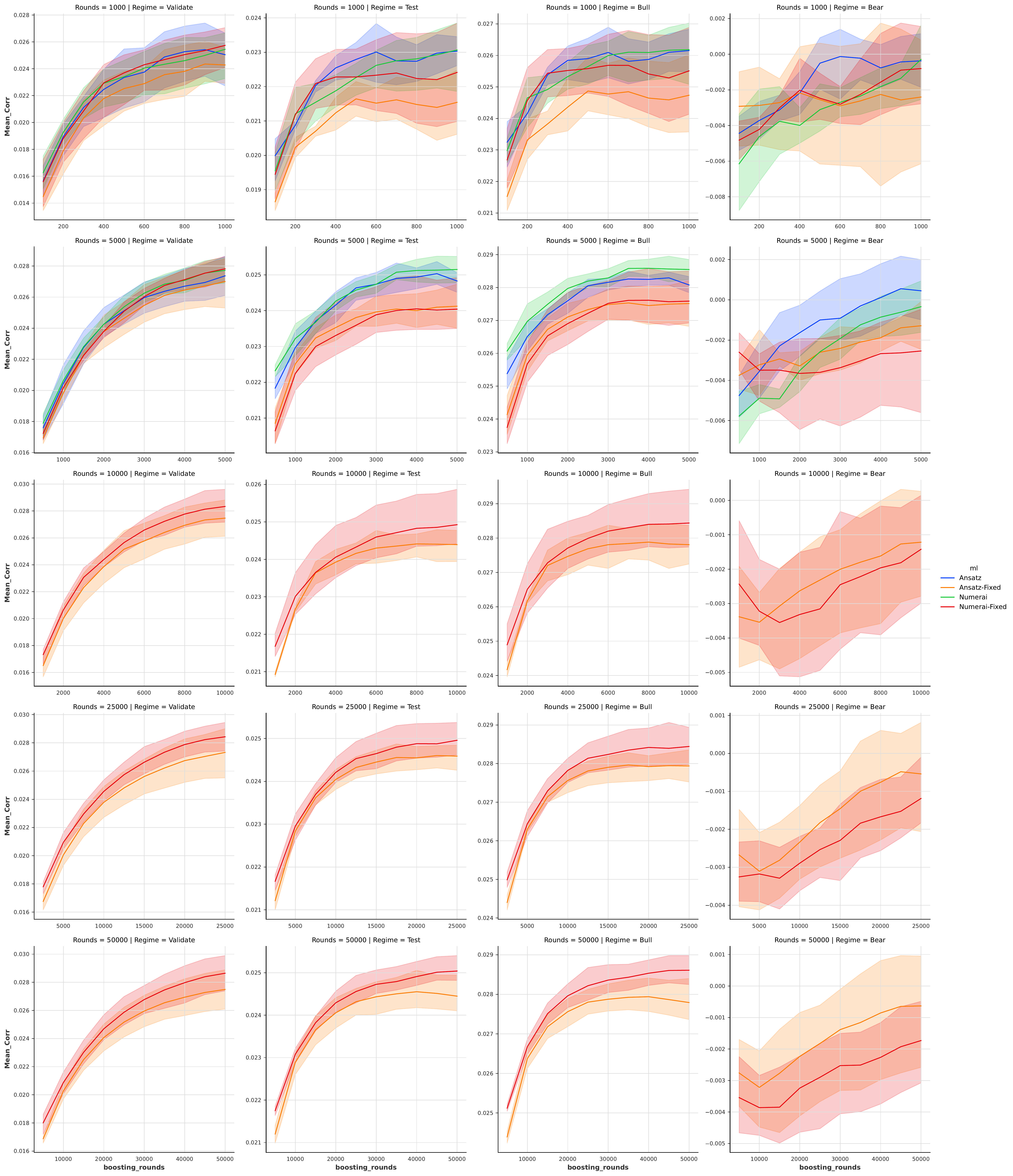}
    \caption{Learning curves of benchmark XGBoost models with different number of boosting rounds $B=1000,5000,10000,25000,50000$ for risk metric Mean Corr under different market regimes}
    \label{fig:Rain-Benchmark-LCurve-MeanCorr}
\end{figure}

\begin{figure}[hbt!]
    \centering
     \includegraphics[width=12.5cm]{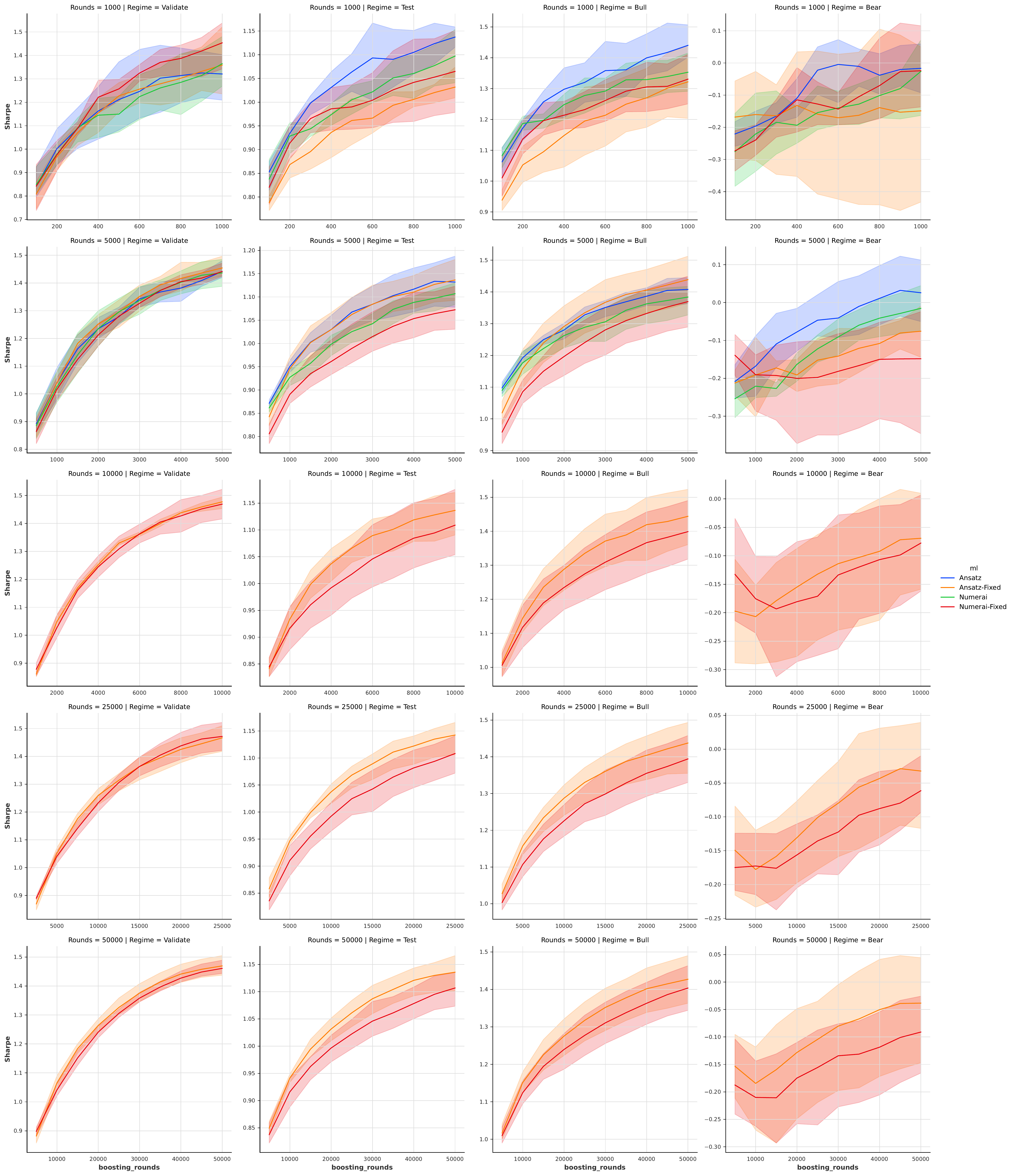}
    \caption{Learning curves of benchmark XGBoost models with different number of boosting rounds $B=1000,5000,10000,25000,50000$ for risk metric Sharpe under different market regimes}
    \label{fig:Rain-Benchmark-LCurve-Sharpe}
\end{figure}

\begin{figure}[hbt!]
    \centering
     \includegraphics[width=12.5cm]{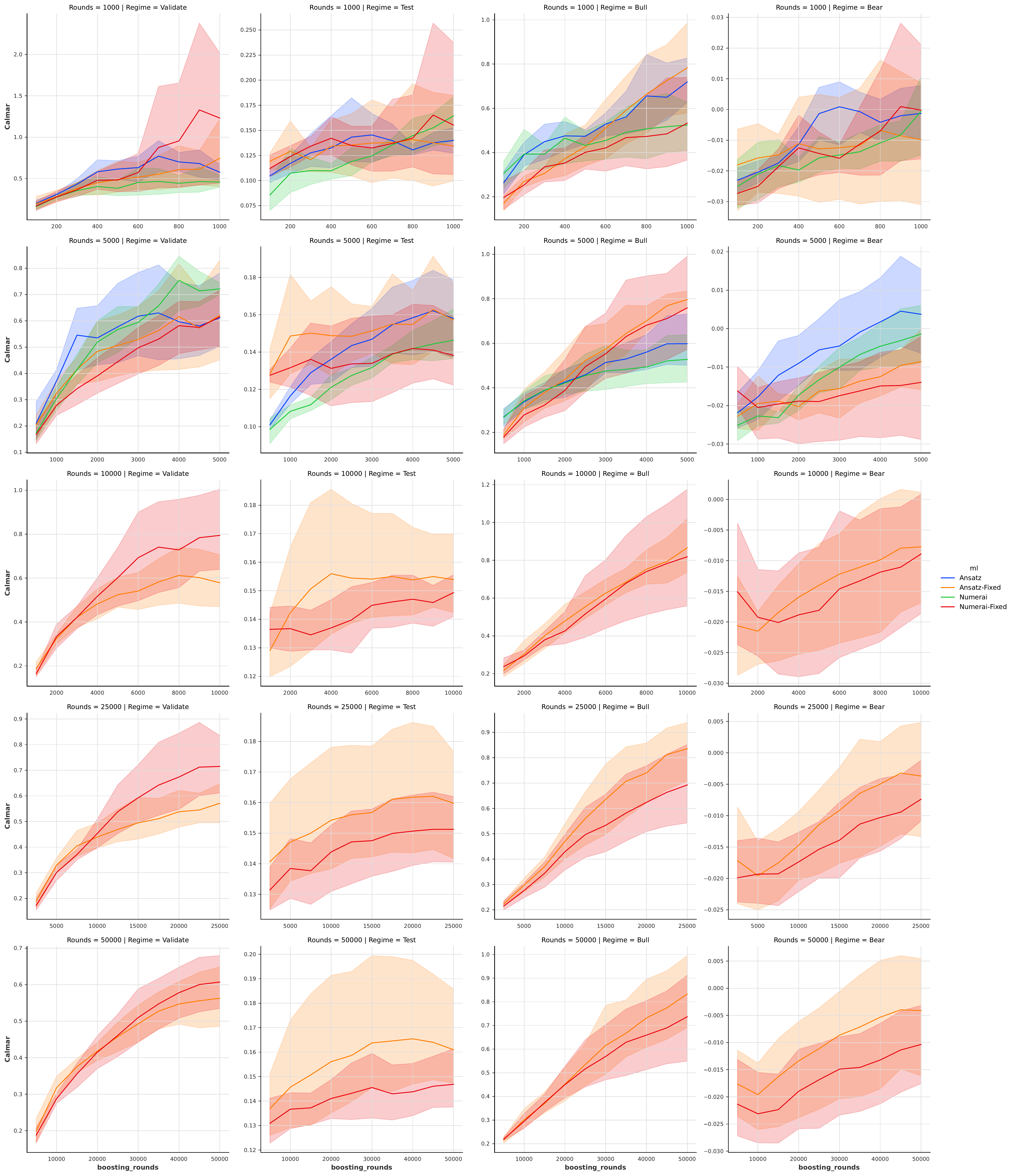}
    \caption{Learning curves of benchmark XGBoost models with different number of boosting rounds $B=1000,5000,10000,25000,50000$ for risk metric Calmar under different market regimes}
    \label{fig:Rain-Benchmark-LCurve-Calmar}
\end{figure}

%% Learning curves 
In Figures \ref{fig:Rain-Benchmark-LCurve-MeanCorr}, \ref{fig:Rain-Benchmark-LCurve-Sharpe}, and \ref{fig:Rain-Benchmark-LCurve-Calmar} in SI, learning curves for Mean Corr and Sharpe ratio of the benchmark XGBoost models are shown under different market regimes.

% Models using the Numerai hyperparameters achieved a higher Mean Corr and Sharpe ratio than models with Ansatz hyperparameters over different stages of process for models with $B=25000$ and $B=50000$. However, consider models with $B=5000$, a simple change from using only the most recent 600 eras of data eras to using all the data from Era 201 in training can significantly improve the performances of models with Ansatz hyperparameters. It suggests the need use the same data for model training in hyperparameter optimisation in order to fairly compare different hyperparameter settings. 

\subsubsection{Sampling data within an era}
\label{section:NumeraiRain-erasample}

In this section, we study the impact of different data sampling strategies on model performances. There are different benefits in using different data sampling schemes in model training. The first is to increase diversity of models. Applying data sampling \textbf{locally} during tree building are shown to improve diversity of trees efficiently. Similarly, applying data sampling \textbf{globally} can enforce our assumptions on the data structure to the model training process to force models to be less correlated to each other by design. Another reason is to reduce computational time in model training, which is critical as newer versions of the Numerai datasets has include more features and data eras.

%% Sampling within era 
We consider two sampling strategies $S_1,S_2$ that can be applied to each data era independently. $S_1$ is the baseline which uses all the data within an era. $S_2$ is the method we propose which uses only around half of the data in each era. 

\begin{itemize}
    \item $S_1$: Using all the stocks within an era
    \item $S_2$: Using all the stocks with target $y \neq 0.5$, which means select all the stocks that is not equal to the median value of target. On average we obtain around $45\%-55\%$ of stocks in each era. 
    %\item $S_3$: Using $50\%$ of stocks sampled by random within an era without replacement 
    %\item $S_4$: Using $25\%$ of stocks sampled by random within an era without replacement 
\end{itemize}

%% Why S2 
The reason to remove data with target values equal to the Median value is that these data provide little information in learning the ranking of stocks near the tails, which has a bigger impact on the trading portfolio. In practise, only stocks at the top and bottom of the rankings are traded due to transaction costs. Another reason to use $S_2$ is that it can reduce computational time by half, therefore allowing researchers to train more base models within a deep IL model. 

To demonstrate whether the new data sampling scheme $S_2$ can work well for a wide range of parameter settings for GBDT models, a grid search on two key hyperparameters namely feature sampling per tree and the depth of trees is performed for each data sampling scheme.
\begin{itemize}
    \item Tree depth: 4,6
    \item Ratio of feature sampling per tree: 0.1,0.25,0.5,0.75,0.9 
\end{itemize}
The Cartesian product over all combinations of the two hyperparameters gives 10 different hyperparameter settings $G_1, \dots G_{10}$.

The grid search procedure is described in Algorithm\ref{alg:Rain-hyperopt}. 

\begin{algorithm}[H]
\caption{Grid Search on hyperparameter settings for XGBoost models}
\label{alg:Rain-hyperopt}
\KwIn{Number of boosting rounds $B$, Training size of Layer 1 $X_1=585$, Data embargo $b=15$}
Set starting Era $D_1=801$ \\
\For{$1 \leq j \leq 2$}{ 
    Prepare training data from Era $D_1-X_1-b$ to $D_1-b$ using one of the data sampling schemes $S_j$ \\
    Set Ansatz learning rate $L = \frac{50}{B}$ \\
    \For{$1 \leq i \leq 10$}{
        Train XGBoost model $M_{i,j}$ with learning rates $L$, hyperparameter setting $G_i$. \\
        Obtain Predictions of models from Era 801 to Era 1070. \\
    }
}
\end{algorithm} 

\begin{figure}[hbt!]
     \centering
        \subfloat[Mean Corr]{
          \includegraphics[width=17.5cm]{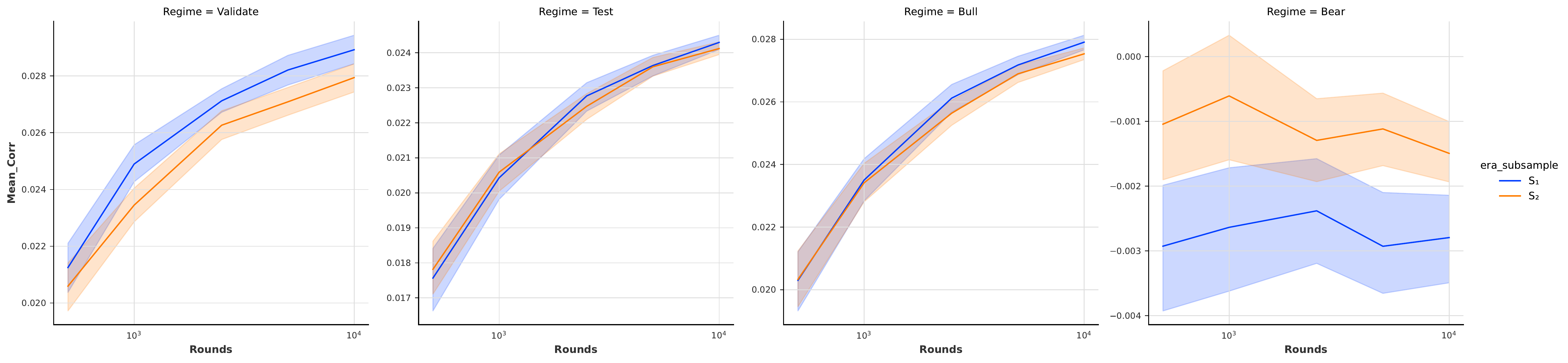}
        }
        \\
        \subfloat[Sharpe]{
          \includegraphics[width=17.5cm]{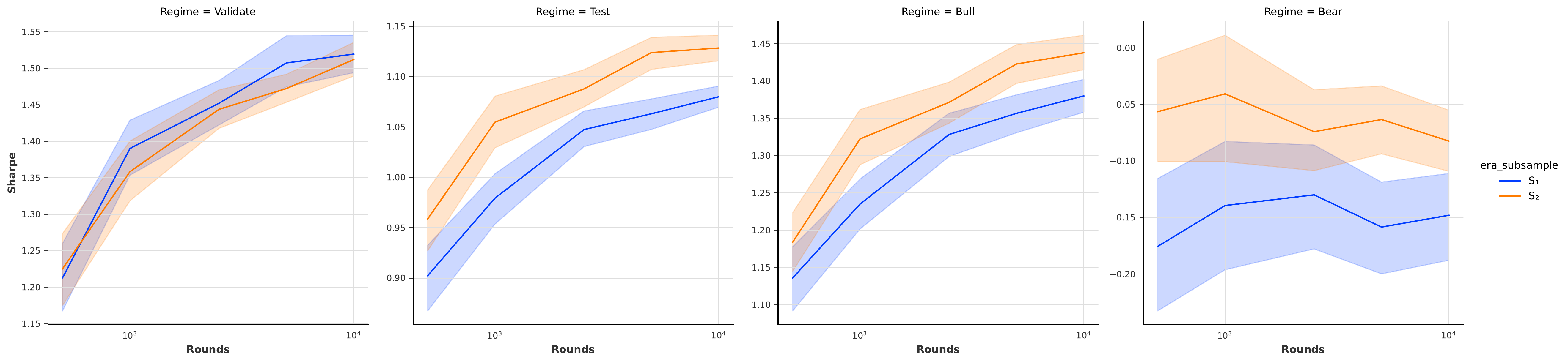}
        }
        \\
        \subfloat[Calmar]{
          \includegraphics[width=17.5cm]{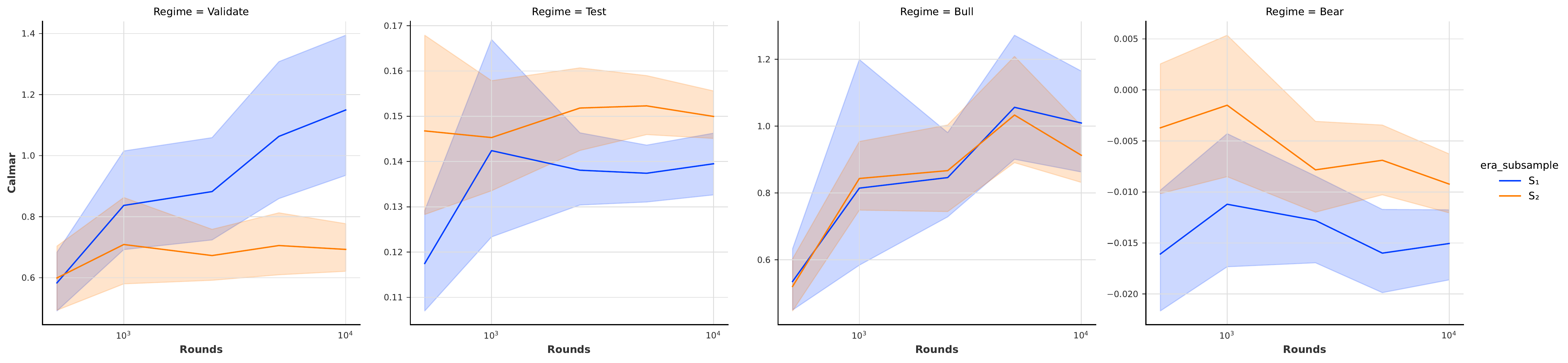}
        }
        \\
    \caption{Comparing the performances of two data sampling schemes $S_1$ and $S_2$ with different number of boosting rounds $B=500,1000,2500,5000$ for risk metrics (a) Mean Corr, (b) Sharpe ratio, (c) Calmar ratio, under different market regimes, over 10 different hyperparameter settings}
    \label{fig:Rain-EraSampling}
\end{figure}

We run the above procedure for different number of boosting rounds $B$, with $B=500,1000,2500,5000$. Each hyperparameter setting is repeated over 4 models using $25\%$ of eras in training data regularly sampled. In Figure \ref{fig:Rain-EraSampling}, we show the risk metrics of the two data sampling schemes, averaged over 10 different hyperparameters settings under different market regimes for different $B$s. Sampling with all the data $S_1$ achieves better Mean Corr in the validation but is not significant in the test period. However, in the test period $S_2$ achieves a better Sharpe and Calmar ratio. 

We then compare the two data sampling schemes under market regimes, which demonstrates the improvement from $S_2$ in the test period can be mostly attributed to improvement during Bear market. Using sampling $S_2$ will not lead to a significant deterioration in model performances in bull market and offers valuable hedging benefits during bear market.

\begin{figure}[hbt!]
     \centering
        \subfloat[Mean Corr]{
          \includegraphics[width=17.5cm]{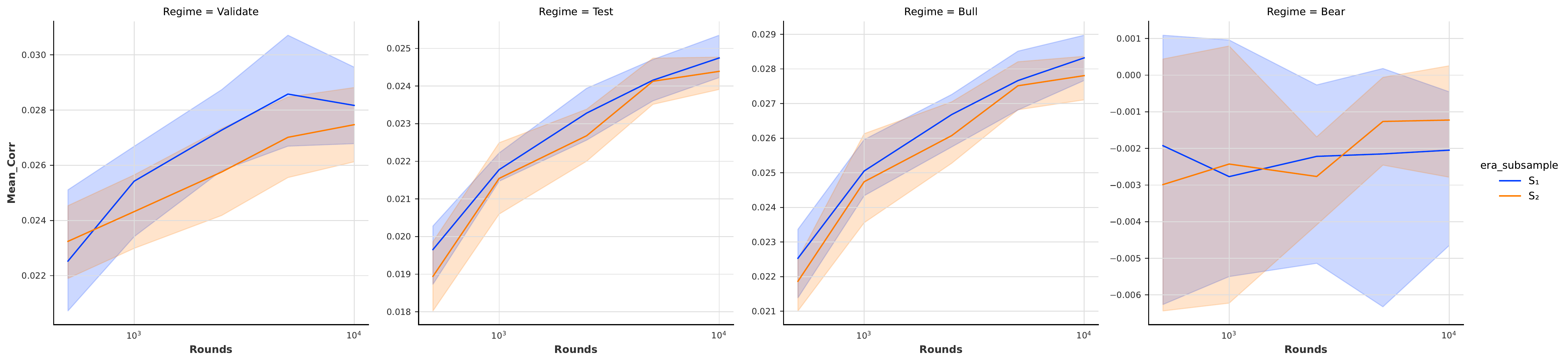}
        }
        \\
        \subfloat[Sharpe]{
          \includegraphics[width=17.5cm]{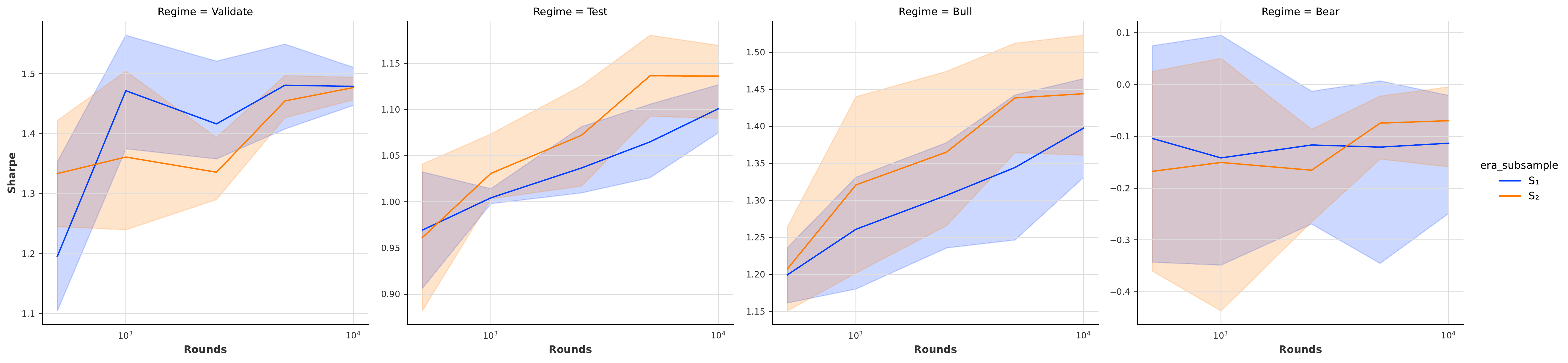}
        }
        \\
        \subfloat[Calmar]{
          \includegraphics[width=17.5cm]{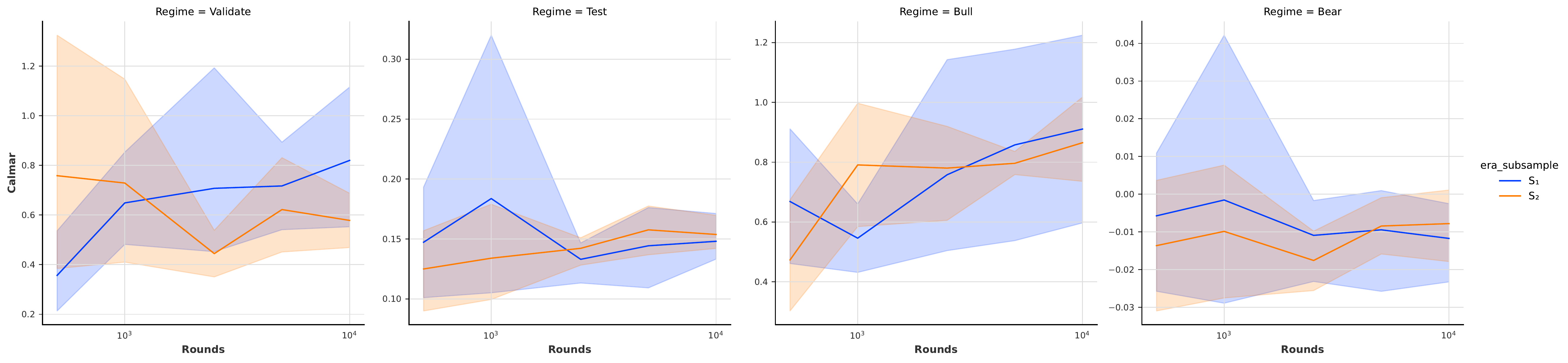}
        }
        \\
    \caption{Comparing the performances of two data sampling schemes with different number of boosting rounds $B=500,1000,2500,5000$ for risk metrics (a) Mean Corr, (b) Sharpe ratio, (c) Calmar ratio, under different market regimes, using the Ansatz hyperparameters Tree Depth = 4 and Ratio of feature sampling per tree = 0.75.}
    \label{fig:Rain-EraSampling-Ansatz}
\end{figure}

Repeating the above analysis using the Ansatz model hyperparameters (Tree Depth = 4 and Ratio of feature sampling per tree = 0.75) only, as shown in Figure \ref{fig:Rain-EraSampling-Ansatz} demonstrated a even smaller performance gap between model performances of $S_1$ and $S_2$. 

Therefore, we use $S_2$ to train train the benchmark models and deep IL XGBoost models.

\subsubsection{Using the Ansatz formula to calculate learning rates of GBDT models}
\label{section:NumeraiRain-LRopt}

We used the Ansatz formula $L = \frac{50}{B}$ to derive the learning rate $L$ for a given number of boosting rounds. Here, we will demonstrate this formula is indeed optimal. 

%% How we select learning rate in current liteature 
Different approaches are used by researchers to select the learning rates and the number of boosting rounds of GBDT and other ML models for tabular datasets. For small tabular datasets, most researchers would use the default values given by the packages without additional tuning. For example, the Gradient Boosting Regression in Scikit-Learn has default values of 100 boosting rounds and 0.1 learning rate. For larger datasets, researchers would perform hyperparameter optimisation to select the optimal learning rate and boosting rounds.

In most benchmark research papers on ML algorithms for tabular datasets \cite{mcelfresh2023neural,Shwartz21,Leo22,}, a random search or other Bayesian approach is use to optimise all the hyperparameters of the ML models, ignoring the joint interactions between hyperparameters. 

In other research papers \cite{Maddock22,Adler22,}, either the number of boosting rounds and/or the learning rate is fixed and then the other hyperparameter is optimised. Recent research \cite{WANG2022288,} suggests the learning rate should be determined dynamically to adapt to concept drift in data. 

Traditional methods of hyper-parameter optimisation based on random or grid searches are problematic as they ignore the key relationship between the two hyperparameters for GBDT models, namely learning rates and the number of boosting rounds. A blind uniform search on the two dimensional hyperparameter space formed by the number of boosting rounds and learning rate is inefficient. 

Intuitively, when learning rate is large, we expect the optimal number of boosting rounds to be small. Similarly, when learning rate is small, the optimal number of boosting rounds should be large. This suggests the optimal learning rate can be written in the form $L = \frac{C}{B} + \mathcal{O}(\frac{1}{B}) $ where B is the number of boosting rounds, and C is a constant that depends on the dataset and other hyperparameters of the GBDT model. For simplicity, we will ignore the higher order terms and assume $L = \frac{C}{B}$.

%% Hypothesis on learning rates of GBDT models 
Using above insights we propose two hypothesis about the learning rates of GBDT models: 

\begin{hypothesis*}[Monotonicity of model performances with respect to learning rate] 
Consider a GBDT model $\mathcal{M}$ with fixed hyper-parameters except the learning rate $l$ and the number of boosting rounds $B$. Let $\mathcal{L}_l(B)$ be the loss function of the GBDT model, parameterised by the learning rate and the number of boosting rounds. 

For any two learning rates $0<l_1<l_2$, we define the minimal value of loss function of model trained with learning rate $l_1$ obtained at boosting round $B_{l_1}$ as $\mathcal{L}^*_{l_1}(B_{l_1})$. Similarly we define the minimal value of loss of model trained with learning rate $l_2$ as $\mathcal{L}^*_{l_2}(B_{l_2})$. We would then have $\mathcal{L}^*_{l_1}(B_{l_1}) \leq \mathcal{L}^*_{l_2}(B_{l_2})$. In other words, as we decrease the learning rate, the theoretical optimal model would become better. 
\end{hypothesis*}

We note that this hypothesis is not contradictory to results obtained by random/grid hyperparameter searches in different experiments as we are working within a finite computational budget. The theoretical optimal model may not be able to be reached if we set the upper bound of the number of boosting rounds to a small value. In this situation, the local optimal model within the hyperparameter grid would not always be the model with the smallest learning rate.

\begin{hypothesis*}[Linear bounds on the number of boosting rounds required to achieve better performances] 
Consider a GBDT model $\mathcal{M}$ with fixed hyper-parameters except the learning rate $l$ and the number of boosting rounds $B$. Let $\mathcal{L}_l(B)$ be the loss function of the GBDT model, parameterised by the learning rate and the number of boosting rounds. 
For any given learning rate $l>0$, number of boosting rounds $B$ and any constant $c>1$, we have $\mathcal{L}_{\frac{l}{c}}(cB) \leq \mathcal{L}_{l}(B)$. 
\end{hypothesis*}

This hypothesis provides us a way to extrapolate optimised learning rates for a given number of boosting rounds to others. This is the basis for the Ansatz learning formula, $L = \frac{C}{B}$.

In Algorithm\ref{alg:Rain-LROpt} we describe how to search over different learning rates for XGBoost models with a given number of boosting rounds$B$. $B$ is set to 1000,2500,5000,50000. Hyperparameters other than the learning rates are the same as the ones used by benchmark Ansatz model. 

For each $B$, we create 5 learning rates based on the Ansatz learning rate $L= \frac{50}{B}$ by considering learning rates larger ($2L$,$4L$) and smaller ($\frac{L}{2}$, $\frac{L}{4}$). In total we have 5 different learning rates including the Ansatz. We train 4 models for each learning rate, where each model is trained using $25\%$ of data eras in the training period, regularly sampled with different start era so the whole dataset is covered. 

\begin{algorithm}[H]
\caption{XGBoost models over learning rate}
\label{alg:Rain-LROpt}
\KwIn{Number of boosting rounds $B$, Training size of Layer 1 $X_1=585$, Retrain Frequency $T=50$, Data embargo $b=15$}
Set starting Era $D_1=801$ \\
Prepare training data from Era $D_1-X_1-b$ to $D_1-b$ \\
Set Ansatz learning rate $L = \frac{50}{B}$ \\
\For{$1 \leq j \leq 5$}{
    Train Layer 1 XGBoost models $M_j$, with learning rates $l_j=\frac{8L}{2^j}$, other hyperparameters are unchanged. \\
    Obtain 10 model snapshot predictions from model $M_j$, taken at boosting rounds $\frac{B}{10}, \frac{2B}{10}, \dots, B$
}
\end{algorithm}

\begin{figure}[hbt!]
     \centering
     \includegraphics[width=17.5cm]{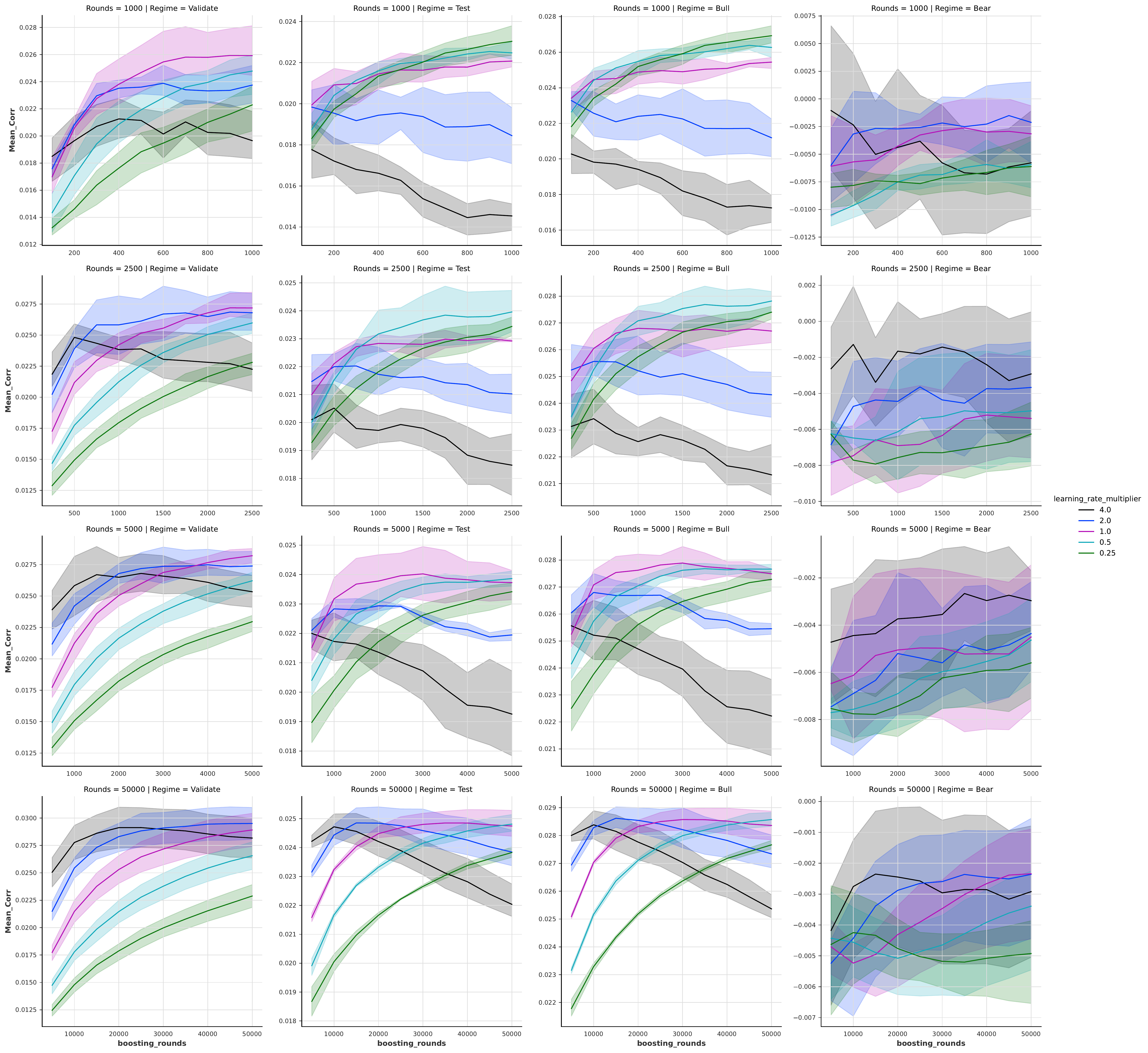}
    \caption{Learning curves of XGBoost models with different learning rates for different number of boosting rounds $B=1000,2500,5000,50000$ for risk metric Mean Corr under different market regimes}
    \label{fig:Rain-LROpt-MeanCorr}
\end{figure}

\begin{figure}[hbt!]
     \centering
     \includegraphics[width=17.5cm]{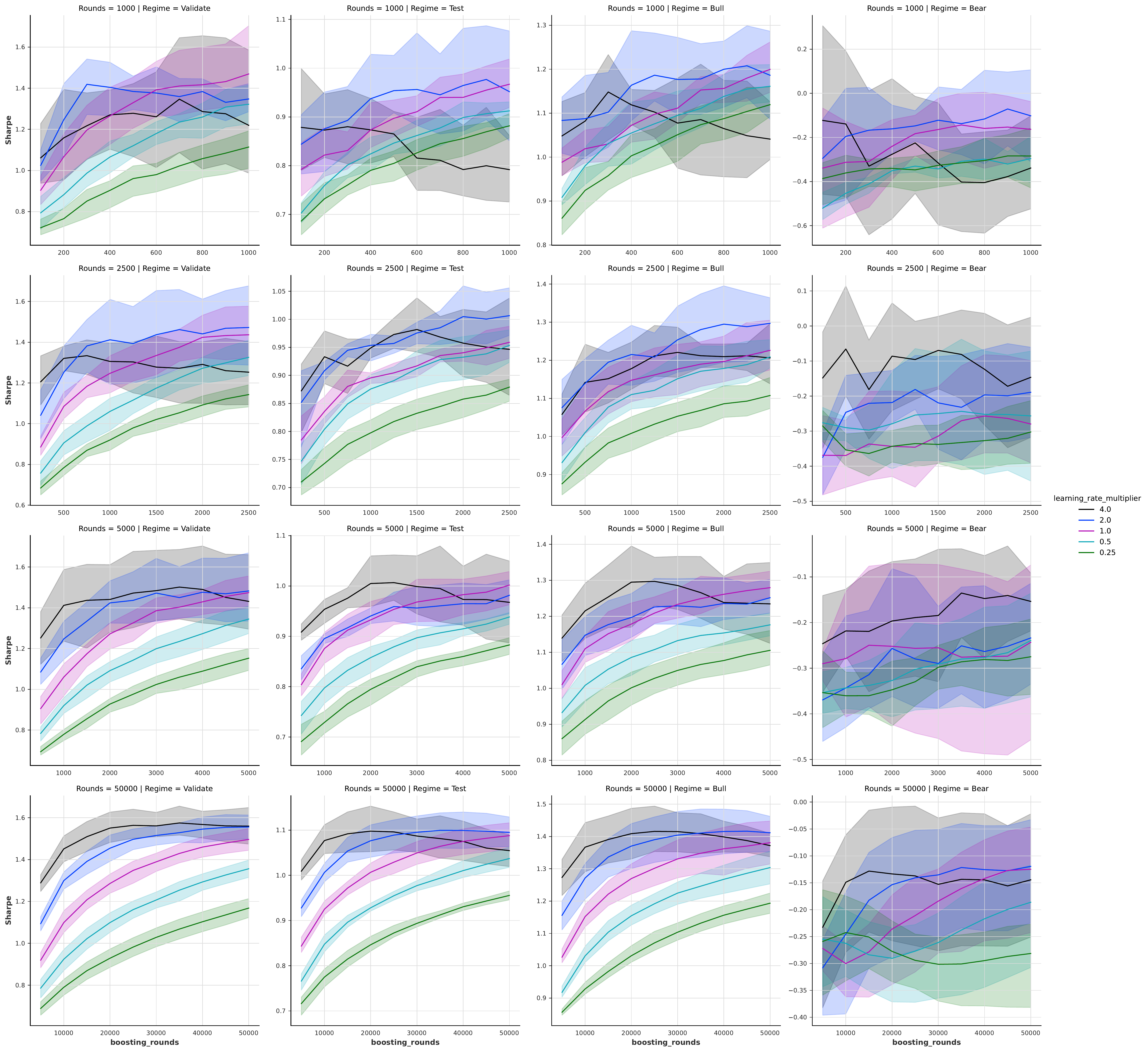}
    \caption{Learning curves of XGBoost models with different learning rates for different number of boosting rounds $B=1000,2500,5000,50000$ for risk metric Sharpe ratio under different market regimes}
    \label{fig:Rain-LROpt-Sharpe}
\end{figure}

\begin{figure}[hbt!]
     \centering
     \includegraphics[width=17.5cm]{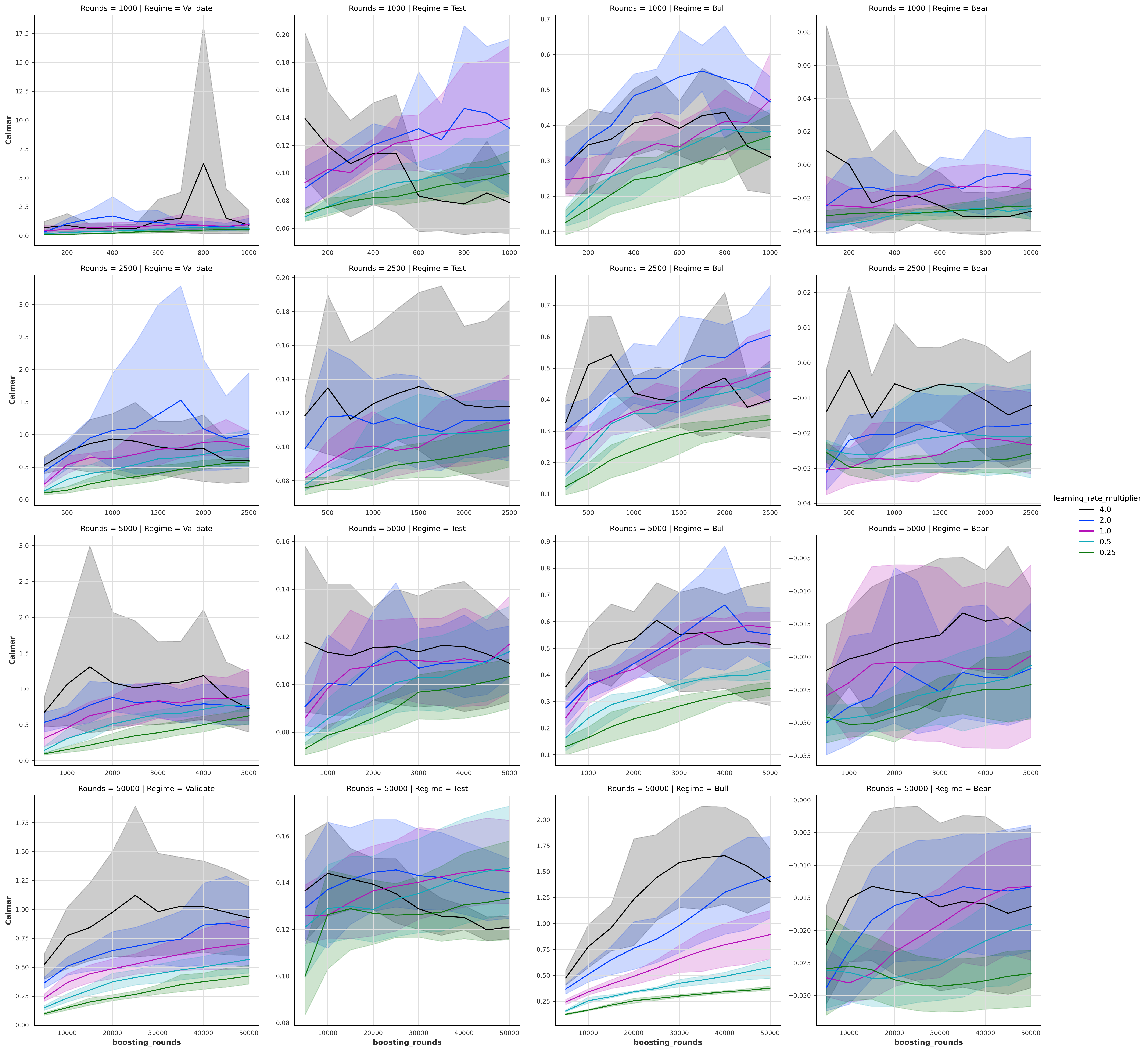}
    \caption{Learning curves of XGBoost models with different learning rates for different number of boosting rounds $B=1000,2500,5000,50000$ for risk metric Calmar ratio under different market regimes}
    \label{fig:Rain-LROpt-Calmar}
\end{figure}

%% Results in Validation Period for Learning curves with differnet learning rate for a given number of boosting rounds 
In Figures \ref{fig:Rain-LROpt-MeanCorr}, \ref{fig:Rain-LROpt-Sharpe} and \ref{fig:Rain-LROpt-Calmar}, the learning curves of XGBoost models with different learning rates are shown for the risk metrics: Mean Corr, Sharpe ratio and Calmar ratio under different regimes. Learning curves show the value of a metric over different stages of the model training process, indicated by the number of boosting rounds.

In the validation period, models of learning rates of $4L$ demonstrated overfitted behaviour. Models of learning rates of $\frac{L}{2}$ and $\frac{L}{4}$ had a lower performance than the models with Ansatz learning rate $L$ in the validation period, suggesting the model is under-fitted. Models with learning rate $2L$ have similar performances with models with learning rate $L$ but with a larger model variance. Therefore, models with learning rate $L$ is indeed optimal in the validation period. 

Models with lower learning rates have a more stable training process. In particular, we observe a monotonic increasing trend of learning curves for learning rate $\frac{L}{4}$ within the computational budget $B$ boosting rounds. This property suggests when we are training large models using a very small learning rate, early stopping is not necessary since we will observe only strictly improving model performances in validation period after removing noise effects. However, using a very small learning rate will require a very long training time, which is infeasible in real applications. 

In test period, learning curves of rates $\frac{L}{4}$ and $\frac{L}{2}$ performed slightly better, but only significant for small Bs where $B \leq 2500$. When $B \geq 5000$, the Anstaz learning rate $L$ gives the best performances. 

Our Ansatz balances both the need of model training efficiency and stability of training process. It gives the upper bound on the optimal learning rate before there is risk of overfitting in the data. Therefore, we cannot further increase learning rate to make model training more efficient without taking additional risks of model overfitting. 

The learning curves of other metrics, such as Sharpe and Calmar ratios are noisy and therefore we do not select learning rates based on those.

\newpage
\subsubsection{Selecting ML models for Temporal Tabular Datasets} 
\label{section:NumeraiSunshine-HyperOpt}

For different tabular models introduced in section \ref{section:NumeraiSunshine-tabular-inc}, hyperparameter optimisation is performed using data before 2018-04-27 (Era 800). The training and validation set is data between 2003-01-03 (Era 1) and 2014-06-26 (Era 600), with the last $25\%$ of data (Era 451 - Era 600) as the validation set. 
Due to memory constraints, era sub-sampling is applied during model training. $25\%$ of the eras in the training period is used with sampling performed at regular intervals. The performance of the models in the evaluation period, from 2014-07-04 (Era 601) to 2018-04-27 (Era 800) is then used to select hyperparameters for the tabular models. The Mean Corr and Sharpe Ratio of the prediction ranking correlation in the evaluation period is reported. 
Due to memory issues for training neural network models, a global feature selection process is used to select $50\%$ of the 1586 features from the V4.1 dataset at the start of each model process by random.

\paragraph{Multi-Layer Perceptron}

Multi-Layer Perceptron (MLP) models without feature projection are trained with different number of encoding and funnel layers using the architecture described in Section \ref{section:NumeraiSunshine-tabular-inc}. 

\begin{itemize}
    \item Number of Feature Eng Layers: 0,1,2,3,4
    \item Number of Funnel Layers: 1,2,3 
\end{itemize}

Other hyperparameters of the neural network models are fixed in the grid search as follows: (Number of epochs: 100, Early Stopping: 10,  Learning Rate: 0.001, Dropout: 0.1, Encoding Neuron Scale: 0.8, Funnel Neuron Scale: 0.8, Gradient Clip: 0.5, Loss Function: Pearson Corr)

In Table \ref{table:MLP1} shows the performances of MLP models with different network architectures over 5 different random seeds.

The architecture with the highest Mean Corr is the model without feature engineering layers and a standard MLP model with 2 linear layers. When the number of funnel layers equals to 1, the MLP model is equivalent to a (regularised) linear model and has the worst performance. Increasing the number of feature engineering layers does not significantly improve Mean Corr. As model complexity increases, model performances are more varied over different random seeds, suggesting the lack of robustness of deep neural network models.

\begin{table}[!ht]
    \centering
    \begin{tabular}{|l|l|l|l|l|l|l|l|}
    \hline
        Feature Eng Layers  & Funnel Layers & Mean Corr & Sharpe  & Calmar  \\ \hline
        \multirow{3}*{0}   & 1  & 0.0159 $\pm$ 0.0016   & 0.8042 $\pm$ 0.0631  & 0.0826 $\pm$ 0.0087 \\ 
         \cline{2-5}
          & 2  & \textbf{0.0235} $\pm$ 0.0001   & \textbf{1.1344} $\pm$ 0.0119  & \textbf{0.2692} $\pm$ 0.0201 \\ 
          \cline{2-5}
          & 3  & 0.0223 $\pm$ 0.0005   & 1.0478 $\pm$ 0.0372  & 0.2117 $\pm$ 0.0072 \\ \hline
        \multirow{3}*{1}  & 1  & 0.0222 $\pm$ 0.0003   & 1.0509 $\pm$ 0.0112  & 0.2118 $\pm$ 0.0068 \\ 
        \cline{2-5}
          & 2  & 0.0216 $\pm$ 0.0003   & 1.0061 $\pm$ 0.0377  & 0.2021 $\pm$ 0.0128 \\ \cline{2-5}
          & 3  & 0.0224 $\pm$ 0.0003   & 1.0575 $\pm$ 0.0121  & 0.2212 $\pm$ 0.0269 \\ \hline
        \multirow{3}*{2}  & 1  & 0.0217 $\pm$ 0.0004   & 1.0176 $\pm$ 0.0357  & 0.2104 $\pm$ 0.0178 \\ 
        \cline{2-5}
          & 2  & 0.0218 $\pm$ 0.0009   & 1.0346 $\pm$ 0.0571  & 0.2005 $\pm$ 0.0076 \\ 
          \cline{2-5}
          & 3  & 0.0226 $\pm$ 0.0006   & 1.0754 $\pm$ 0.0352  & 0.2348 $\pm$ 0.0242 \\ \hline
        \multirow{3}*{3}  & 1  & 0.0224 $\pm$ 0.0006   & 1.0467 $\pm$ 0.0402  & 0.2226 $\pm$ 0.0281 \\ 
        \cline{2-5}
          & 2  & 0.0221 $\pm$ 0.0009   & 1.0564 $\pm$ 0.0441  & 0.2332 $\pm$ 0.0291 \\ 
          \cline{2-5}
          & 3  & 0.0217 $\pm$ 0.0007   & 1.0245 $\pm$ 0.0414  & 0.2049 $\pm$ 0.0156 \\ 
          \hline
        \multirow{3}*{4}  & 1  & 0.0215 $\pm$ 0.0006   & 1.0131 $\pm$ 0.0192  & 0.1980 $\pm$ 0.0146 \\ 
        \cline{2-5}
          & 2  & 0.0219 $\pm$ 0.0010   & 1.0490 $\pm$ 0.0673  & 0.2229 $\pm$ 0.0309 \\ 
          \cline{2-5}
          & 3  & 0.0218 $\pm$ 0.0017   & 1.0459 $\pm$ 0.0880  & 0.2513 $\pm$ 0.0229 \\ \hline
    \end{tabular}
    \smallskip
        \caption{Neural Network models between  2014-07-04 (Era 601) and 2018-04-27 (Era 800)}
            \label{table:MLP1}
\end{table}

\paragraph{XGBoost}

Root Mean Square Error (RMSE), the standard loss function for regression problems is used to train the XGBoost models. Early-stopping based on Pearson correlation in the validation set is applied to control the model complexity if needed. A grid search is performed to select the data sub-sample and feature sub-sample ratios of the XGBoost models. 

\begin{itemize}
    \item Max Depth: 4,6,8 
    \item Data subsample by tree: 0.25,0.5,0.75
    \item Feature subsample by tree: 0.25,0.5,0.75
    \item L1 regularisation: 0, 0.001, 0.01
    \item L2 regularisation: 0, 0.001, 0.01
\end{itemize}

Other hyperparameters of the XGBoost models are fixed as follows: (Number of boosting rounds: 5000, Learning rate: 0.01, Grow policy: Depth-wise, Min Samples per node: 10, Feature subsample by level/node: 1)

%% Comparison of XGBoost performances over different parameters 
Table \ref{table:XGB1} compares performances of XGBoost models by different data subsample ratios, feature subsample ratios and max depth, mean and standard deviation over 45 models of the 9 combinations of L1 and L2 regularisation each with 5 different random seeds are reported.

Calmar ratio is the performance metric with the most variance, suggesting selecting models based on Calmar ratio is not robust. Mean Corr is the least varied metric between random seeds and therefore we use it for hyperparameter selection. 

Models with data sub-sampling ratio of $75\%$ performed better than models with data sub-sampling ratio of $50\%$ and $25\%$, with a lower variance between model performances over different random seeds also. Models with feature sub-sampling ratio of $75\%$ also performed better. XGBoost models with max depth of 4 performed better than models with max depth of 6 and 8 for each fixed data and feature sub-sampling ratios. 

Table \ref{table:XGB2} compares performances of XGBoost models by different L1 and L2 regularisation and max depth with fixed data and feature sub-sampling ratio of $75\%$. Mean and standard deviation over 5 models with different random seeds are reported. There are no significant difference between model performances over different L1 and L2 regularisation when other model hyperparameters are fixed. Therefore, we set the L1 and L2 regularisation penalty to be zero when training XGBoost models.

We conclude the optimised hyperparameters as follows: (Number of boosting rounds: 5000, Learning rate: 0.01, Max Depth: 4,Data subsample by tree: 0.75,  Feature subsample by tree: 0.75, L1 regularisation: 0, L2 regularisation: 0, Grow policy: Depth-wise, Min Samples per node: 10, Feature subsample by level/node: 1)

\paragraph{Conclusion} 

%% Complex MLP models are not helpful 
Increasing complexity of MLP models cannot improve model performances. The best performance is achieved by a standard MLP with two layers, which provides the minimal amount of non-linearity required so that the model does not degenerate to a ridge regression model. As suggested in \cite{Teresa22}, the performance of over-parameterised models are affected by a myriad of factors including model architecture and training process. It cannot be ruled out that there are other model architectures that can make deep learning models performing better than XGBoost. However, as suggested from research on bench-marking of tabular ML models \cite{Leo22}, recent deep learning models for tabular data such as TabNet does not always perform better than MLP models. It is unlikely there are advance neural network architectures that are efficient and performed better than MLP.  

%% XGBoost models performed better than MLP 
XGBoost models performed better than MLP models over a wide range of hyperparameters in the evaluation period. The binned nature of features favours the use of decision trees over neural networks, and this view is shared by different reviews on ML algorithms for tabular data \cite{Shwartz21,Leo22,}. Moreover, MLP models take longer computational time to train and suffers from memory constraints. Therefore, we do not consider the use of MLP in building deep model ensemble for the Numerai dataset.  For similar reasons,  other advanced neural architectures are not explored here given their high computational resources requirement. These architectures are also known to have a high variation of performances over random seeds \cite{gundersen2023sources} and their hyperparameters are difficult to tune.

\begin{table}[!htb]
    \centering
    \begin{tabular}{|l|l|l|l|l|l|l|l|l|}
    \hline
    Data Sample & Feature Sample & Depth & Mean Corr  & Sharpe  & Calmar \\  \hline
    \multirow{9}*{0.25}        &  \multirow{3}*{0.25}          & 4     & 0.0242    $\pm$ 0.0014    & 1.2126 $\pm$ 0.0992 & 0.3451 $\pm$ 0.1237 \\ 
    \cline{3-6} &    & 6     & 0.0225    $\pm$ 0.0018    & 1.1502 $\pm$ 0.1034 & 0.3275 $\pm$ 0.0727 \\ 
     \cline{3-6} &           & 8     & 0.0187    $\pm$ 0.0015    & 1.0045 $\pm$ 0.0904 & 0.2227 $\pm$ 0.0858 \\ 
    \cline{2-6} &   \multirow{3}*{0.5}       & 4     & 0.0236    $\pm$ 0.0014    & 1.1929 $\pm$ 0.0706 & 0.2804 $\pm$ 0.0413 \\ 
    \cline{3-6} &            & 6     & 0.0222    $\pm$ 0.0014    & 1.1193 $\pm$ 0.0825 & 0.2495 $\pm$ 0.075  \\ 
    \cline{3-6} &               & 8     & 0.0189    $\pm$ 0.0012    & 0.9999 $\pm$ 0.066  & 0.23   $\pm$ 0.0791 \\ 
    \cline{2-6} &   \multirow{3}*{0.75}          & 4     & 0.0249    $\pm$ 0.0016    & 1.258  $\pm$ 0.1066 & 0.3501 $\pm$ 0.1275 \\ 
    \cline{3-6} &             & 6     & 0.0228    $\pm$ 0.0013    & 1.1414 $\pm$ 0.0666 & 0.2974 $\pm$ 0.0864 \\ 
    \cline{3-6} &          & 8     & 0.0188    $\pm$ 0.0023    & 0.9734 $\pm$ 0.1425 & 0.1848 $\pm$ 0.075  \\ 
    \hline
    \multirow{9}*{0.5}        &  \multirow{3}*{0.25}
             & 4     & 0.0259    $\pm$ 0.0009    & 
    1.2751 $\pm$ 0.055  & 0.3641 $\pm$ 0.0809 \\ 
    \cline{3-6}
    &          & 6     & 0.0248    $\pm$ 0.0012    & 1.2453 $\pm$ 0.0768 & 0.3862 $\pm$ 0.1135 \\ 
     \cline{3-6}
    &           & 8     & 0.0217    $\pm$ 0.0018    & 1.1244 $\pm$ 0.1076 & 0.3684 $\pm$ 0.143  \\ 
     \cline{2-6}
    &  \multirow{3}*{0.5}       & 4     & 0.0267    $\pm$ 0.001     & 1.3394 $\pm$ 0.0908 & 0.4423 $\pm$ 0.1279 \\ 
     \cline{3-6}
     &               & 6     & 0.0255    $\pm$ 0.001     & 1.2733 $\pm$ 0.0603 & 0.4521 $\pm$ 0.1521 \\
     \cline{3-6}
    &             & 8     & 0.0224    $\pm$ 0.0011    & 1.1622 $\pm$ 0.063  & 0.4375 $\pm$ 0.1299 \\ 
    \cline{2-6}
    & \multirow{3}*{0.5}          & 4     & 0.0268    $\pm$ 0.0011    & 1.3173 $\pm$ 0.0842 & 0.413  $\pm$ 0.0998 \\ 
    \cline{3-6}
    &           & 6     & 0.0255    $\pm$ 0.0011    & 1.2716 $\pm$ 0.075  & 0.4429 $\pm$ 0.1468 \\ 
    \cline{3-6}
    &           & 8     & 0.0226    $\pm$ 0.0014    & 1.1566 $\pm$ 0.1021 & 0.4315 $\pm$ 0.146  \\ 
    \hline
     \multirow{9}*{0.75}        &  \multirow{3}*{0.25}           & 4     & 0.0265    $\pm$ 0.0009    & 1.3146 $\pm$ 0.0605 & 0.4388 $\pm$ 0.0731 \\ 
     \cline{3-6}
     &           & 6     & 0.0268    $\pm$ 0.0009    & 1.3439 $\pm$ 0.0778 & 0.6006 $\pm$ 0.2169 \\ 
    \cline{3-6}
    &           & 8     & 0.0235    $\pm$ 0.0005    & 1.2071 $\pm$ 0.048  & 0.5044 $\pm$ 0.1404 \\ 
    \cline{2-6}
     & \multirow{3}*{0.5}          & 4     & 0.0270     $\pm$ 0.0007    & 1.3345 $\pm$ 0.0477 & 0.4345 $\pm$ 0.0665 \\ 
     \cline{3-6}
    &            & 6     & 0.0271    $\pm$ 0.0007    & 1.3469 $\pm$ 0.0479 & \textbf{0.6300}   $\pm$ 0.1526 \\ 
    \cline{3-6}
    &            & 8     & 0.0241    $\pm$ 0.0012    & 1.234  $\pm$ 0.0702 & 0.4843 $\pm$ 0.2099 \\ 
    \cline{2-6}
    & \multirow{3}*{0.75}           & 4     & \textbf{0.0273}    $\pm$ 0.0006    & \textbf{1.3624} $\pm$ 0.0485 & 0.4885 $\pm$ 0.1032 \\ 
    \cline{3-6}
    &            & 6     & 0.0267    $\pm$ 0.0009    & 1.3373 $\pm$ 0.0566 & 0.5509 $\pm$ 0.1314 \\ 
    \cline{3-6}
    &        & 8     & 0.0237    $\pm$ 0.0005    & 1.2369 $\pm$ 0.065  & 0.5501 $\pm$ 0.1776 \\ 
    \hline
    \end{tabular}
    \bigskip
       \caption{XGBoost models with different data subsample ratios, feature subsample ratios and max depths between  2014-07-04 (Era 601) and 2018-04-27 (Era 800)}
           \label{table:XGB1}
\end{table}

\begin{table}[!htb]
    \centering
    \begin{tabular}{|l|l|l|l|l|l|l|l|l|}
    \hline
    Max Depth & L2-reg & L1-reg & Mean Corr & Sharpe & Calmar \\ 
    \hline
       \multirow{9}*{4}        &  \multirow{3}*{0.0} 
            & 0.0    & \textbf{0.0276}    $\pm$ 0.0007    & \textbf{1.3829} $\pm$ 0.0515 & 0.5305 $\pm$ 0.1077 \\ 
    \cline{3-6}
             &    & 0.001  & 0.0274    $\pm$ 0.0006    & 1.3764 $\pm$ 0.0485 & 0.5161 $\pm$ 0.1213 \\ 
    \cline{3-6}
             &    & 0.1    & 0.0268    $\pm$ 0.0008    & 1.3283 $\pm$ 0.0627 & 0.4224 $\pm$ 0.0937 \\ 
    \cline{2-6}
             & \multirow{3}*{0.001}  & 0.0    & \textbf{0.0276}    $\pm$ 0.0007    & \textbf{1.3829} $\pm$ 0.0515 & 0.5305 $\pm$ 0.1077 \\ 
    \cline{3-6}
             &  & 0.001  & 0.0274    $\pm$ 0.0006    & 1.3764 $\pm$ 0.0485 & 0.5161 $\pm$ 0.1213 \\ 
    \cline{3-6}
             &  & 0.1    & 0.0268    $\pm$ 0.0008    & 1.3283 $\pm$ 0.0627 & 0.4224 $\pm$ 0.0938 \\ 
    \cline{2-6}
             & \multirow{3}*{0.1}   & 0.0    & 0.0271    $\pm$ 0.0002    & 1.3489 $\pm$ 0.0253 & 0.4771 $\pm$ 0.0824 \\ 
    \cline{3-6}
             &    & 0.001  & 0.0274    $\pm$ 0.0004    & 1.368  $\pm$ 0.0358 & 0.4977 $\pm$ 0.1186 \\ 
    \cline{3-6}
             &   & 0.1    & 0.0274    $\pm$ 0.0004    & 1.3694 $\pm$ 0.0362 & 0.4833 $\pm$ 0.0918 \\ 
    \hline
    \multirow{9}*{6}        &  \multirow{3}*{0.0} 
             & 0.0    & 0.0266    $\pm$ 0.0008    & 1.3354 $\pm$ 0.0453 & 0.5574 $\pm$ 0.1282 \\ 
    \cline{3-6}
             &    & 0.001  & 0.0267    $\pm$ 0.0011    & 1.3353 $\pm$ 0.061  & 0.5422 $\pm$ 0.1697 \\ 
    \cline{3-6}
             &    & 0.1    & 0.0267    $\pm$ 0.0011    & 1.3201 $\pm$ 0.0638 & 0.5665 $\pm$ 0.1578 \\ 
    \cline{2-6}
             & \multirow{3}*{0.001}  & 0.0    & 0.0265    $\pm$ 0.0007    & 1.3347 $\pm$ 0.0441 & 0.5711 $\pm$ 0.1244 \\ 
    \cline{3-6}
             &  & 0.001  & 0.0268    $\pm$ 0.0011    & 1.3403 $\pm$ 0.0657 & 0.5184 $\pm$ 0.1226 \\ 
    \cline{3-6}
             &  & 0.1    & 0.0267    $\pm$ 0.0011    & 1.3201 $\pm$ 0.0638 & 0.5665 $\pm$ 0.1578 \\ 
    \cline{2-6}
            & \multirow{3}*{0.1}    & 0.0    & 0.0269  $\pm$ 0.0009    & 1.3551 $\pm$ 0.0559 & \textbf{0.6187} $\pm$ 0.1823 \\ 
    \cline{3-6}
             &    & 0.001  & 0.0267    $\pm$ 0.0013    & 1.3307 $\pm$ 0.0818 & 0.4871 $\pm$ 0.1164 \\ 
    \cline{3-6}
             &    & 0.1    & 0.0269    $\pm$ 0.0009    & 1.3638 $\pm$ 0.0563 & 0.5304 $\pm$ 0.0598 \\ 
    \hline
    \multirow{9}*{8}        &  \multirow{3}*{0.0} 
            & 0.0    & 0.0237    $\pm$ 0.0008    & 1.2226 $\pm$ 0.0592 & 0.5127 $\pm$ 0.0875 \\ 
    \cline{3-6}
         &    & 0.001  & 0.0238    $\pm$ 0.0003    & 1.2445 $\pm$ 0.0623 & 0.5269 $\pm$ 0.1949 \\ 
    \cline{3-6}
         &    & 0.1    & 0.0237    $\pm$ 0.0004    & 1.2408 $\pm$ 0.0633 & 0.5298 $\pm$ 0.0905 \\ 
    \cline{2-6}
        & \multirow{3}*{0.001}  & 0.0    & 0.0238    $\pm$ 0.0007    & 1.243  $\pm$ 0.0942 & 0.5657 $\pm$ 0.2615 \\ 
    \cline{3-6}
        &  & 0.001  & 0.0238    $\pm$ 0.0004    & 1.2383 $\pm$ 0.0254 & 0.4905 $\pm$ 0.1505 \\ 
    \cline{3-6}
        &  & 0.1    & 0.0238    $\pm$ 0.0003    & 1.2429 $\pm$ 0.0748 & 0.6487 $\pm$ 0.1551 \\ 
    \cline{2-6}
        & \multirow{3}*{0.1}    & 0.0    & 0.0235    $\pm$ 0.0006    & 1.2102 $\pm$ 0.0655 & 0.6036 $\pm$ 0.2693 \\ 
    \cline{3-6}
        &    & 0.001  & 0.024     $\pm$ 0.0003    & 1.2461 $\pm$ 0.0552 & 0.53   $\pm$ 0.1934 \\ 
    \cline{3-6}
        &    & 0.1    & 0.0235    $\pm$ 0.0008    & 1.2438 $\pm$ 0.1056 & 0.5426 $\pm$ 0.2102 \\ 
    \hline
    \end{tabular}
    \smallskip
        \caption{XGBoost models with different L1 and L2 regularisation with fixed data and feature sub-sampling ratios of $75\%$ between  2014-07-04 (Era 601) and 2018-04-27 (Era 800)}
            \label{table:XGB2}
\end{table}

\newpage
\subsection{Additional Results}

%% Layer 1 Models Feature JK 
\begin{figure}[hbt!]
     \centering
        \subfloat[Mean Corr]{
          \includegraphics[width=8cm]{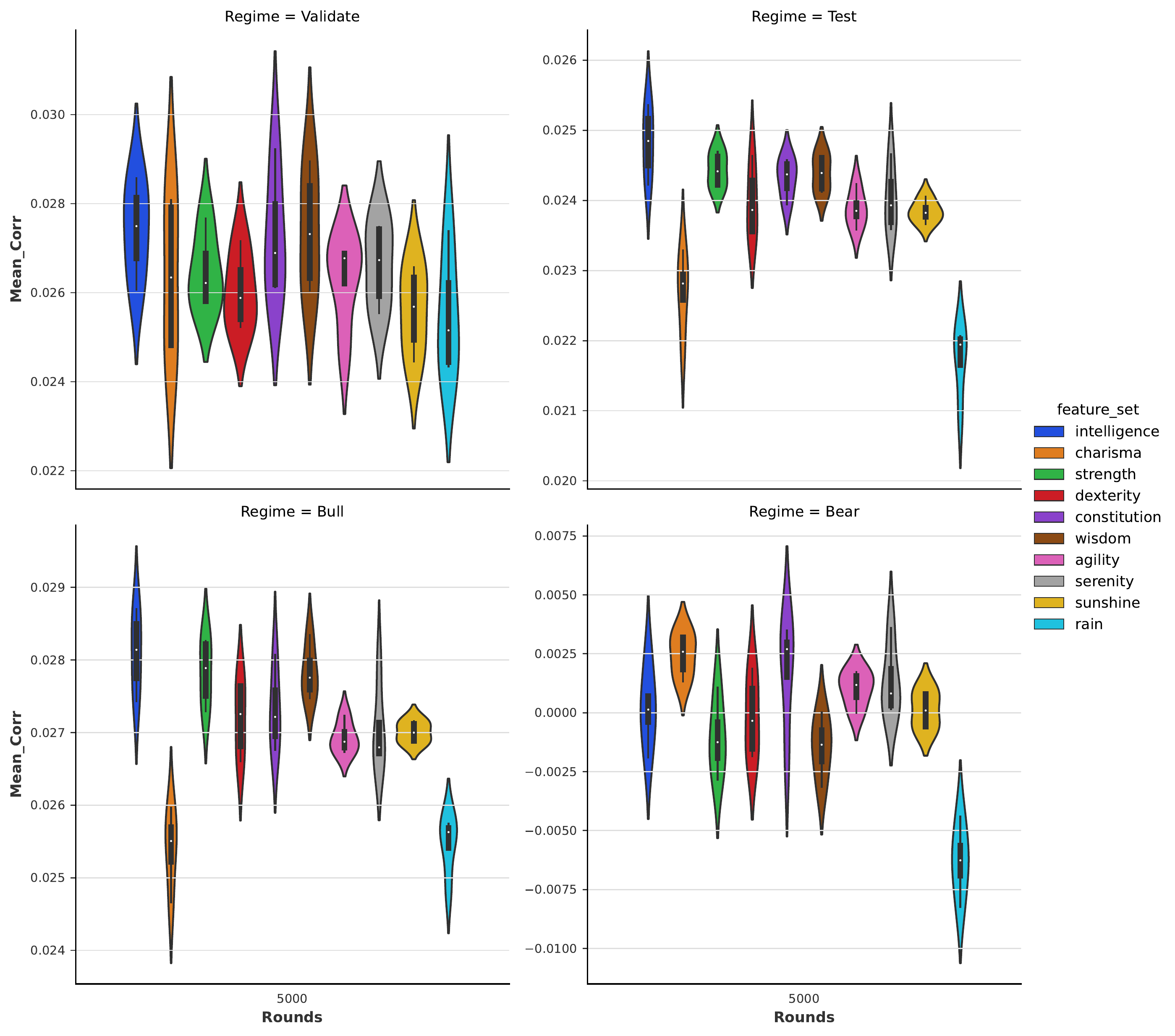}
        }
        \subfloat[Sharpe]{
          \includegraphics[width=8cm]{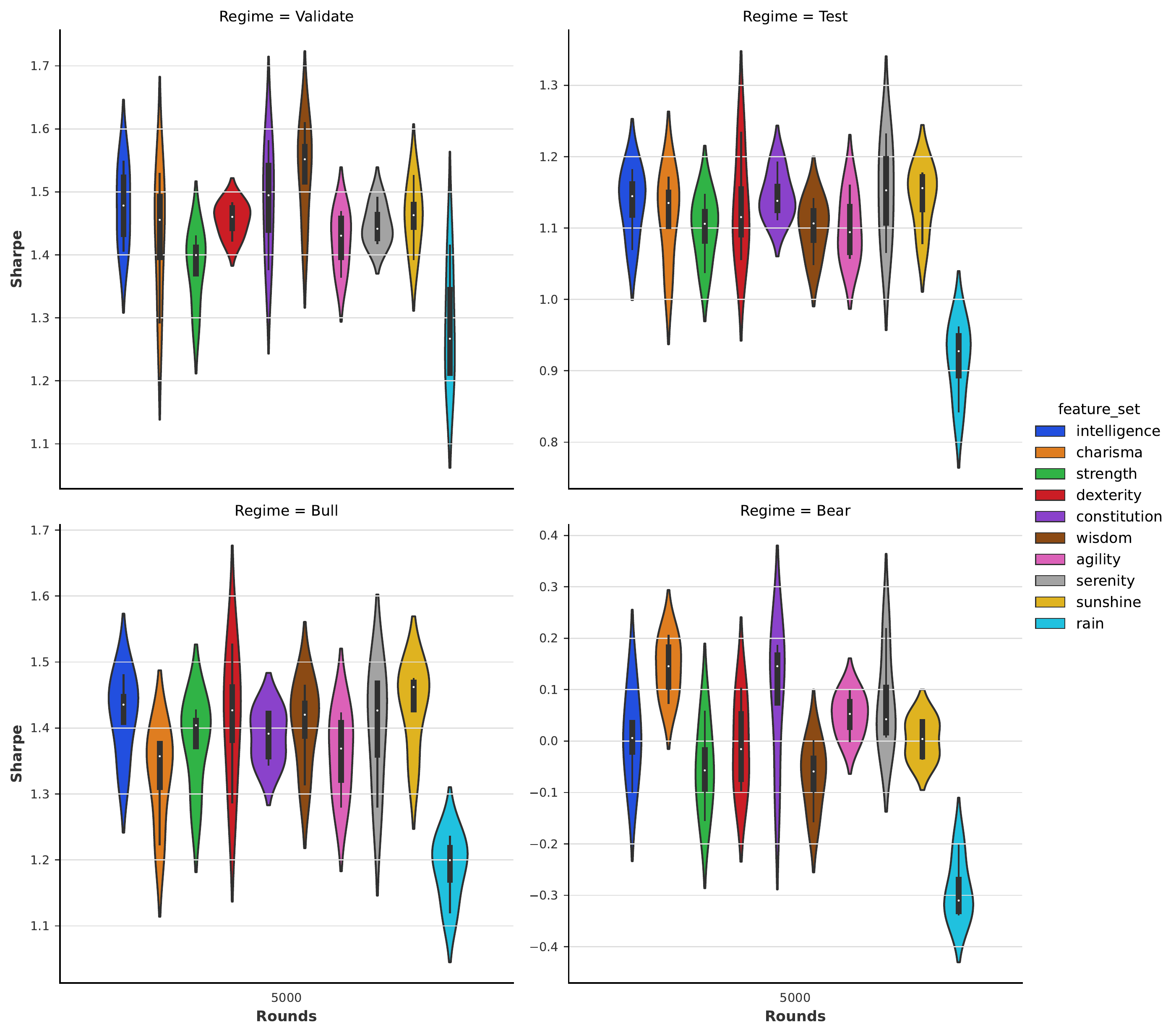}
        }
        \\
        \subfloat[Calmar]{
          \includegraphics[width=8cm]{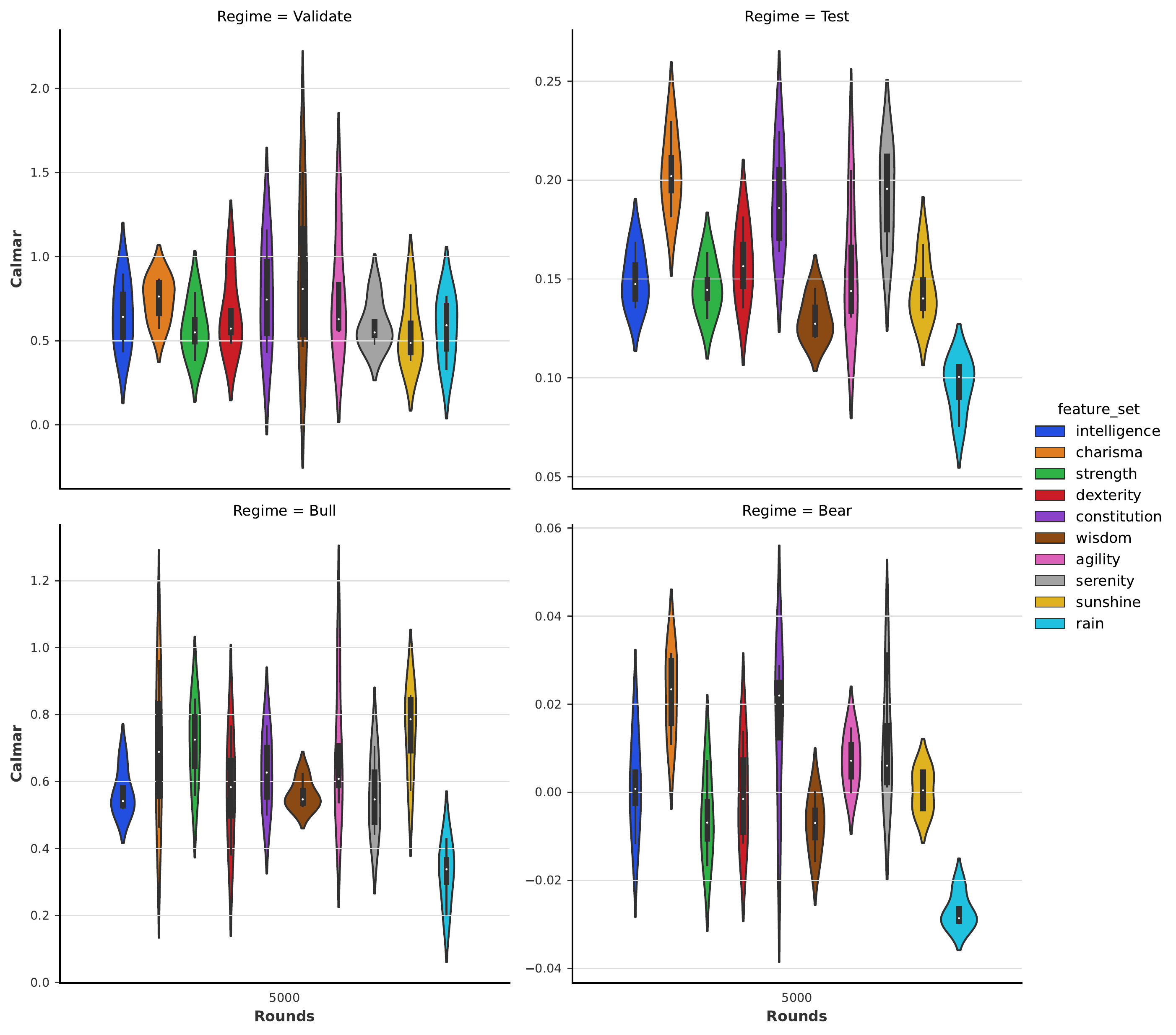}
        }
        \\
    \caption{Performances, (a) Mean Corr, (b) Sharpe ratio, and (c) Calmar ratio of the deep IL XGBoost models with Jackknife feature sampling under different market regimes. }
    \label{fig:Rain-IL-FeatureJK-Layer1}
\end{figure}

%% Layer 1 Models Feature Random 
\begin{figure}[hbt!]
     \centering
        \subfloat[Mean Corr]{
          \includegraphics[width=8cm]{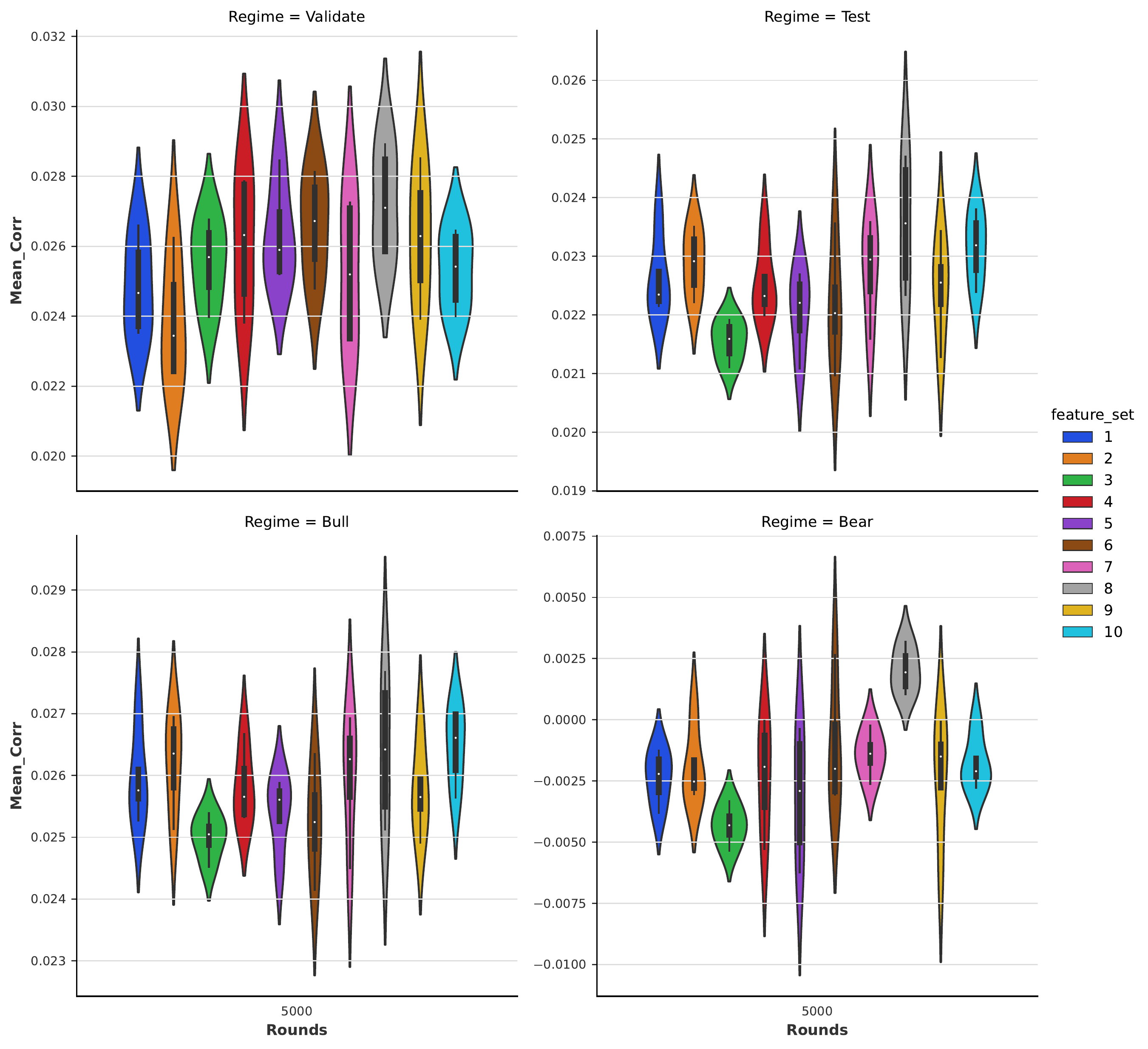}
        }
        \subfloat[Sharpe]{
          \includegraphics[width=8cm]{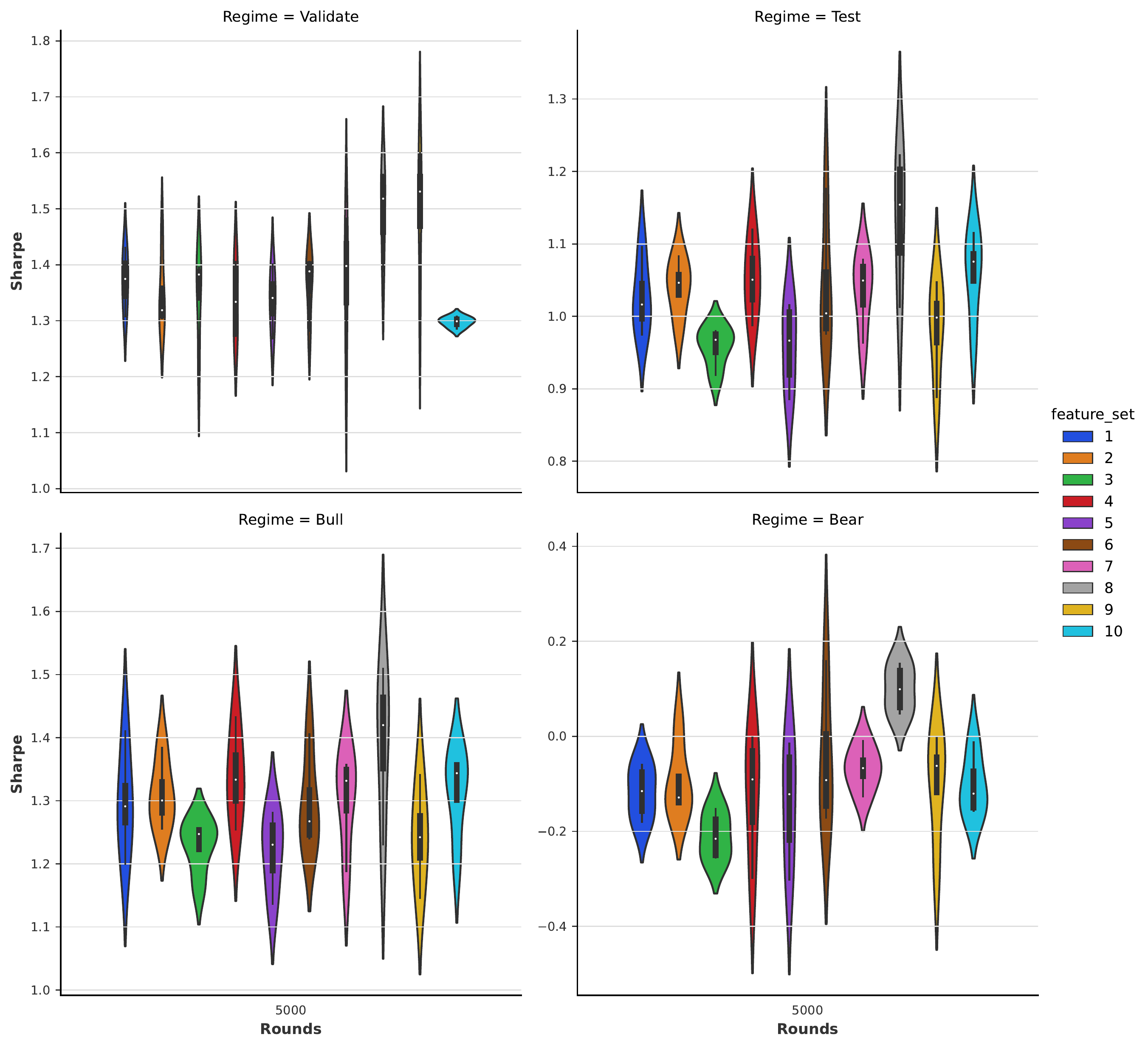}
        }
        \\
        \subfloat[Calmar]{
          \includegraphics[width=8cm]{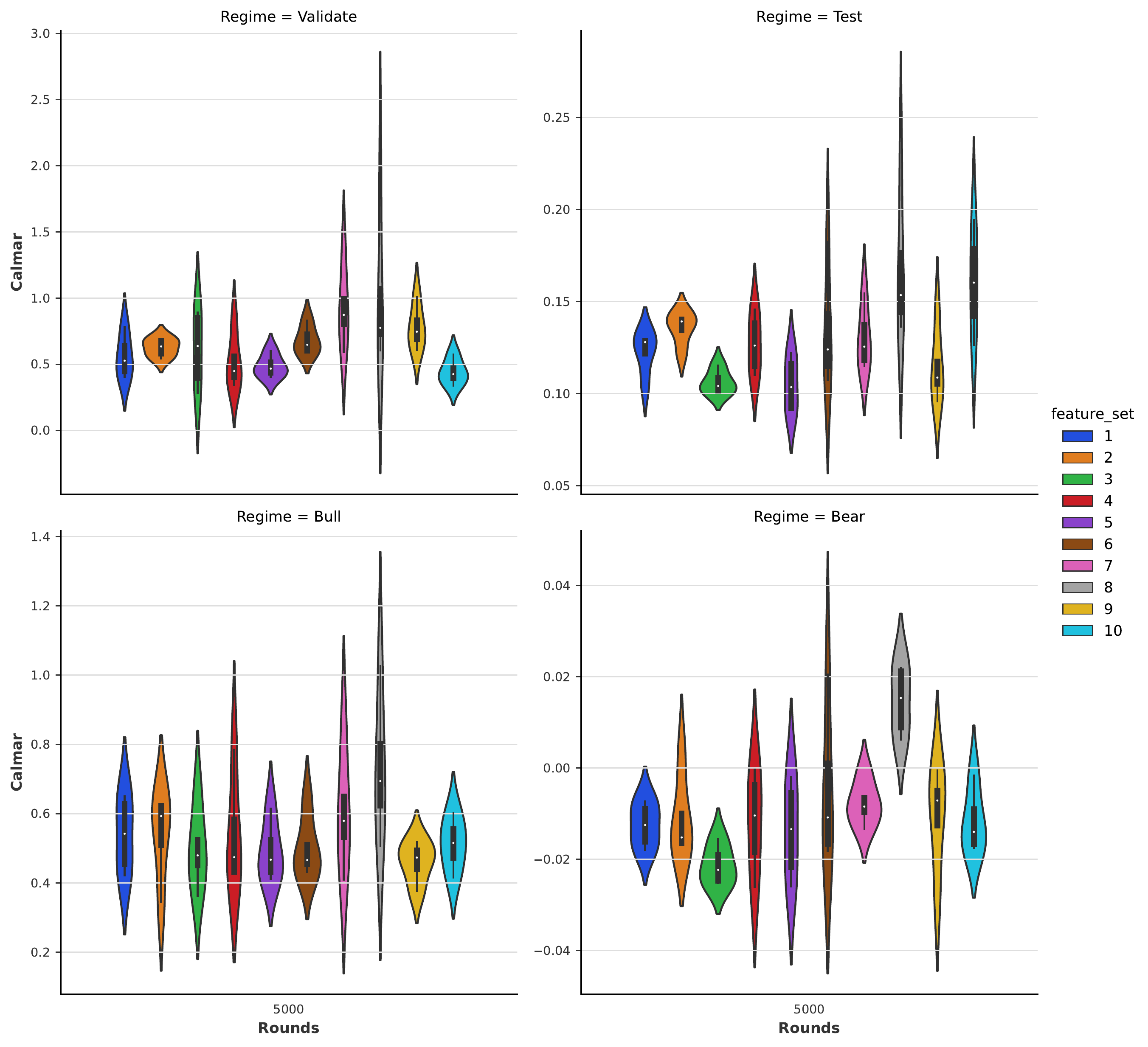}
        }
        \\
    \caption{Performances, (a) Mean Corr, (b) Sharpe ratio, and (c) Calmar ratio of the deep IL XGBoost models with random feature sampling under different market regimes. }
    \label{fig:Rain-IL-FeatureRandom-Layer1}
\end{figure}

%% Layer 1 Models Training Size
\begin{figure}[hbt!]
     \centering
        \subfloat[Mean Corr]{
          \includegraphics[width=12cm]{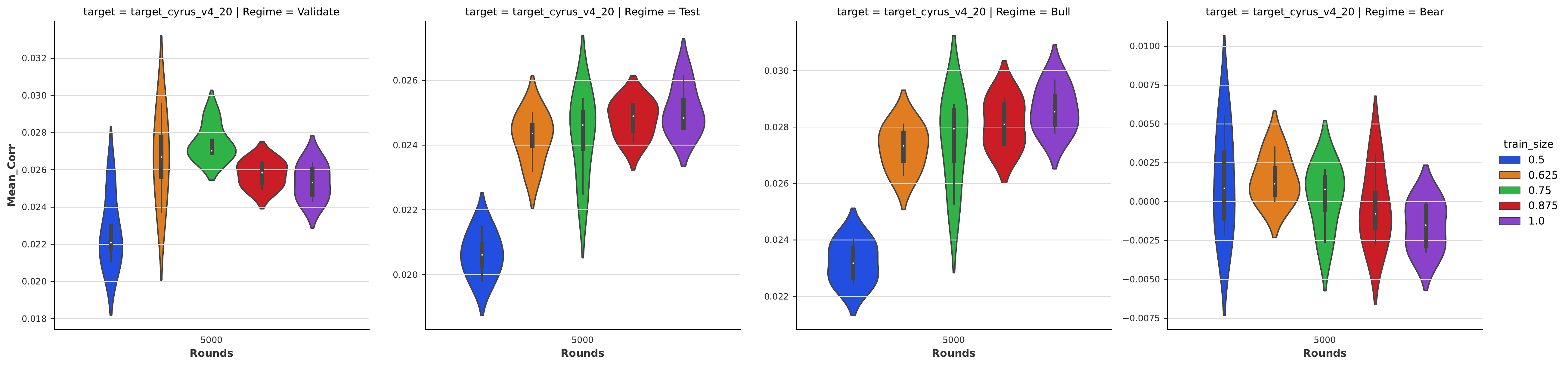}
        }
        \\
        \subfloat[Sharpe]{
          \includegraphics[width=12cm]{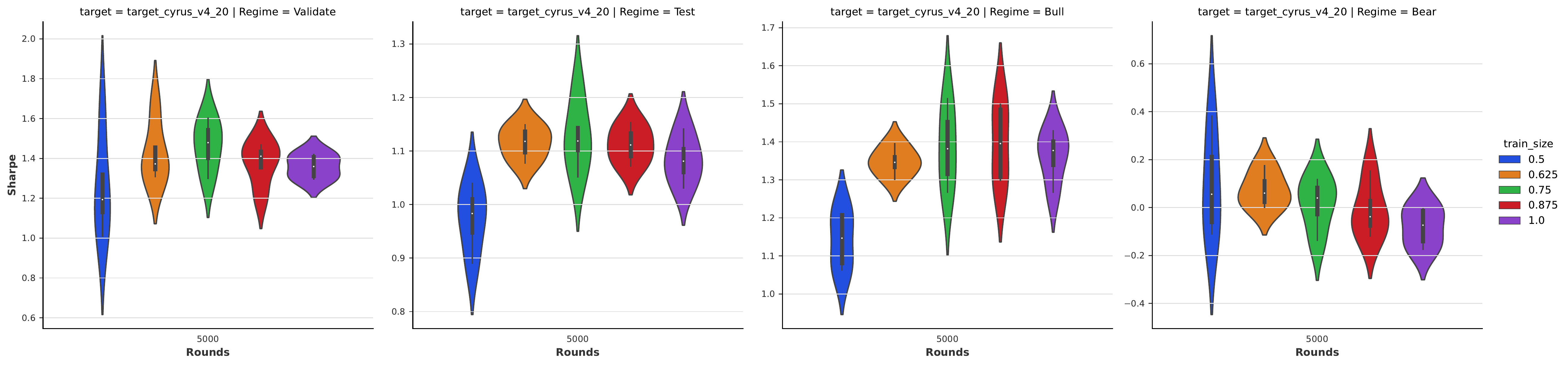}
        }
        \\
        \subfloat[Calmar]{
          \includegraphics[width=12cm]{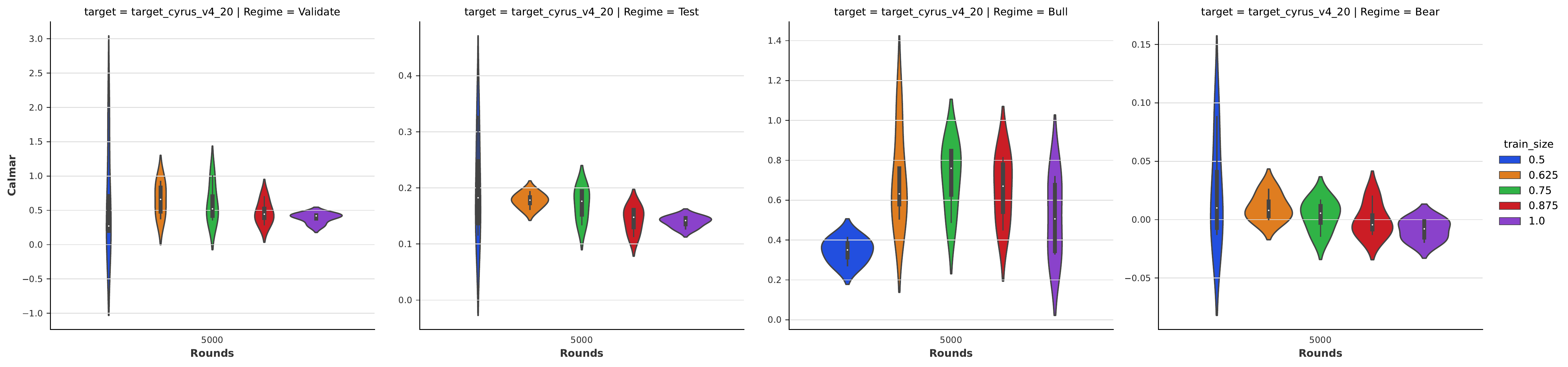}
        }
        \\
    \caption{Performances, (a) Mean Corr, (b) Sharpe ratio, and (c) Calmar ratio of the deep IL XGBoost models with different training sizes under different market regimes. }
    \label{fig:Rain-IL-TrainingSize-Layer1}
\end{figure}

%% Layer 1 Models Learning Rates Ensemble

%% Layer 1 Models Target Learning Rates Ensemble
\begin{figure}[hbt!]
     \centering
        \subfloat[Mean Corr]{
          \includegraphics[width=8cm]{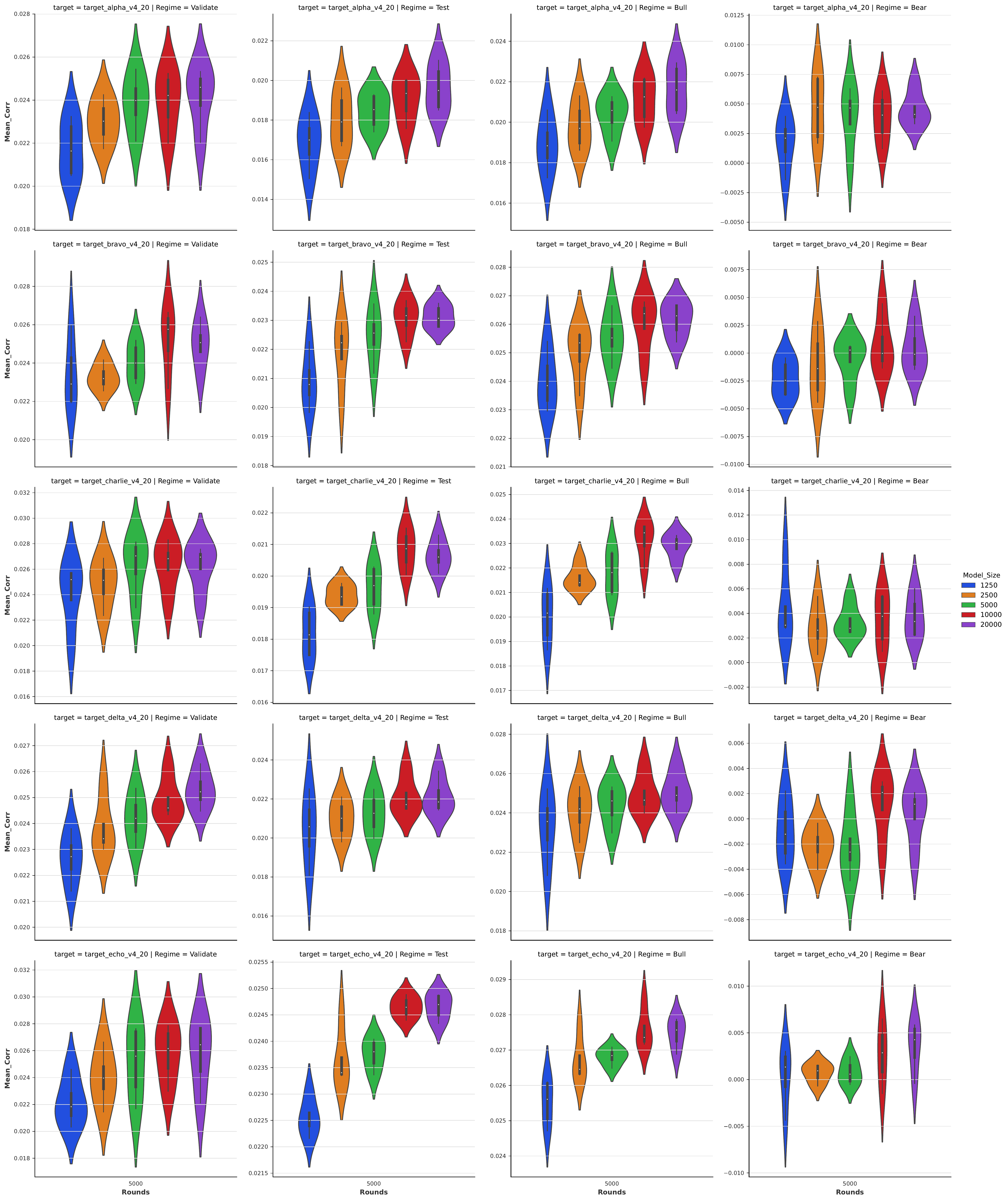}
        }
        \subfloat[Sharpe]{
          \includegraphics[width=8cm]{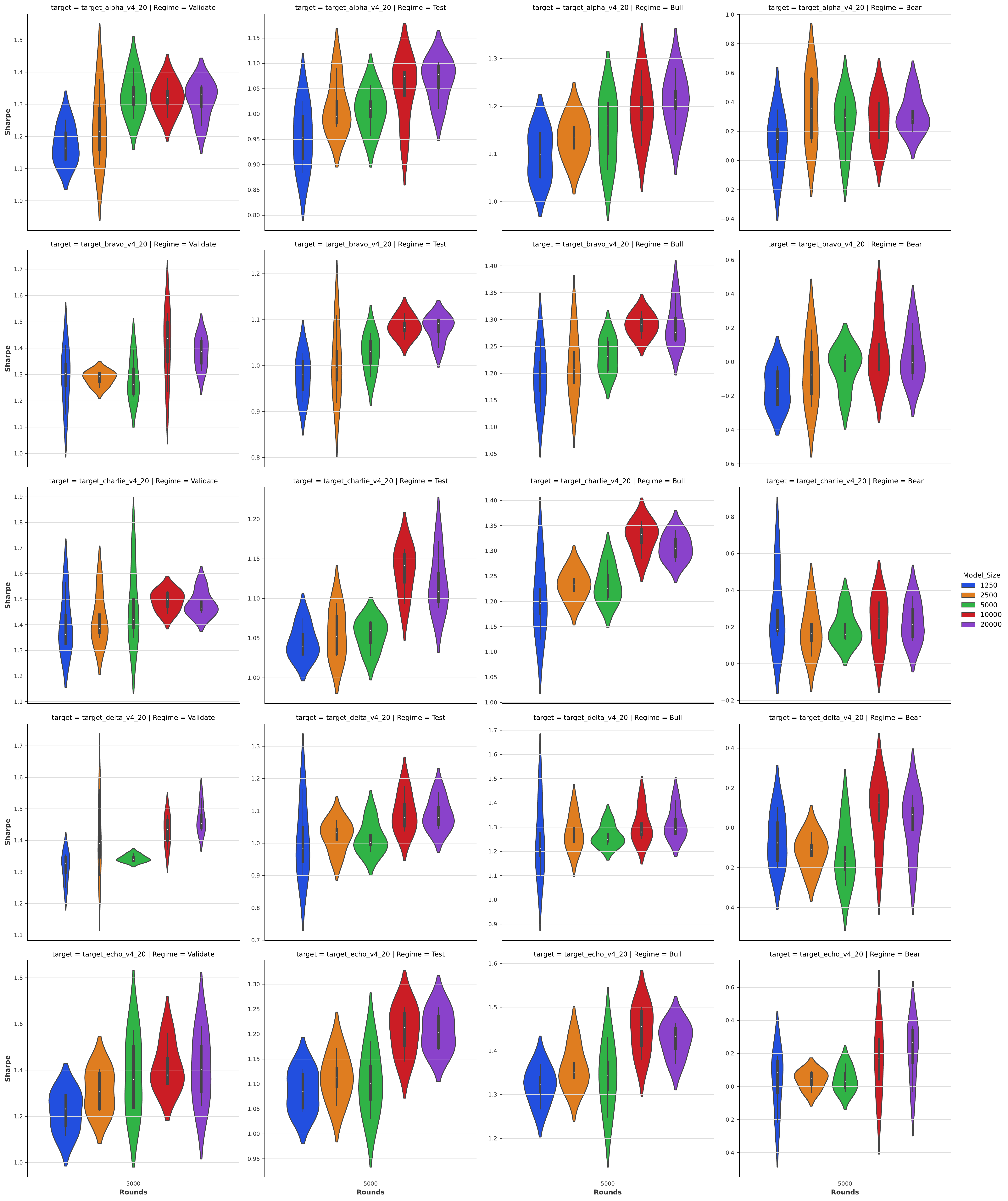}
        }
        \\
        \subfloat[Calmar]{
          \includegraphics[width=8cm]{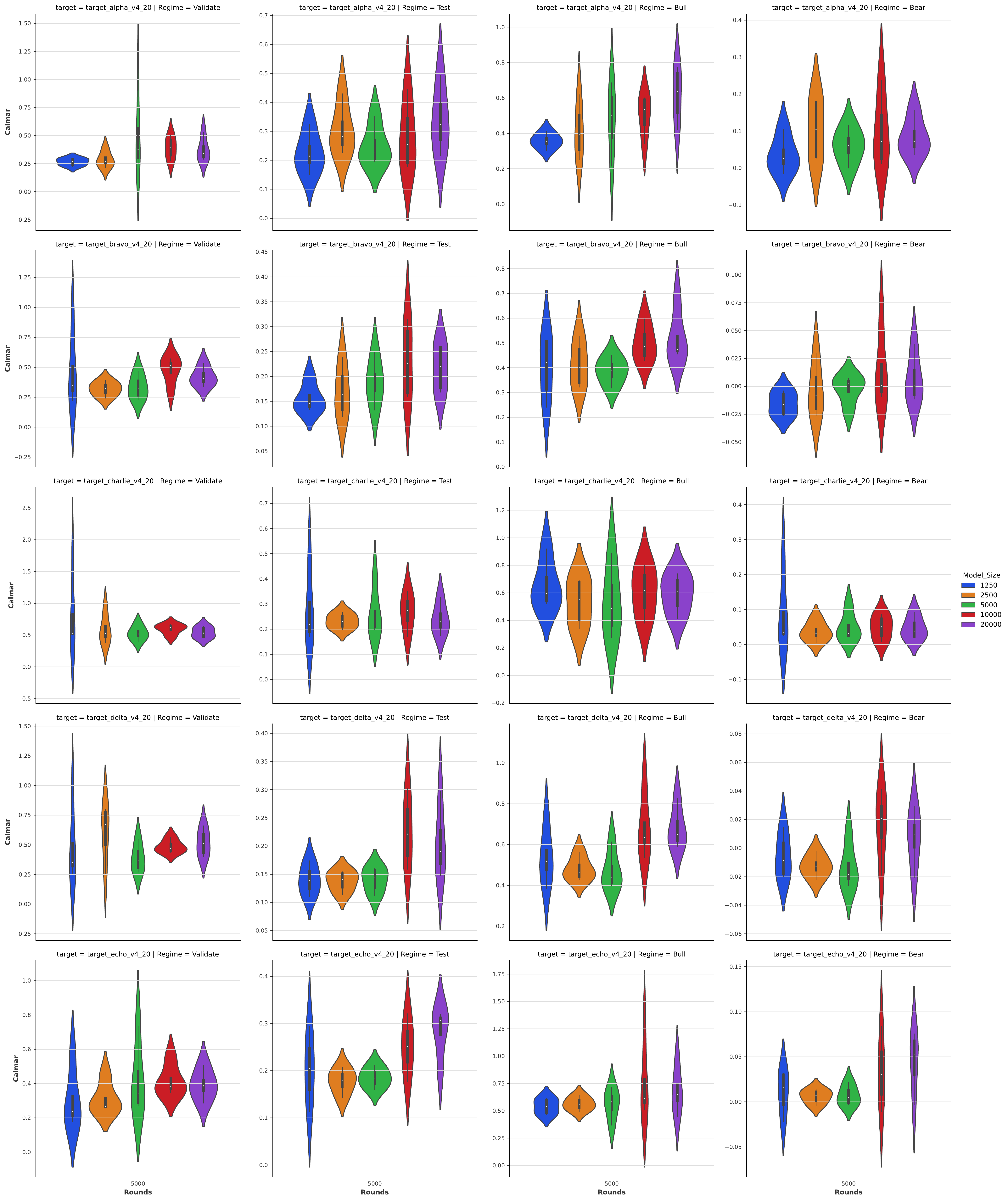}
        }
        \\
    \caption{Performances, (a) Mean Corr, (b) Sharpe ratio, and (c) Calmar ratio of the deep IL XGBoost models with different targets and learning rates under different market regimes. }
    \label{fig:Rain-IL-Target-LREnsemble-Layer1}
\end{figure}

%% Era Sampling 

%%% Learning Rate Optimisation 

%% FT Models 
\begin{figure}[hbt!]
     \centering
        \subfloat[EMA models]{
          \includegraphics[width=15cm]{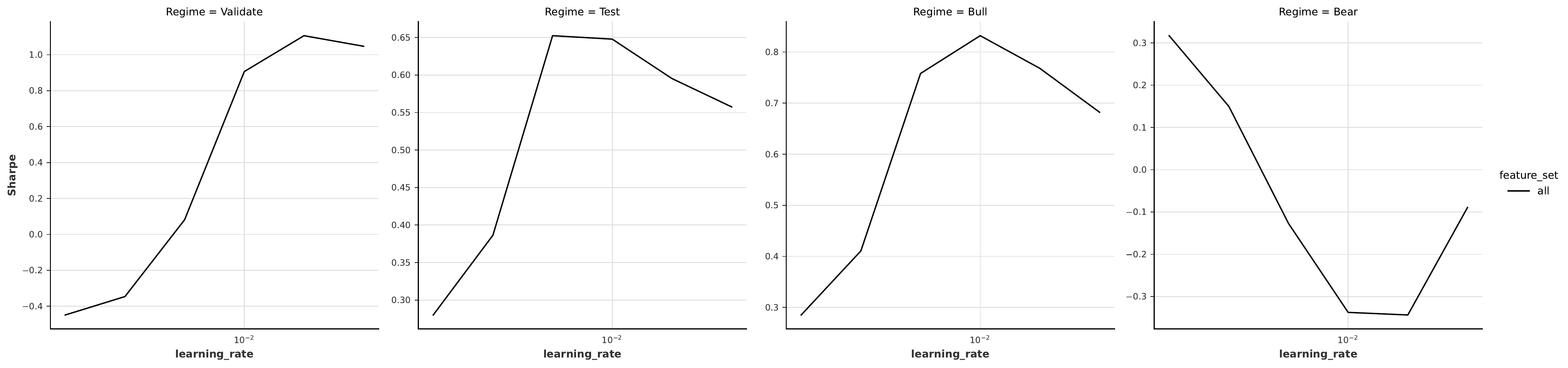}
        }
        \\
        \subfloat[Feature Transform factor-timing models]{
          \includegraphics[width=15cm]{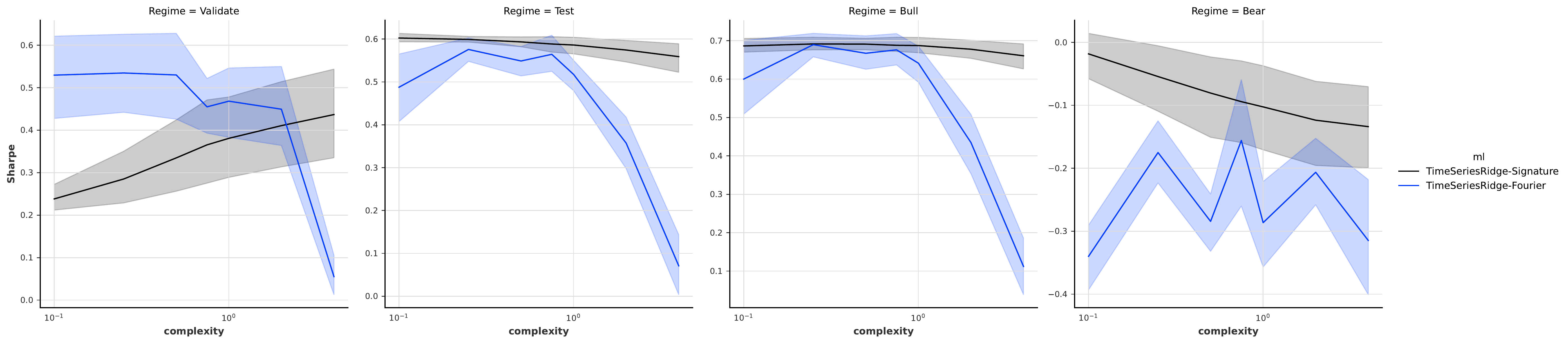}
        }
        \\
    \caption{Sharpe ratio of EMA and Feature Transform factor-timing models under different market regimes}
    \label{fig:Rain-Trend-Summary-Sharpe}
\end{figure}

%% FT Models 
\begin{figure}[hbt!]
     \centering
        \subfloat[EMA models]{
          \includegraphics[width=15cm]{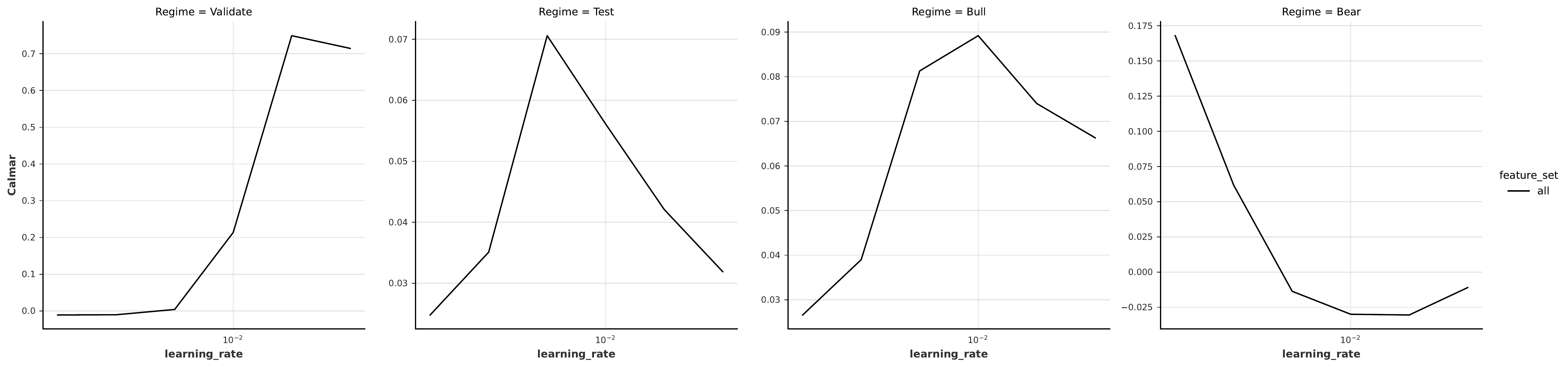}
        }
        \\
        \subfloat[Feature Transform factor-timing models]{
          \includegraphics[width=15cm]{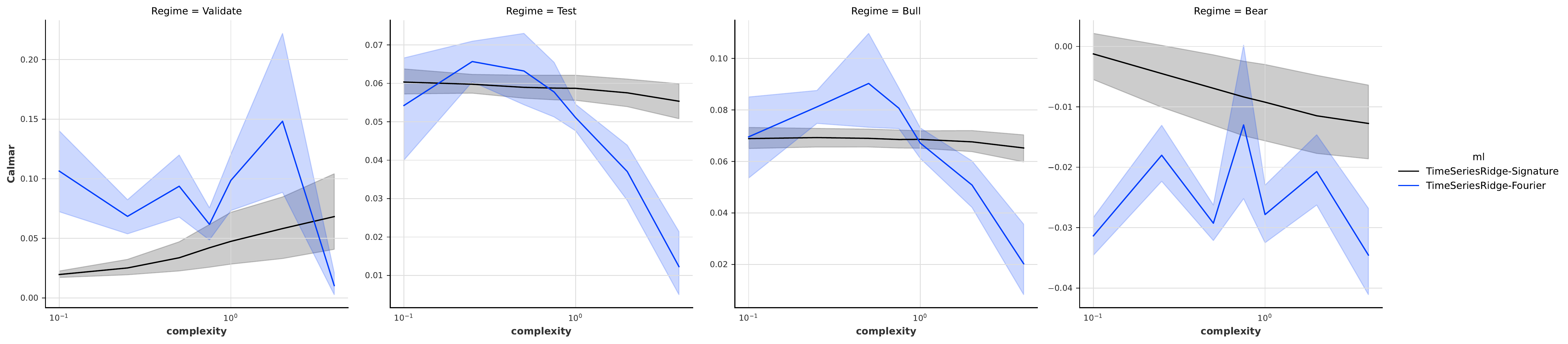}
        }
        \\
    \caption{Calmar ratio of EMA and Feature Transform factor-timing models under different market regimes}
    \label{fig:Rain-Trend-Summary-Calmar}
\end{figure}

%%% Dynamic Hedged Random Feature Sampling 
\begin{table}[hbt!]
\begin{tabular}{|l|c|l|l|l|}
\hline
Regime & Strategy & Mean Corr & Sharpe & Max Drawdown  \\ \hline
\multirow{5}{*}{Test}  & Example Model           & 0.0264  & 0.9626  & 0.2608  \\ \cline{2-5} 
                       & Baseline Model          & 0.0257  & 1.0983  & 0.1839  \\ \cline{2-5} 
                       & Tail Risk Model         & 0.0022  & 0.1607  & 0.2161  \\ \cline{2-5} 
                       & Static Hedged Model     & 0.0176  & 1.0507  & 0.0586  \\ \cline{2-5}
                       & Dynamic Hedged Model    & 0.0224  & 1.2378  & 0.0415  \\ \hline
\multirow{5}{*}{Bull}  & Example Model           & 0.0307  & 1.2512  & 0.0693 	 \\ \cline{2-5} 
                       & Baseline Model          & 0.0294  & 1.3962  & 0.0493  \\ \cline{2-5} 
                       & Tail Risk Model         & 0.0011  & 0.0794  & 0.2161  \\ \cline{2-5} 
                       & Static Hedged Model     & 0.0195  & 1.1858  & 0.0504  \\ \cline{2-5}
                       & Dynamic Hedged Model    & 0.0246  & 1.3731  & 0.0415  \\ \hline
\multirow{5}{*}{Bear}  & Example Model           & -0.0060 & -0.2306 & 0.2608  \\ \cline{2-5} 
                       & Baseline Model          & -0.0018 & -0.0831 & 0.1839  \\ \cline{2-5} 
                       & Tail Risk Model         & 0.0106  & 1.0225  & 0.0027  \\ \cline{2-5} 
                       & Static Hedged Model     & 0.0033  & 0.2970  & 0.0586  \\ \cline{2-5}
                       & Dynamic Hedged Model    & 0.0057  & 0.7733  & 0.0184   \\ \hline
\end{tabular}
\caption{Performances of Dynamic Hedged deep IL XGBoost ensemble model based on random feature sampling and V4.2 Example Model from Era 901 to Era 1070 under different market regimes. }
\label{table:Rain-DynamicHedge-FeatureRandom}
\end{table}

%%% Dynamic Hedged Training Set Size 
\begin{table}[hbt!]
\centering
\begin{tabular}{|l|c|l|l|l|}
\hline
Regime & Strategy & Mean Corr & Sharpe & Max Drawdown  \\ \hline
\multirow{5}{*}{Test}  & Example Model           & 0.0264  & 0.9626  & 0.2608  \\ \cline{2-5} 
                       & Baseline Model          & 0.0266  & 1.1559  & 0.1646  \\ \cline{2-5} 
                       & Tail Risk Model         & 0.0016  & 0.1068  & 0.3728  \\ \cline{2-5} 
                       & Static Hedged Model     & 0.0207  & 1.1337  & 0.0742  \\ \cline{2-5}
                       & Dynamic Hedged Model    & 0.0225  & 1.2646  & 0.0330  \\ \hline
\multirow{5}{*}{Bull}  & Example Model           & 0.0307  & 1.2512  & 0.0693 	 \\ \cline{2-5} 
                       & Baseline Model          & 0.0301  & 1.4305  & 0.0351  \\ \cline{2-5} 
                       & Tail Risk Model         & 0.0007  & 0.0481  & 0.3728  \\ \cline{2-5} 
                       & Static Hedged Model     & 0.0227  & 1.2941  & 0.0380  \\ \cline{2-5}
                       & Dynamic Hedged Model    & 0.0246  & 1.4305  & 0.0330  \\ \hline
\multirow{5}{*}{Bear}  & Example Model           & -0.0060 & -0.2306 & 0.2608  \\ \cline{2-5} 
                       & Baseline Model          & 0.0003  & 0.0151  & 0.1646  \\ \cline{2-5} 
                       & Tail Risk Model         & 0.0081  & 0.5826  & 0.0219  \\ \cline{2-5} 
                       & Static Hedged Model     & 0.0054  & 0.3409  & 0.0742  \\ \cline{2-5}
                       & Dynamic Hedged Model    & 0.0069  & 0.4871  & 0.0312   \\ \hline
\end{tabular}
\caption{Performances of Dynamic Hedged deep IL XGBoost ensemble model based on different training set sizes and V4.2 Example Model from Era 901 to Era 1070 under different market regimes. }
\label{table:Rain-DynamicHedge-TrainingSize}
\end{table}

%%% Dynamic Hedged Learning Rate  
\begin{table}[hbt!]
\centering
\begin{tabular}{|l|c|l|l|l|}
\hline
Regime & Strategy & Mean Corr & Sharpe & Max Drawdown  \\ \hline
\multirow{5}{*}{Test}  & Example Model           & 0.0264  & 0.9626  & 0.2608  \\ \cline{2-5} 
                       & Baseline Model          & 0.0265  & 1.1943  & 0.1562  \\ \cline{2-5} 
                       & Tail Risk Model         & 0.0015  & 0.1044  & 0.1754  \\ \cline{2-5} 
                       & Static Hedged Model     & 0.0199  & 0.9978  & 0.1460  \\ \cline{2-5}
                       & Dynamic Hedged Model    & 0.0207  & 1.0760  & 0.0871  \\ \hline
\multirow{5}{*}{Bull}  & Example Model           & 0.0307  & 1.2512  & 0.0693 	 \\ \cline{2-5} 
                       & Baseline Model          & 0.0300  & 1.5073  & 0.0343  \\ \cline{2-5} 
                       & Tail Risk Model         & 0.0011  & 0.0743  & 0.1754  \\ \cline{2-5} 
                       & Static Hedged Model     & 0.0227  & 1.2135  & 0.0377  \\ \cline{2-5}
                       & Dynamic Hedged Model    & 0.0233  & 1.2755  & 0.0434  \\ \hline
\multirow{5}{*}{Bear}  & Example Model           & -0.0060 & -0.2306 & 0.2608  \\ \cline{2-5} 
                       & Baseline Model          & -0.0001 & -0.0053 & 0.1562  \\ \cline{2-5} 
                       & Tail Risk Model         & 0.0051  & 0.2925  & 0.0743  \\ \cline{2-5} 
                       & Static Hedged Model     & -0.0008 & -0.0483 & 0.1460 \\ \cline{2-5}
                       & Dynamic Hedged Model    & 0.0015  & 0.0998  & 0.0871  \\ \hline
\end{tabular}
\caption{Performances of Dynamic Hedged deep IL XGBoost ensemble model based on different learning rates and V4.2 Example Model from Era 901 to Era 1070 under different market regimes. }
\label{table:Rain-DynamicHedge-LearningRate}
\end{table}

%%% Dynamic Hedged Targets  
\begin{table}[hbt!]
\centering
\begin{tabular}{|l|c|l|l|l|}
\hline
Regime & Strategy & Mean Corr & Sharpe & Max Drawdown  \\ \hline
\multirow{5}{*}{Test}  & Example Model           & 0.0264  & 0.9626  & 0.2608  \\ \cline{2-5} 
                       & Baseline Model          & 0.0247  & 1.1915  & 0.1026  \\ \cline{2-5} 
                       & Tail Risk Model         & -0.0001  & -0.0064  & 0.5444  \\ \cline{2-5} 
                       & Static Hedged Model     & 0.0189  & 0.9937  & 0.0401  \\ \cline{2-5}
                       & Dynamic Hedged Model    & 0.0220  & 1.2399  & 0.0457  \\ \hline
\multirow{5}{*}{Bull}  & Example Model           & 0.0307  & 1.2512  & 0.0693 	 \\ \cline{2-5} 
                       & Baseline Model          & 0.0277  & 1.4344  & 0.0411  \\ \cline{2-5} 
                       & Tail Risk Model         & -0.0017  & -0.0990  & 0.5444  \\ \cline{2-5} 
                       & Static Hedged Model     & 0.0204  & 1.0636  & 0.0401  \\ \cline{2-5}
                       & Dynamic Hedged Model    & 0.0234  & 1.3173  & 0.0457 \\ \hline
\multirow{5}{*}{Bear}  & Example Model           & -0.0060 & -0.2306 & 0.2608  \\ \cline{2-5} 
                       & Baseline Model          & 0.0020 & 0.1219 & 0.1026  \\ \cline{2-5} 
                       & Tail Risk Model         & 0.0118  & 0.8287  & 0.0148  \\ \cline{2-5} 
                       & Static Hedged Model     & 0.0078 & 0.5788 & 0.0383 \\ \cline{2-5}
                       & Dynamic Hedged Model    & 0.0115  & 0.8475  & 0.0244  \\ \hline
\end{tabular}
\caption{Performances of Dynamic Hedged deep IL XGBoost ensemble model based on different targets and V4.2 Example Model from Era 901 to Era 1070 under different market regimes. }
\label{table:Rain-DynamicHedge-Target}
\end{table}

\end{document}